\documentclass[journal]{IEEEtran}
\usepackage{bm}
\usepackage{amsfonts}
\usepackage{subfigure}
\usepackage{graphicx}
\usepackage{epstopdf}
\usepackage{float}

\usepackage{rotating}
\usepackage{algorithm,algorithmicx}
\usepackage[noend]{algpseudocode}
\usepackage{longtable}
\usepackage{multirow}
\usepackage{booktabs}
\usepackage{amsmath} 
\usepackage{array}
\usepackage{enumerate}
\ifCLASSINFOpdf
\else
\fi
\begin{document}
	%
	\title{Hyperspectral Image Classification With Context-Aware Dynamic Graph Convolutional Network}
	%
	%
	%
	
	\author{Sheng~Wan,
		Chen~Gong,~\IEEEmembership{Member,~IEEE},
		Ping~Zhong,~\IEEEmembership{Senior~Member,~IEEE},
		Shirui~Pan,
		Guangyu~Li,
		and~Jian~Yang,~\IEEEmembership{Member,~IEEE}
		
		\thanks{S. Wan, C. Gong, G. Li, and J. Yang are with the PCA Lab, the Key Laboratory of Intelligent Perception and Systems for High-Dimensional Information of Ministry of Education, the Jiangsu Key Laboratory of Image and Video Understanding for Social Security, and the School of Computer Science and Engineering, Nanjing University of Science and Technology, Nanjing, 210094, P.R. China. (e-mail: wansheng315@hotmail.com; chen.gong@njust.edu.cn; liguangyu0627@hotmail.com; csjyang@njust.edu.cn).}
		\thanks{P. Zhong is with the National Key Laboratory of Science and Technology on ATR, National University of Defense Technology, Changsha 410073, China (e-mail: zhongping@nudt.edu.cn).}
		\thanks{S. Pan is with Faculty of Information Technology, Monash University, Clayton, VIC 3800, Australia (Email: shirui.pan@monash.edu).}		
		\thanks{\emph{Corresponding~author:~Chen~Gong.}}
	}
	%
	%

	\markboth{}%
	{Shell \MakeLowercase{\textit{et al.}}: Bare Demo of IEEEtran.cls for IEEE Journals}
	%



	\maketitle
	
	\begin{abstract}
		In hyperspectral image (HSI) classification, spatial context has demonstrated its significance in achieving promising performance. However, conventional spatial context-based methods simply assume that spatially neighboring pixels should correspond to the same land-cover class, so they often fail to correctly discover the contextual relations among pixels in complex situations, and thus leading to imperfect classification results on some irregular or inhomogeneous regions such as class boundaries. To address this deficiency, we develop a new HSI classification method based on the recently proposed Graph Convolutional Network (GCN), as it can flexibly encode the relations among arbitrarily structured non-Euclidean data. Different from traditional GCN, there are two novel strategies adopted by our method to further exploit the contextual relations for accurate HSI classification. First, since the receptive field of traditional GCN is often limited to fairly small neighborhood, we proposed to capture long range contextual relations in HSI by performing successive graph convolutions on a learned region-induced graph which is transformed from the original 2D image grids. Second, we refine the graph edge weight and the connective relationships among image regions by learning the improved adjacency matrix and the `edge filter', so that the graph can be gradually refined to adapt to the representations generated by each graph convolutional layer. Such updated graph will in turn result in accurate region representations, and vice versa. The experiments carried out on three real-world benchmark datasets demonstrate that the proposed method yields significant improvement in the classification performance when compared with some state-of-the-art approaches.

	\end{abstract}

	%
	\IEEEpeerreviewmaketitle

	\section{Introduction}
	%
	%
	%
	%

	\IEEEPARstart{H}{yperspectral} image (HSI) has recently received considerable attention in a variety of applications such as military target detection, mineral identification, and disaster prevention \cite{298007}. In contrast to traditional panchromatic and multispectral remote sensing images, HSI consists of hundreds of contiguous spectral bands, which are helpful to distinguishing the targets with different materials. Thanks to the high spectral resolution, HSI has shown its advantages in identifying various land-cover types or targets \cite{Zhang2018Diverse}.
	
	Up to now, significant efforts have been made in developing diverse kinds of HSI classification methods. The early-staged algorithms are mainly based on the simple combination of spectral signatures and conventional pattern recognition methods, such as nearest neighbor classifier and Support Vector Machines (SVM) \cite{1323134, 5766028}. However, these methods isolatedly classify each image pixel without considering the spatial correlation among the pixels, so they will encounter the spectral variability problem \cite{8101519} and generate imperfect classification results.
	
	To address this shortcoming, the spatial context naturally becomes another type of useful information in addition to the spectra. It is now commonly acknowledged that the introduction of spatial context offers probability to improve HSI classification results and is the key to generating discriminative features for classification \cite{6297992}, hence there is a huge demand for the algorithms which can effectively discover and incorporate spatial context. During the past decades, researchers have reported various HSI classification methods utilizing spatial context. The first attempt was accomplished by Landgrebe and Kettig \cite{4071761}, where the well-known ECHO classifier is proposed to extract contextual information. After that, Markov Random Field (MRF) \cite{Wang2005A}, which is an undirected graphical model \cite{8271995}, became a popular approach to include spatial context for HSI classification. For instance, in \cite{5779697}, a relative homogeneity index for each pixel is introduced in MRF-based classification to determine an appropriate weighting coefficient for the contextual contribution. Apart from this, a novel framework combining Support Vector Machine (SVM) and MRF is proposed for contextual HSI classification \cite{6305471}. Meanwhile, many other models, like mathematical morphology, Gabor filtering, have also emerged as powerful tools for integrating the spatial context of HSI. For example, Benediktsson \emph{et} \emph{al.} \cite{1396321} constructed extended morphological profiles based on the use of opening and closing morphological transforms for preprocessing HSI. Besides, in \cite{5887411}, a set of complex Gabor wavelets with different frequencies and orientations are designed to represent the spatial context of pixels. Additionally, multiple kernel learning \cite{1576697} based on spatial context is proposed to improve the classification performance of SVM classifier on hyperspectral data. Although the use of spatial context improves classification results in smooth image areas \cite{7147814}, the aforementioned methods do not explicitly investigate the contextual relations among individual pixels (or regions), and they only implicitly assume that nearby pixels have a large probability to take the same class label regardless whether they are in object boundary regions or homogeneous regions \cite{7117347}. As a result, the semantic meaning carried by image patches cannot be well preserved, and the classification errors may appear within the object area of a certain class. Additionally, the pixels around some irregular or inhomogeneous regions are also very likely to be misclassified due to the inappropriate utilization of local contextual information. Consequently, the simple assumption of smoothness and homogeneity over the whole image is unreasonable.
	
	In order to effectively and precisely exploit the contextual relations in HSI, in this paper, we propose a novel `Context-Aware Dynamic Graph Convolutional Network' (CAD-GCN) which includes the following three key techniques: (1) The incorporation of Graph Convolutional Network (GCN) for sufficiently exploiting contextual relations among pixels; (2) The employment of graph projection and re-projection framework for exploring long range contextual relations; and (3) The utilization of dynamic graph refinement for accurately characterizing contextual relations and timely finding precise region representations.
	
	Specifically, in our CAD-GCN, the recently proposed GCN \cite{Kipf2016Semi} is utilized. GCN is the extension of Convolutional Neural Network (CNN) for the non-grid data, and is able to aggregate features and propagate information across graph nodes. Consequently, the convolution operation of GCN is adaptively dominated by the neighborhood structure and can be applied to the non-Euclidean irregular data based on the graph which encodes contextual relations among graph nodes. As a result, the complex regions such as target boundaries in HSI can be flexibly preserved by GCN.
	
	Since graph convolution is usually conducted based on neighborhood structure, its receptive field is merely limited to a fairly small region, and thus it fails to capture long range contextual relations among faraway pixels. Although constructing a dense fully-connected graph can address this deficiency, the computational cost will become unbearably heavy \cite{Wang2017Non, Yin2018Beyond}. Meanwhile, deep stacking of local operations for creating a large receptive field with long range context has also been proven inefficient \cite{Luo2016Understanding}. Different from above approaches, by learning to project the original HSI into a region-induced graph, our proposed CAD-GCN moves beyond regular squared image grids and encodes contextual relations among regions. Then inference can be performed on the graph through passing messages between regions and along the edges connecting them. Therefore, this inference can not only update the region features, but also connect the regions which are originally far away in the 2D space by successive graph convolutions. As a result, the long range relations between faraway image regions can be effectively exploited. After that, the proposed CAD-GCN can learn an efficient graph representation with only a small number of nodes. Finally, the learned region-level features can be interpolated into the 2D feature map by reverting the pixel-to-region assignment from the previous graph projection step, so that the pixel-level features can be obtained to fully comply with the existing networks.
	
	Nevertheless, the contextual relations revealed by a predefined fixed graph \cite{Kipf2016Semi} for implementing GCN is still inadequate for HSI classification. Since the predefined graph based on the Euclidean distance may not be suitable for measuring their real similarities \cite{Li2018Adaptive}, we aim to learn the improved similarity measurement between image regions, in order to better characterize the contextual relationships among them. Specifically, the graph can be dynamically updated to adapt to the region representations generated by each graph convolutional layer, which will in turn make the representations more accurate. Meanwhile, since the learned graph may contain improper inter-class connections, especially around the boundaries between the regions of different classes, we introduce the `edge filter' which can filter out the incorrect inter-class edges and refine the contextual relations represented by the graph. Intensive experimental results on three typically used HSI datasets reveal the superiority of the proposed CAD-GCN when compared with the exiting methods.

	\section{Related Works}
	\label{Relatedworks}
	
	In this section, we review some representative works on HSI classification and GCN, as they are related to this work.
	
	\subsection{Hyperspectral Image Classification}
	
	As a traditional yet critical technique, HSI classification has been an active research topic in the field of remote sensing. During the past few decades, diverse kinds of methods have been developed for HSI classification, which can be roughly divided into two categories: spectral-based methods and spectral-spatial-based methods. Many classical HSI classification approaches are only based on spectral information \cite{1323134, 5942156} and ignore the crucial spatial information contained in HSI, which may decrease the classification performance \cite{8271995}. As a result, spatial context, which has been observed to be arguably more effective than spectral signatures \cite{6297992, Zhang2016Simultaneous}, has been incorporated, to acquire better classification results. In the following, several typical methodologies for capturing the spectral-spatial information of HSI are reviewed:
	
	\begin{enumerate}
		\item \emph{Structural filtering}. Structural Filtering-based methods have been widely studied for spatial preprocessing of HSI, where spatial features are often generated via structural filtering. Many researchers have been working on this direction, and one of the most simple but effective ways is to extract spatial information from a given region based on the moment criteria, such as the mean or standard deviation of adjacent pixels in a window \cite{8101519}. The inclusion of moments or cumulants has been intensively investigated in the field of composite kernel learning and multiple kernel learning \cite{6912942, 6926746}. Meanwhile, the adaptations of neighboring moments or cumulants have been widely adopted for HSI classification to explore the spatial homogeneity, which can preserve the image details as well \cite{4683346}. In additional to structural filtering, another research direction is to perform local harmonic analysis, which includes spatial translation-invariant spectral-spatial wavelet features, spectral-spatial Gabor features, and empirical mode decomposition-based features \cite{6923420, 7046411}. Besides, another trend in spatial filtering is to extract features with adaptive structures, such as adaptive region-based filtering \cite{7080913}. 
		\item \emph{Mathematical morphology}. Mathematical morphology is a powerful tool for analyzing and processing geometrical structures in spatial domain \cite{7064745}. In \cite{905239}, the morphological profile (MP) is introduced for classifying images with very high spatial resolutions by using a sequence of geodesic opening and closing operations. Furthermore, Benediktsson \emph{et} \emph{al.} \cite{1396321} developed the Extended MP (EMP), where an MP is computed on each component after reducing the dimensionality of data. However, the method in \cite{1396321} does not fully exploit the spectral information of hyperspectral data. In order to overcome this deficiency, Fauvel \emph{et} \emph{al.} \cite{4686022} proposed a spectral and spatial fusion approach based on EMP and the original data. Meanwhile, by extending the concept of MP and EMP, Attribute Profile (AP) \cite{5482208} was proposed to extract additional spatial features for HSI classification. Since then, AP and its extensions, including extended AP \cite{5664759} and extended multi-attribute profiles \cite{6410414}, have attracted increasing attention in HSI classification.
		\item \emph{Superpixel establishment}. In HSI, each superpixel corresponds to a cluster of spatially connected and spectrally similar pixels. The superpixels can also be regarded as small local regions with diverse shapes and sizes. Recently, some works focus on developing segmentation-based methods for HSI classification with superpixel technique \cite{7097693, 7119598}, in order to jointly combine the spectral-spatial correlations and discrimination to improve classification performance. For instance, in \cite{8291607}, superpixel technique is utilized to generate homogeneous region before constructing a graph on superpixels, which produces satisfactory classification results.
	\end{enumerate}
	
	\subsection{Graph Convolutional Network}
	
	Following the practice of conventional CNN, the concept of graph convolution can be defined by spatial methods in the vertex domain. Concretely, the convolution is defined as a weighted average function over the neighbors of each node, where the function characterizes the impact exerting to the target node from its neighboring nodes. For instance, in GraphSAGE \cite{Hamilton2017Inductive}, the weighting function is built by using various aggregators over the neighboring nodes. Besides, graph attention network \cite{velivckovic2017graph} proposes to learn the weighting function via self-attention mechanism. For spatial methods, one of the open challenges is how to construct an appropriate neighborhood for the target node when defining graph convolution.	
	
	In additional to spatial methods, graph convolution can also be defined by spectral methods via convolution theorem. As the pioneering work of spectral methods, spectral CNN \cite{Bruna2014Spectral} converts signals defined in vertex domain into spectral domain by leveraging graph Fourier transform, where the convolution kernel is taken as a set of learnable coefficients associated with Fourier bases (i.e., the eigenvectors of Laplacian matrix). However, this method depends on the eigen-decomposition of Laplacian matrix, which will lead to extremely high computational complexity on large-scale graphs. Subsequently, ChebyNet \cite{Defferrard2016Convolutional} considers the convolution kernel as a polynomial function of the diagonal matrix containing the eigenvalues of Laplacian matrix. Afterwards, GCN \cite{Kipf2016Semi} was proposed by Kipf and Welling via using a localized first-order approximation to ChebyNet, which brings about more efficient filtering operations than spectral CNN.
	
	Recently, GCN gains remarkable success in processing graph-structured data and has been widely adopted in many areas, such as social network mining \cite{Lei2009Relational}, recommendation system \cite{Ying2018GraphSIGKDD}, natural language processing \cite{8029788}, and scene understanding \cite{8100020}. Due to the effectiveness of GCN in handling the non-Euclidean data, we plan to employ GCN to capture the contextual relations among pixels in HSI. To the best of our knowledge, only one prior work has employed GCN for HSI classification, i.e., \cite{8474300}. However, \cite{8474300} only utilizes the original fixed structure of GCN, and thus the intrinsic relations among pixels cannot be precisely explored. Moreover, the limited receptive field will degrade the convolution performance in HSI. To address these problems, we proposed a novel context-aware dynamic GCN which dynamically refines the graph along with graph convolution process and captures long range contextual relations among the image pixels via using a graph projection technique. As a result, an improved graph representation can be learned, and the performance of HSI classification can be enhanced as well.

	\section{The Proposed Method}
	
	\begin{figure*}[!t]
		\centering
		\centering
		\includegraphics[width=17cm]{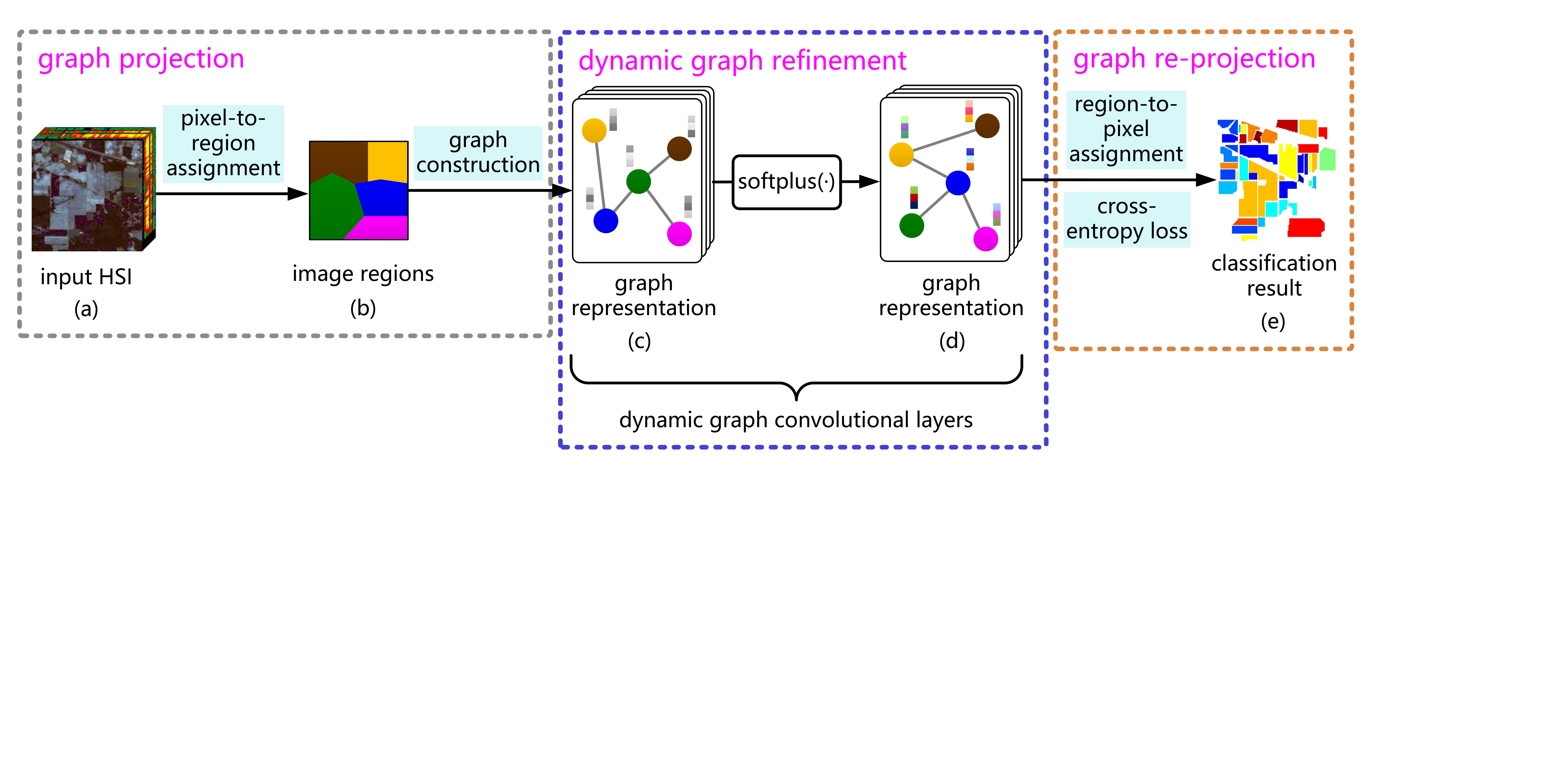}\hspace{0pt}
		\caption{The framework of our proposed CAD-GCN. (a) is the original HSI. (b) shows the five regions from the original HSI obtained via pixel-to-region assignment. (c) and (d) denote two dynamic graph convolutional layers, where the circles with different colors correspond to different image regions (i.e., graph nodes) and the gray lines represent graph edges. From (c) to (d), both the edge weight and the connective relationships among regions can be dynamically refined during the convolution operation, and thus the improved graph structure and node representations can be obtained. In our model, softplus \cite{7280459} is utilized as the activation function. In (e), the learned graph representation can be interpolated back into 2D image grids based on the region-to-pixel assignment, and then the cross-entropy loss is used to penalize the label differences between the network output and the originally labeled pixels.}
		\label{Overview}
	\end{figure*}	
	
	This section details the proposed CAD-GCN model, of which the pipeline is presented in Fig.~\ref{Overview}. Given an input image (Fig.~\ref{Overview}(a)), we first obtain its region features (Fig.~\ref{Overview}(b)) by learning to project the original image with 2D pixel grids into graph data. Then dynamic graph convolution (Fig.~\ref{Overview}(c) and Fig.~\ref{Overview}(d)) is conducted to refine the acquired region graph, along with encoding features for each region. Finally, the classification result (Fig.~\ref{Overview}(e)) is produced by interpolating the learned graph representation into 2D grids based on the region-to-pixel assignment. The critical operations in the proposed CAD-GCN will be detailed by presenting the GCN backbone (Section~\ref{GCNBackbone}), explaining the graph projection with pixel-to-region assignment (Section~\ref{SecP2R}), describing the dynamic graph refinement (Section~\ref{DynamicGraph}), and elaborating the graph re-projection with region-to-pixel assignment (Section~\ref{SecR2P}).

	\subsection{Graph Convolutional Network}
	\label{GCNBackbone}
	
	Inspired by CNN, GCN \cite{Kipf2016Semi} is a multi-layer neural network which directly operates on a graph and aims to extract high-level features through aggregating feature information from the neighborhoods of graph nodes. In GCN, an undirected graph is formally defined as $\mathcal{G}=(\mathcal{V}, \mathcal{E})$ with $\mathcal{V}$ and $\mathcal{E}$ denoting the sets of nodes and edges, respectively. The notation $\mathbf{A}$ denotes the adjacency matrix of $\mathcal{G}$ which indicates the existence of an edge between each pair of nodes, and its $(i, j)^{\rm{th}}$ element can be calculated as
	\begin{equation}
	\label{InitialAdjacencyMatirx}
	{\mathbf{A}_{ij}} = \left\{ {\begin{array}{*{20}{l}}
		{{e^{ - \gamma {{\left\| {{\mathbf{x}_i} - {\mathbf{x}_j}} \right\|}^2}}}}\,\;{\rm{if}}\,\;{\mathbf{x}}_{i} \in N(\mathbf{x}_j)\,\;{\rm{or}}\,\;{\mathbf{x}}_{j} \in N(\mathbf{x}_i)\\
		0 \;\;\;\;\;\;\;\;\;\;\;\;\;\;\;\;\;\;\rm{otherwise}
		\end{array}} \right.,
	\end{equation}
	where the parameter $\gamma$ is empirically set to 0.2 in our experiments, $\mathbf{x}_i$ and $\mathbf{x}_j$ represent two graph nodes (i.e., image regions in this paper), and $N(\mathbf{x}_j)$ is the set of neighbors of $\mathbf{x}_j$.
	
	First, in order to conduct node embedding for $\mathcal{G}$, spectral filtering on the graph is defined, which can be expressed as the multiplication of a signal $\mathbf{x}$ with a filter $g_{\bm{\theta}}=\rm{diag}(\bm{\theta})$ in the Fourier domain, i.e.,
	\begin{equation}
	\label{FilterFourier}
	{g_{\bm{\theta}} }\star\mathbf{x} = \mathbf{U}{g_{\bm{\theta}} }{\mathbf{U}^{\top}}\mathbf{x},
	\end{equation}
	where $\mathbf{U}$ is the matrix of eigenvectors of normalized graph Laplacian $\mathbf{L}=\mathbf{I}-{\mathbf{D}^{ - \frac{1}{2}}}\mathbf{A}{\mathbf{D}^{ - \frac{1}{2}}}=\mathbf{U}\mathbf{\Lambda}\mathbf{U}^{\top}$. Here $\mathbf{\Lambda}$ denotes a diagonal matrix composed of the eigenvalues of $\mathbf{L}$, $\mathbf{D}$ is the degree matrix with the diagonal element ${\mathbf{D}_{ii}} = \sum\nolimits_j {{\mathbf{A}_{ij}}}$, and $\mathbf{I}$ represents the identity matrix with proper size throughout this paper. Then $g_{\bm{\theta}}$ can be understood as a function of eigenvalues of $\mathbf{L}$, i.e., $g_{\bm{\theta}}(\mathbf{\Lambda})$. To reduce the computational cost of eigen-decomposition in Eq.~\eqref{FilterFourier}, Hammond \emph{et} \emph{al.} \cite{Hammond2009Wavelets} approximated $g_{\bm{\theta}}(\mathbf{\Lambda})$ using a truncated expansion in terms of Chebyshev polynomials $T_k(\mathbf{x})$ up to $K^{\rm{th}}$-order, namely
	\begin{equation}
	{g_{\bm{\theta} '}}(\mathbf{\Lambda} ) \approx \sum\limits_{k = 0}^K {\bm{\theta} {'_k}{T_k}(\widetilde{ \mathbf{\Lambda}} )},
	\end{equation}
	where $\bm{\theta} {'}$ denotes a vector of Chebyshev coefficients, and $\widetilde{\mathbf{\Lambda}}=\frac{2}{{{\lambda _{\max }}}}\mathbf{\Lambda}  - {\mathbf{I}}$ with $\lambda _{\max }$ being the largest eigenvalue of $\mathbf{L}$. According to \cite{Hammond2009Wavelets}, the Chebyshev polynomials can be defined as ${T_k}(\mathbf{x}) = 2\mathbf{x}{T_{k - 1}}(\mathbf{x}) - {T_{k - 2}}(\mathbf{x})$, where $T_{0}(\mathbf{x})=1$ and $T_{1}(\mathbf{x})=\mathbf{x}$. Hence, we have the convolution of a signal $\mathbf{x}$ as
	\begin{equation}
	\label{ConvoX}
	{g_{\bm{\theta}{'}}  }\star\mathbf{x} \approx \sum\limits_{k = 0}^K {\bm{\theta} {'_k}{T_k}(\widetilde{\mathbf{L}} )\mathbf{x}},
	\end{equation}
	where $\widetilde{\mathbf{L}}=\frac{2}{{{\lambda _{\max }}}}\mathbf{L}  - {\mathbf{I}}$ is the scaled Laplacian matrix. Eq.~\eqref{ConvoX} can be easily verified according to the fact that $(\mathbf{U}\mathbf{\Lambda}\mathbf{U}^{\top})^k=\mathbf{U}\mathbf{\Lambda}^k\mathbf{U}^{\top}$. As is can be observed, this expression is a $K^{\rm{th}}$-order polynomial regarding the Laplacian (i.e., $K$-localized). In other words, the filtering only depends on the nodes that are at most $K$ steps away from the central node. In our CAD-GCN model, the first-order neighborhood is considered, i.e., $K=1$, and thus Eq.~\eqref{ConvoX} turns to a linear function on the graph Laplacian spectrum with respect to $\mathbf{L}$.
	
	Afterwards, a neural network based on graph convolutions can be built by stacking multiple convolutional layers in the form of Eq.~\eqref{ConvoX}, where each layer is followed by an element-wise non-linear operation (i.e., softplus($\cdot$) \cite{7280459}). By this way, we can derive diverse classes of convolutional filter functions through stacking multiple layers of the same configuration. With the linear formulation, Kipf and Welling \cite{Kipf2016Semi} further approximated $\lambda_{\rm{max}}\approx 2$, considering that the network parameters can adapt to this change in scale during the training process. Therefore, Eq.~\eqref{ConvoX} is simplified to
	\begin{equation}
	\label{ConvoSim1}
	{g_{\bm{\theta}{'}} }\star\mathbf{x} \approx \bm{\theta} {'_0}\mathbf{x}+\bm{\theta} {'_1}(\mathbf{L}-\mathbf{I})\mathbf{x}=\bm{\theta} {'_0}\mathbf{x}-\bm{\theta} {'_1}{\mathbf{D}^{ - \frac{1}{2}}}\mathbf{A}{\mathbf{D}^{ - \frac{1}{2}}}\mathbf{x},
	\end{equation}
	where $\bm{\theta} {'_0}$ and $\bm{\theta} {'_1}$ are two free parameters. Since reducing the number of parameters helps to avoid overfitting, Eq.~\eqref{ConvoSim1} is further converted to
	\begin{equation}
	\label{ConvoSim2}
	{g_{\bm{\theta}}}\star\mathbf{x} \approx {\bm\theta}(\mathbf{I}+\mathbf{D}^{ - \frac{1}{2}}\mathbf{A}{\mathbf{D}^{ - \frac{1}{2}}})\mathbf{x}
	\end{equation}
	by letting $\bm{\theta}=\bm{\theta} {'_0}=-\bm{\theta} {'_1}$. As $\mathbf{I}+\mathbf{D}^{ - \frac{1}{2}}\mathbf{A}{\mathbf{D}^{ - \frac{1}{2}}}$ has the eigenvalues in the range $[0, 2]$, repeatedly applying this operator will result in numerical instabilities and exploding/vanishing gradients in a deep network. To solve this deficiency, Kipf and Welling \cite{Kipf2016Semi} performed the re-normalization trick $\mathbf{I}+\mathbf{D}^{ - \frac{1}{2}}\mathbf{A}{\mathbf{D}^{ - \frac{1}{2}}} \to \widetilde{\mathbf{D}}^{ - \frac{1}{2}}\widetilde{\mathbf{A}}{\widetilde{\mathbf{D}}^{ - \frac{1}{2}}}$ with $\widetilde{\mathbf{A}} = \mathbf{A} + \mathbf{I}$ and $\widetilde{\mathbf{D}}_{ii} = \sum\nolimits_j {{\widetilde{\mathbf{A}}_{ij}}}$. As a result, the convolution operation of GCN model can then be expressed as
	\begin{equation}
	{\mathbf{H}^{(l)}} = \sigma (\widetilde{\mathbf{A}}{\mathbf{H}^{(l-1)}}{\mathbf{W}^{(l)}}),
	\end{equation}
	where ${\mathbf{H}^{(l)}}$ denotes the output of the $l^{\rm{th}}$ layer, $\sigma ( \cdot )$ represents an activation function, such as the softplus function \cite{7280459} used in our proposed CAD-GCN, and $\mathbf{W}^{(l)} $ is the trainable weight matrix involved in the $l^{\rm{th}}$ layer.
	
	\subsection{Pixel-to-Region Assignment}
	\label{SecP2R}

	Although GCN is able to capture contextual relations among image pixels, the receptive field of pixel-level graph convolution is often limited \cite{Yin2018Beyond}. In order to effectively characterize long range relations among pixels, we intend to move beyond regular 2D image grids and encode contextual relations among regions, since the dependencies among image regions are of much longer than those captured by pixel-level convolutions \cite{Yin2018Beyond}. The main idea is learning pixel-to-region assignment which groups pixels with similar features into coherent regions, in order to capture contextual relations among the regions originally far away in the original 2D space.

	Different from the conventional region-based methods \cite{7117347, 8291607} which start by coarsely grouping pixels into certain regions, we aim at learning to transform the original HSI into a region graph, and this process is called graph projection. Specifically, a soft assignment matrix which is parameterized by $\mathbf{V}\in\mathbb{R}^{d\times c}$ will be learned by the network to assign each pixel $\mathbf{z}_i\in\mathbb{R}^{d}$ to its neighboring regions, where $d$ denotes the spectral dimensionality of each pixel, $c$ is the number of image regions, and each column $\mathbf{v}_i\in\mathbb{R}^{d}$ of $\mathbf{V}$ corresponds to the anchor point of a region. Then the soft assignment matrix $\mathbf{P}\in\mathbb{R}^{n\times c}$ can be computed as
	\begin{equation}
	\label{AssignmentMatrixP}
	{{\bf{P}}_{ij}} = \left\{ {\begin{array}{*{20}{l}}
		{{e^{ - \gamma {{\left\| {{{\bf{z}}_i} - {{\bf{v}}_j}} \right\|}^2}}}  }\,\;\;\;{\rm{if}}\,\;{\mathbf{v}_j \in \widetilde N({{\bf{z}}_i})}\\
		0 \;\;\;\;\;\;\;\;\;\;\;\;\;\;\;\;\;\;\;\,\rm{otherwise}
		\end{array}} \right.,
	\end{equation}
	where $n$ is the number of image pixels, $\widetilde{N}(\mathbf{z}_i)$ denotes the set of neighboring regions connected to the pixel $\mathbf{z}_i$. To be more specific, $\widetilde{N}(\mathbf{z}_i)$ includes not only the central region where $\mathbf{z}_i$ resides, but also the regions adjacent that are to the central one. In Eq.~\eqref{AssignmentMatrixP}, the element $\mathbf{P}_{ij}$ defines the soft assignment of a pixel $\mathbf{z}_i$ to $\mathbf{v}_j$. With the learned pixel-to-region assignment, the region feature $\mathbf{x}_j$ can be encoded by
	\begin{equation}
	\label{GraphProjection}
	{{\mathbf{x}}_j} = \frac{{\sum\nolimits_i {{{\mathbf{P}}_{ij}}{{\mathbf{z}}_i}} }}{{\sum\nolimits_i {{{\mathbf{P}}_{ij}}} }}.
	\end{equation}
	By learning the features for each region, the negative impact of inaccurate pre-computed region features can be reduced.
	
	\begin{figure}[!t]
		\centering
		\centering
		\resizebox*{!}{5cm}{\includegraphics{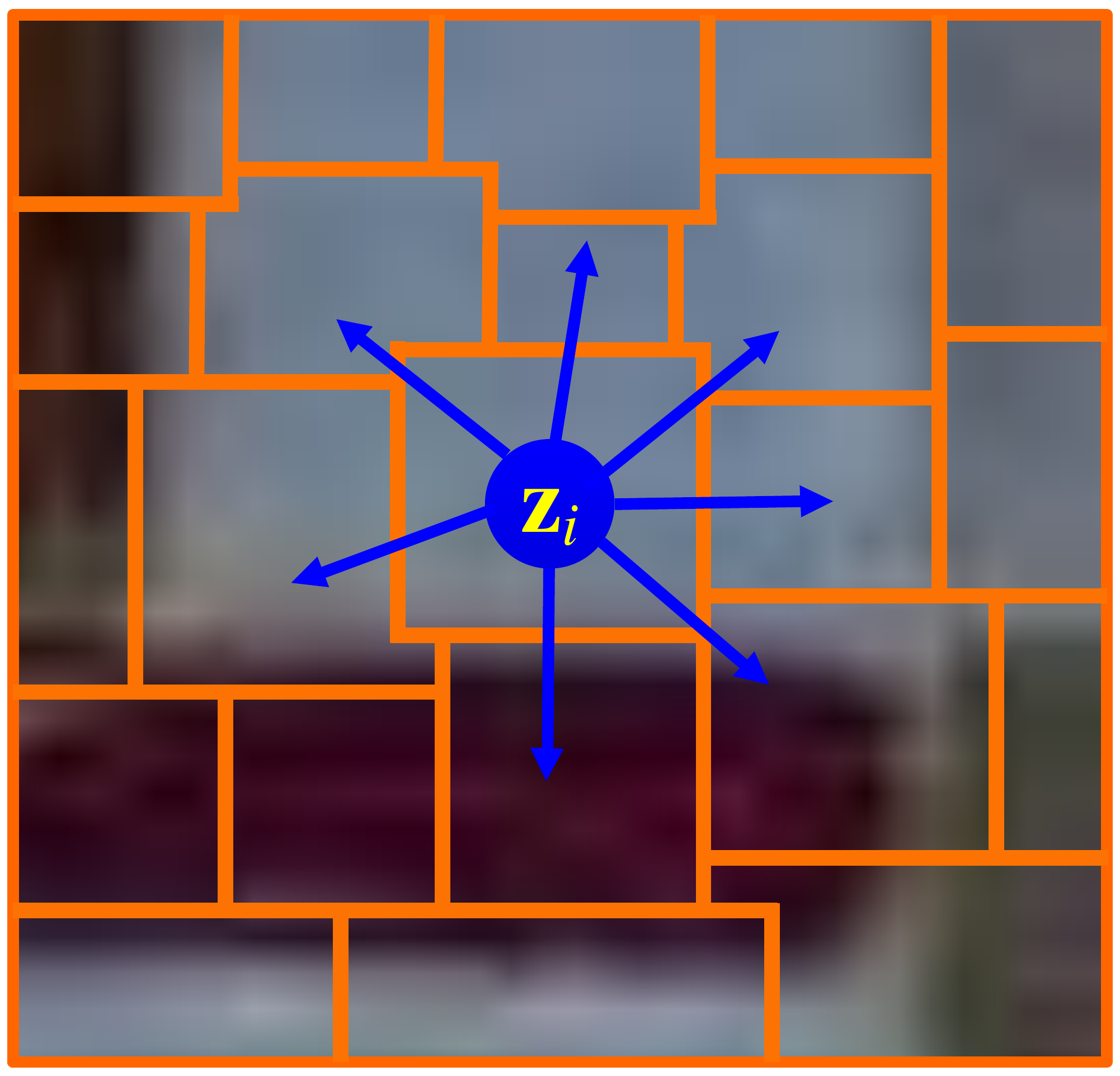}}\hspace{0pt}
		\caption{Illustration of the soft pixel-to-region assignment used in our CAD-GCN model. Each of the initialized image regions is surrounded by yellow lines, and the blue arrows denote the assignment regarding the pixel $\mathbf{z}_i$ to its neighboring regions.}
		\label{SoftAssignment}
	\end{figure}
	
	However, there still exists an optimization challenge, since most or even all of the image pixels may be assigned to a single region in some extreme circumstances. This is probably because that the anchor point matrix for image regions $\mathbf{V}$ is initialized improperly, which will subsequently result in an ill-posed assignment matrix $\mathbf{P}$. Moreover, the imbalanced assignment will lead to unfavorable graph structure, and thus the contextual relations cannot be sufficiently exploited. To cope with this problem, instead of initializing $\mathbf{V}$ randomly, we take the spatial information into consideration and initialize $\mathbf{V}$ by utilizing a segmentation technique. Specifically, the Simple Linear Iterative Clustering (SLIC) algorithm \cite{Radhakrishna2012SLIC}, which has been widely used for image segmentation, is employed to obtain the initial regions. Here the average spectral signatures of the pixels involved in the corresponding region will be utilized to initialize each $\mathbf{v}_i$, and the matrix $\mathbf{V}$ can be further updated via using gradient descent. This segmentation-based initialization technique can yield more stable training performance and produce more meaningful graph representation than random initialization \cite{Yin2018Beyond}. Fig.~\ref{SoftAssignment} exhibits the pixel-to-region assignment regarding a pixel $\mathbf{z}_i$. With the learned region features, the corresponding region graph can be naturally acquired by using Eq.~\eqref{InitialAdjacencyMatirx}. After that, the region features $\mathbf{X}$ will be re-computed by performing graph convolution \cite{Kipf2016Semi} which aggregates information along the edges. Moreover, through successive graph convolutions, long range dependencies among the regions that are far away in the original 2D space can be captured.

	\subsection{Dynamic Graph Refinement}
	\label{DynamicGraph}
	
	The performance of graph convolution largely depends on the quality of the predefined graph which encodes the similarities and connective relationships among graph nodes. However, the Euclidean distance, which is widely used for characterizing node similarities (e.g., in Eq.~\eqref{InitialAdjacencyMatirx}), may not be a good metric for graph structured data \cite{Li2018Adaptive}. In order to address this weakness, we aim to learn an improved distance metric. Specifically , we construct a symmetric positive semi-definite matrix $\mathbf{M} = {\mathbf{W}_d}\mathbf{W}_d^{\top}$ with $\mathbf{W}_d$ being a trainable weight matrix. Then the Generalized Mahalanobis distance can be formulated as follows:
	\begin{equation}
	\label{Dxixj}
	\mathcal{D}({\mathbf{x}_i},{\mathbf{x}_j}) = \sqrt {{{({\mathbf{x}_i} - {\mathbf{x}_j})}^{\top}}\mathbf{M}({\mathbf{x}_i} - {\mathbf{x}_j})}.
	\end{equation}
	Afterwards, the adjacency matrix $\mathbf{A}$ in Eq.~\eqref{InitialAdjacencyMatirx} can be rewritten as
	\begin{equation}
	\label{ADxixj}
	{\mathbf{A}_{ij}} = \left\{ {\begin{array}{*{20}{l}}
		{{e^{ - \gamma ({{\mathcal{D}({\mathbf{x}_i},{\mathbf{x}_j})}})^2}}}\,\;{\rm{if}}\,\;{\mathbf{x}}_{i} \in N(\mathbf{x}_j)\,\;{\rm{or}}\,\;{\mathbf{x}}_{j} \in N(\mathbf{x}_i)\\
		0 \;\;\;\;\;\;\;\;\;\;\;\;\;\;\;\;\;\;\;\;\,\rm{otherwise}
		\end{array}} \right..
	\end{equation}
	Since the graph representation is updated along with the graph convolutional layers, learning a single matrix $\mathbf{M}$ is insufficient to accurately measure node similarities for all the layers. Therefore, we adaptively learn the symmetric positive semi-definite parameter matrix $\mathbf{M}^{(l)}$ for the adjacency matrix $\mathbf{A}^{(l)}$ which is utilized in the $l^{\rm{th}}$ layer, in order to acquire the improved node similarities. Then Eq.~\eqref{ADxixj} can be rewritten as	 
	\begin{equation}
	\label{ADxixj_layer}
	{\mathbf{A}_{ij}^{(l+1)}} = \left\{ {\begin{array}{*{20}{l}}
		{{e^{ - \gamma ({{\mathcal{D}^{(l)}({\mathbf{h}_i^{(l)}},{\mathbf{h}_j^{(l)}})}})^2}}}\,\;\;{\rm{if}}\,\;{\mathbf{x}}_{i} \in N(\mathbf{x}_j)\,\;{\rm{or}}\\
		\;\;\;\;\;\;\;\;\;\;\;\;\;\;\;\;\;\;\;\;\;\;\;\;\;\;\;\;\;\;{\mathbf{x}}_{j} \in N(\mathbf{x}_i)\\
		0 \;\;\;\;\;\;\;\;\;\;\;\;\;\;\;\;\;\;\;\;\;\;\;\;\;\;\;\;\rm{otherwise}
		\end{array}} \right.,	 
	\end{equation}
	where $\mathbf{h}_i^{(l)}$ is the representation of $\mathbf{x}_i$ generated by the $l^{\rm{th}}$ layer with $\mathbf{h}_i^{(0)} = \mathbf{x}_i$, and ${{\mathcal{D}^{(l)}({\mathbf{h}_i^{(l)}},{\mathbf{h}_j^{(l)}})}}$ can be formulated as $\sqrt {{{({\mathbf{h}_i^{(l)}} - {\mathbf{h}_j^{(l)}})}^{\top}}\mathbf{M}^{(l)}({\mathbf{h}_i^{(l)}} - {\mathbf{h}_j^{(l)}})}$.

	During graph construction, connections among the regions from different classes may be incorporated, which will lead to the aggregation of inter-class feature information and further degrade the discriminability of graph convolution results. To overcome this deficiency, we propose to use the edge filter, which aims to refine the contextual relations by reducing undesirable inter-class edges of the graph. Since the intra-class examples are generally more similar than the inter-class ones, it is believed that the element $\mathbf{A}_{ij}^{(l)}$ with relatively small value is more likely to represent inter-class relations than the $\mathbf{A}_{ij}^{(l)}$ with large value. Therefore, we employ a threshold $\beta^{(l)}$ for each graph convolutional layer to filter out the inter-class relations and reduce the adverse effect of inter-class feature aggregation. The selection of $\beta^{(l)}$ will be discussed in Section~\ref{Experiments}. Specifically, in the $l^{\rm{th}}$ layer, the edge filter $\mathcal{F}( \cdot )$ used can be simply expressed as
	\begin{equation}
	\label{EdgeFilter}
	\mathcal{F}(\mathbf{A}_{ij}^{(l)}) = \left\{ {\begin{array}{*{20}{l}}
		{\mathbf{A}_{ij}^{(l)}}\,\;\,\;\,\;\,\;\;\;\;\;\;\;{\rm{if}}\,\;\mathcal{F}(\mathbf{A}_{ij}^{(l)})>\beta^{(l)}\\
		0\;\;\;\;\;\;\;\;\;\;\;\;\;\;\;\;\;\rm{otherwise}
		\end{array}} \right..
	\end{equation}
	In practice, constraining the number of parameters can be beneficial to address the problem of overfitting \cite{Kipf2016Semi}, and thus we set $\beta^{(l)} = \beta$ for all the layers. With the edge filter, the graph convolutional layer can then be reformulated as
	\begin{equation}
	\label{CADGCN_gc_layer}
	{\mathbf{H}^{(l)}} = \sigma (\mathcal{F}(\mathbf{A}^{(l)}){\mathbf{H}^{(l-1)}}{\mathbf{W}^{(l)}}),
	\end{equation}
	with $\mathbf{H}^{(0)} = \mathbf{X}$.

	\subsection{Region-to-Pixel Assignment}
	\label{SecR2P}
	
	After conducting the dynamic graph convolution on the region level, we need to re-project the new region features (i.e., the learned graph representation) $\mathbf{H}^{(L)}$ back into 2D image with grids of pixels, and this process is called graph re-projection, Specifically, the region-to-pixel assignment is accomplished by linearly interpolating pixel features based on the soft assignment matrix $\mathbf{P}$, namely ${\mathbf{P}}{{\mathbf{H}}^{(L)}}$, where $L$ denotes the number of graph convolutional layers. Note that all the re-projected pixels will have diverse feature representations, even if some of them are assigned to the same region. Therefore, the contextual details of the HSI can be well preserved.
	
	With the region-to-pixel assignment, the output of our proposed CAD-GCN can be obtained as
	\begin{equation}
	\label{CADGCN_output}
	{\mathbf{O}} =  \mathbf{P}\mathbf{H}^{(L)}.
	\end{equation}
	In our CAD-GCN model, the cross-entropy error is employed to penalize the differences between the network output and the labels of labeled pixels, namely
	\begin{equation}
	\label{ErrorTerm}
	\mathcal{L} =  - \sum\limits_{g \in {\mathbf{y}_G}} {\sum\limits_{f = 1}^C {{\mathbf{Y}_{gf}}\ln {\mathbf{O}_{gf}}} },
	\end{equation}
	where $C$ is the number of classes, $\mathbf{y}_G$ denotes the set of indices corresponding to the labeled pixels, and $\mathbf{Y}$ represents the label matrix. Here, we let $\mathbf{Y}_{ij}$ be 1 if the pixel $\mathbf{z}_i$ belongs to the $j^{\rm{th}}$ class, and 0 otherwise. It is noticeable that our model can be trained via an end-to-end way. Similar to \cite{Kipf2016Semi}, full-batch gradient descent is utilized to update the network parameters for CAD-GCN. Algorithm~\ref{Algorithm_CADGCN} shows the summarization of our proposed CAD-GCN classification method.

	\begin{algorithm}[!t] 
		\caption{The Proposed CAD-GCN for HSI Classification} 
		\label{Algorithm_CADGCN} 
		\begin{algorithmic}[1] 
			\Require 
			Input image; number of iterations $\mathcal{T}$; learning rate $\eta$; number of graph convolutional layers $L$;
			\State Initialize the anchor point matrix $\mathbf{V}$ with SLIC algorithm;
			\State // Train the CAD-GCN model
			\For {$t=1$ to $\mathcal{T}$}
			\State Learn the region features $\mathbf{X}$ through Eq.~\eqref{AssignmentMatrixP} and Eq.~\eqref{GraphProjection};
			\State Dynamically refine the graph $\mathbf{A}^{(l)}$ using Eq.~\eqref{ADxixj_layer} and Eq.~\eqref{EdgeFilter} along with the graph convolution operation of Eq.~\eqref{CADGCN_gc_layer};
			\State Interpolate the region features back into the original 2D grids by Eq.~\eqref{CADGCN_output};
			\State Calculate the error term according to Eq.~\eqref{ErrorTerm}, and update the weight matrices $\mathbf{W}^{(l)}$ $( 1\leq l\leq L)$ using full-batch gradient descent;
			\EndFor 
			\State \textbf{end for}
			\State Conduct label prediction via Eq.~\eqref{CADGCN_gc_layer} and Eq.~\eqref{CADGCN_output};		
			\Ensure 
			Predicted label for each pixel.
		\end{algorithmic} 
	\end{algorithm}	 
	
	\section{Experimental Results}
	\label{Experiments}
	
	To test the effectiveness of the proposed CAD-GCN model, in this section, we conduct exhaustive experiments on three real-world benchmark datasets, namely Indian Pines, University of Pavia, and Salinas. We first compare CAD-GCN with other state-of-the-art methods, where four metrics including per-class accuracy, overall accuracy (OA), average accuracy (AA), and kappa coefficient, are used to evaluate the model performance. Then we study the influence of the number of labeled pixels on the classification performance. After that, we investigate the impact of hyperparameters incorporated by the proposed CAD-GCN. Finally, we present the ablation study and also investigate the running time of our model.
	
	\subsection{Datasets}
	\label{dataset}
	
	The performance of our proposed CAD-GCN is evaluated on three real-world benchmark datasets, i.e., the Indian Pines\footnote{http://www.ehu.eus/ccwintco/index.php?title=Hyperspectral\_Remote\_Sen\\sing\_Scenes\#Indian\_Pines}, the University of Pavia\footnote{http://www.ehu.eus/ccwintco/index.php?title=Hyperspectral\_Remote\_Sen\\sing\_Scenes\#Pavia\_University\_scene}, and the Salinas\footnote{http://www.ehu.eus/ccwintco/index.php?title=Hyperspectral\_Remote\_Sen\\sing\_Scenes\#Salinas\_scene}, which will be introduced below.
	
	\subsubsection{Indian Pines}
	
\begin{figure}[!t]
		\centering
		
		\subfigure[]{%
			\resizebox*{3cm}{!}{\includegraphics{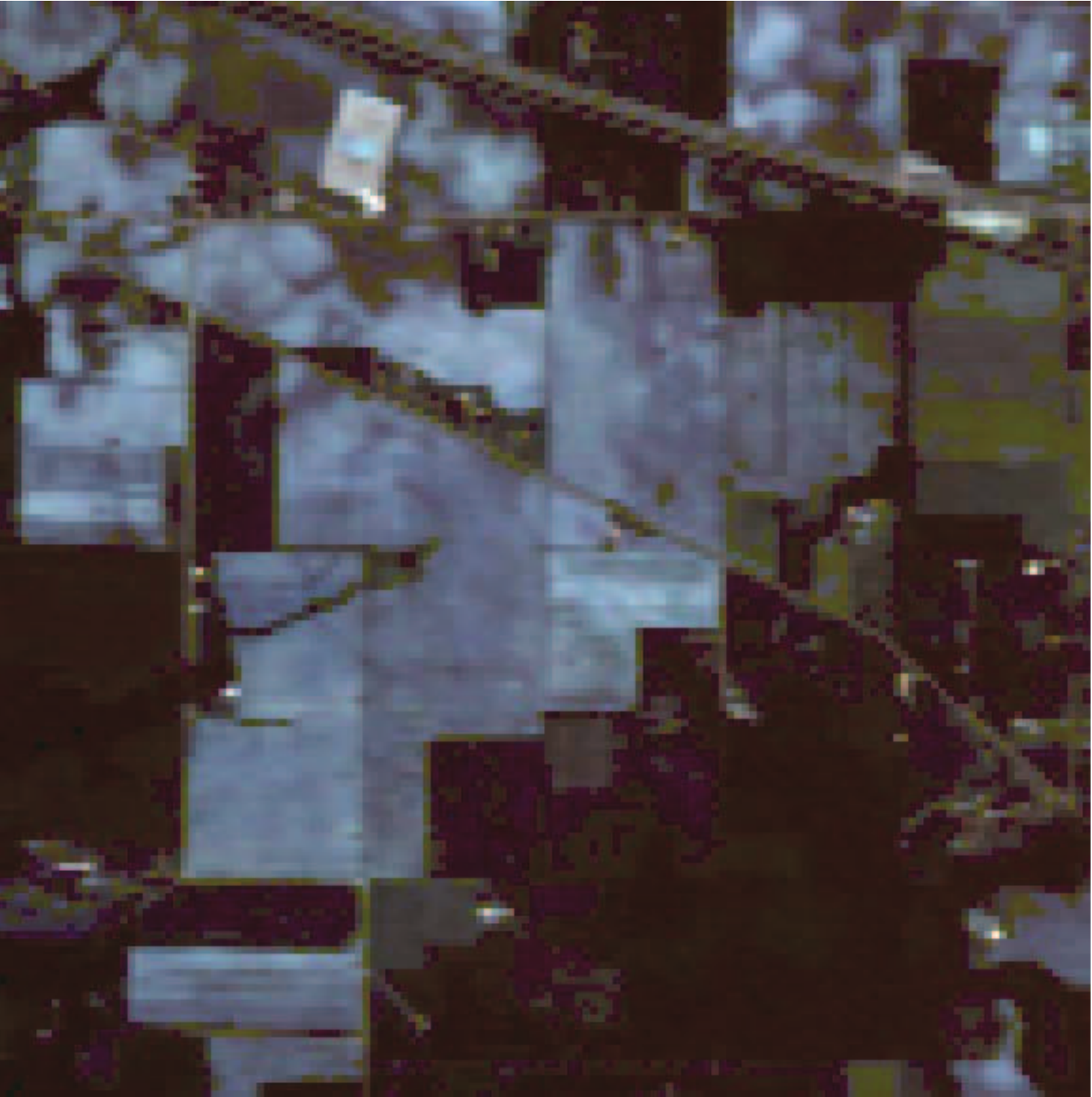}}}\hspace{0pt}	
		\subfigure[]{%
			\label{IPclassmap1}
			\resizebox*{3cm}{!}{\includegraphics{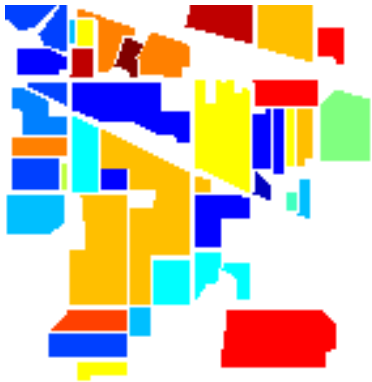}}}\hspace{0pt}
		
		\subfigure {%
			\resizebox*{!}{0.2cm}{\includegraphics{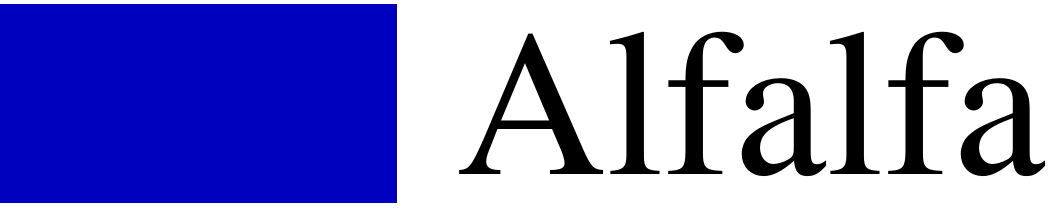}}}\hspace{1pt}
		\subfigure {%
			\resizebox*{!}{0.2cm}{\includegraphics{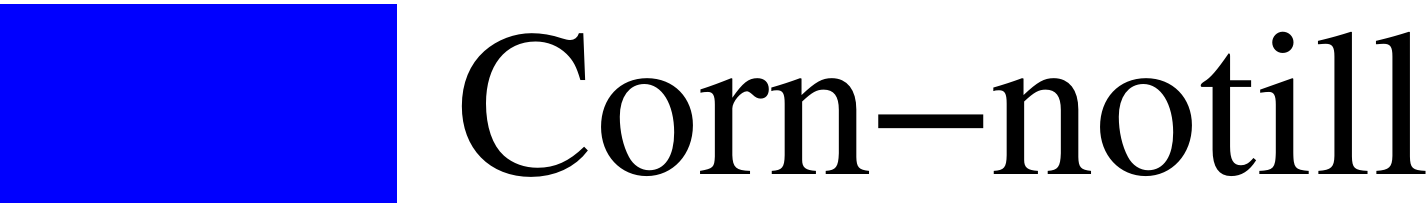}}}\hspace{1pt}
		\subfigure {%
			\resizebox*{!}{0.2cm}{\includegraphics{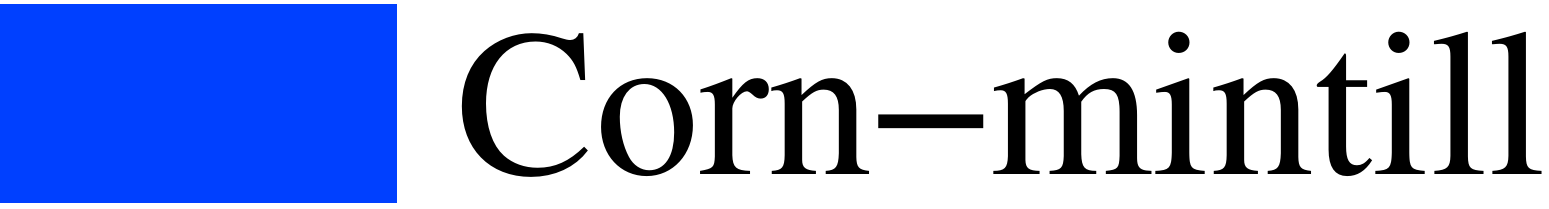}}}\hspace{1pt}
		\subfigure {%
			\resizebox*{!}{0.2cm}{\includegraphics{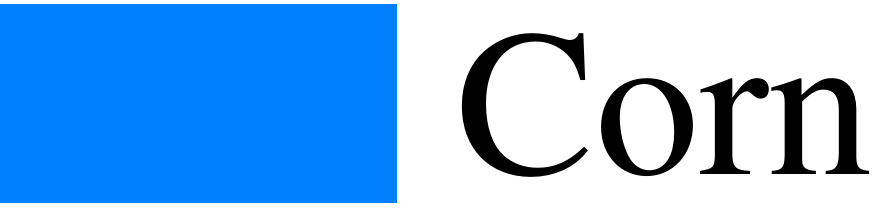}}}\hspace{1pt}
		\subfigure {%
			\resizebox*{!}{0.2cm}{\includegraphics{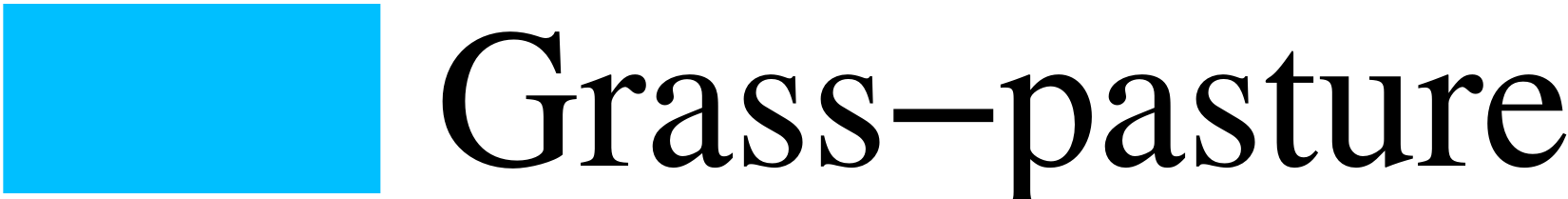}}}\hspace{1pt}
		\subfigure {%
			\resizebox*{!}{0.2cm}{\includegraphics{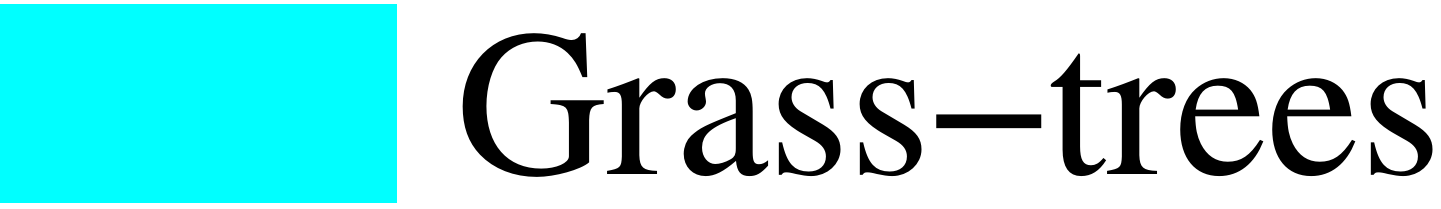}}}\hspace{1pt}
		\subfigure {%
			\resizebox*{!}{0.2cm}{\includegraphics{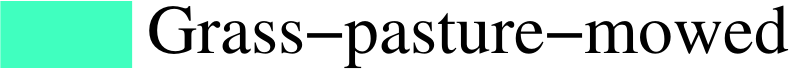}}}\hspace{1pt}
		\subfigure {%
			\resizebox*{!}{0.2cm}{\includegraphics{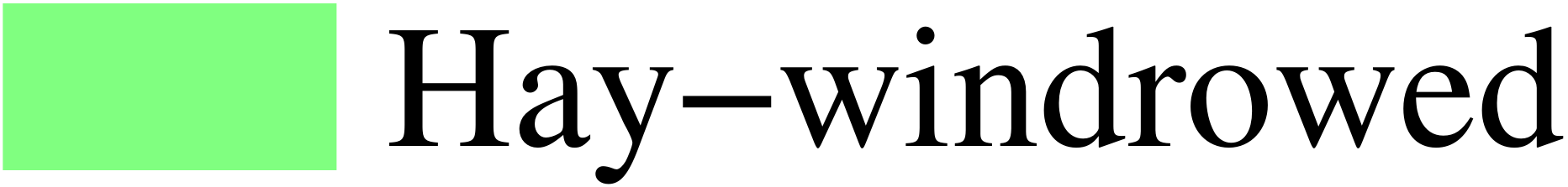}}}\hspace{1pt}
		\subfigure {%
			\resizebox*{!}{0.2cm}{\includegraphics{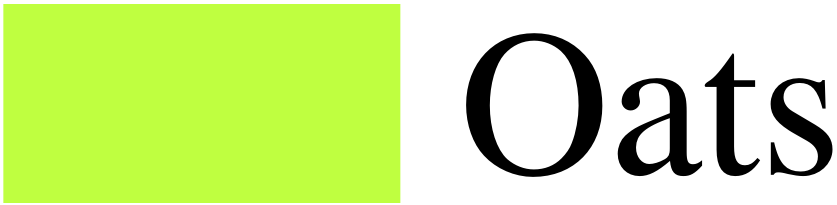}}}\hspace{1pt}
		\subfigure {%
			\resizebox*{!}{0.2cm}{\includegraphics{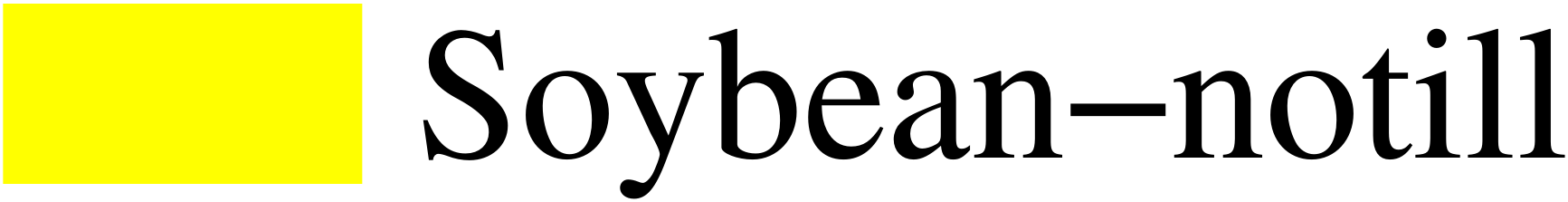}}}\hspace{1pt}
		\subfigure {%
			\resizebox*{!}{0.2cm}{\includegraphics{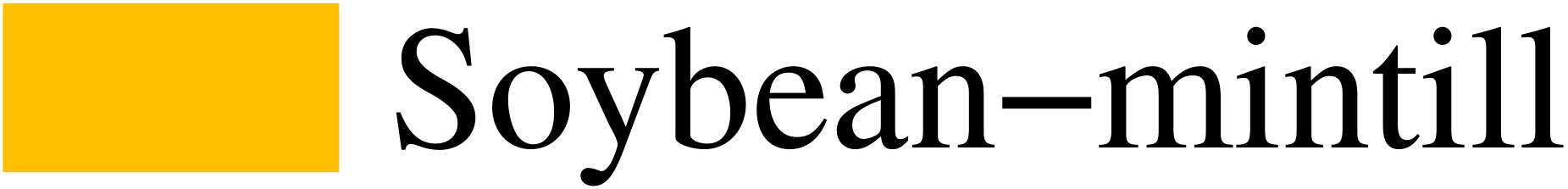}}}\hspace{1pt}
		\subfigure {%
			\resizebox*{!}{0.2cm}{\includegraphics{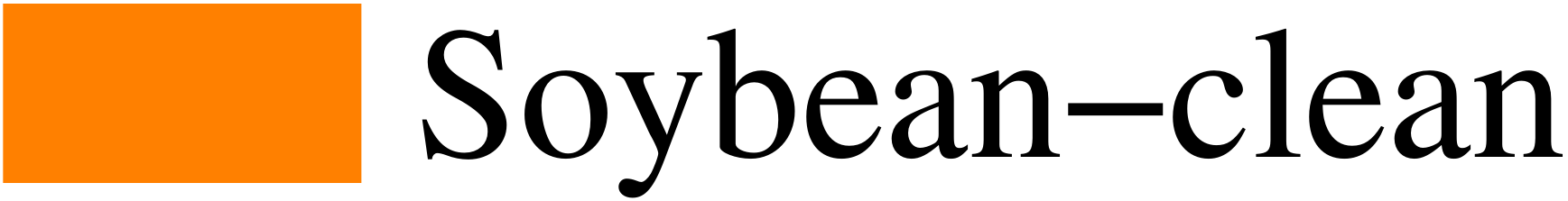}}}\hspace{1pt}
		\subfigure {%
			\resizebox*{!}{0.19cm}{\includegraphics{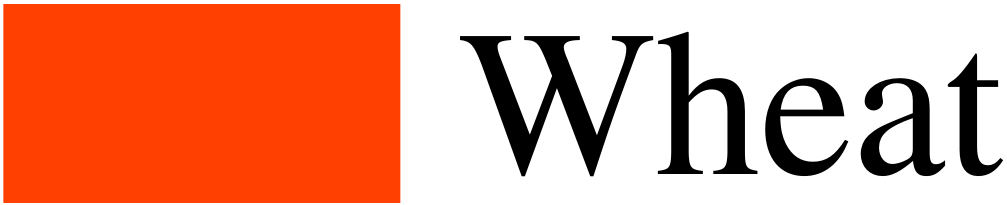}}}\hspace{0pt}
		\subfigure {%
			\resizebox*{!}{0.19cm}{\includegraphics{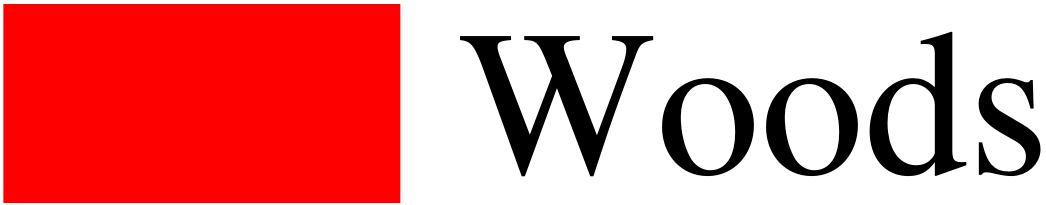}}}\hspace{0pt}
		\subfigure {%
			\resizebox*{!}{0.23cm}{\includegraphics{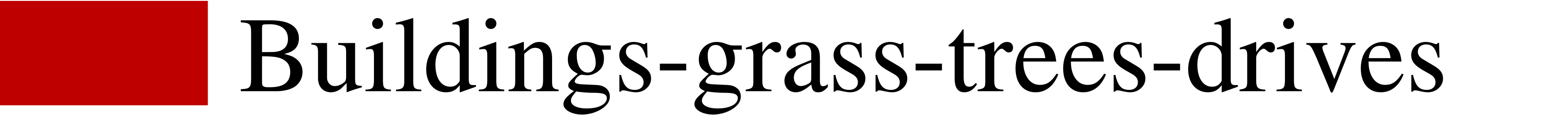}}}\hspace{0pt}
		\subfigure {%
			\resizebox*{!}{0.23cm}{\includegraphics{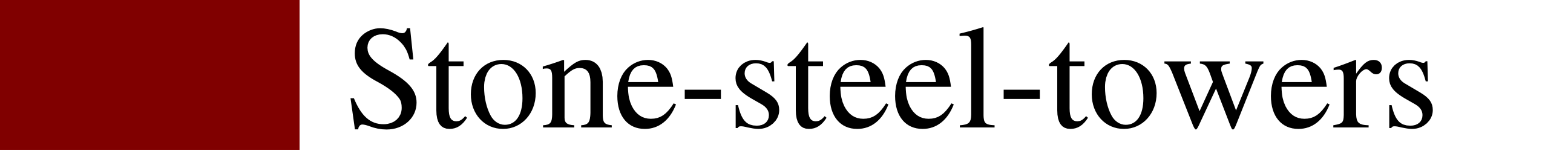}}}\hspace{1pt}
		
		\caption{Indian Pines. (a) False color image. (b) Ground truth map.}
		\label{IPgtfc}
	\end{figure}

	The Indian Pines dataset was gathered by Airborne Visible/Infrared Imaging Spectrometer sensor in 1992, which records north-western India. This dataset consists of $145\times145$ pixels with a spatial resolution of 20 m $\times$ 20 m, and there are 220 spectral channels covering the range from 0.4 $\mu$m to 2.5 $\mu$m. As a usual step, 20 water absorption and noisy bands are removed, and the remaining 200 bands are retained. The original ground truth of the Indian Pines dataset includes 16 land-cover classes, such as `Alfalfa', `Corn-notill', `Corn-mintill', etc. Fig.~\ref{IPgtfc} exhibits the false color image and ground truth map of the Indian Pines dataset. The amounts of labeled and unlabeled pixels of various classes are listed in Table~\ref{IPnum}.
	
	\begin{table}[!t]
		\centering
		\caption{Numbers of Labeled and Unlabeled Pixels of All Classes in Indian Pines Dataset}
		\begin{tabular}{cccc}
			\toprule
			ID   & Class & \#Labeled  & \#Unlabeled \\
			\midrule
			1     & Alfalfa & 30    & 16 \\
			2     & Corn-notill & 30    & 1398 \\
			3     & Corn-mintill & 30    & 800 \\
			4     & Corn  & 30    & 207 \\
			5     & Grass-pasture & 30    & 453 \\
			6     & Grass-trees & 30    & 700 \\
			7     & Grass-pasture-mowed & 15    & 13 \\
			8     & Hay-windrowed & 30    & 448 \\
			9     & Oats  & 15    & 5 \\
			10    & Soybean-notill & 30    & 942 \\
			11    & Soybean-mintill & 30    & 2425 \\
			12    & Soybean-clean & 30    & 563 \\
			13    & Wheat & 30    & 175 \\
			14    & Woods & 30    & 1235 \\
			15    & Buildings-grass-trees-drives & 30    & 356 \\
			16    & Stone-steel-towers & 30    &63 \\
			\bottomrule
		\end{tabular}%
		\label{IPnum}%
	\end{table}%
	
	\subsubsection{University of Pavia}
	
\begin{figure}[!t]
		\centering
		\subfigure[]{%
			\resizebox*{3cm}{!}{\includegraphics{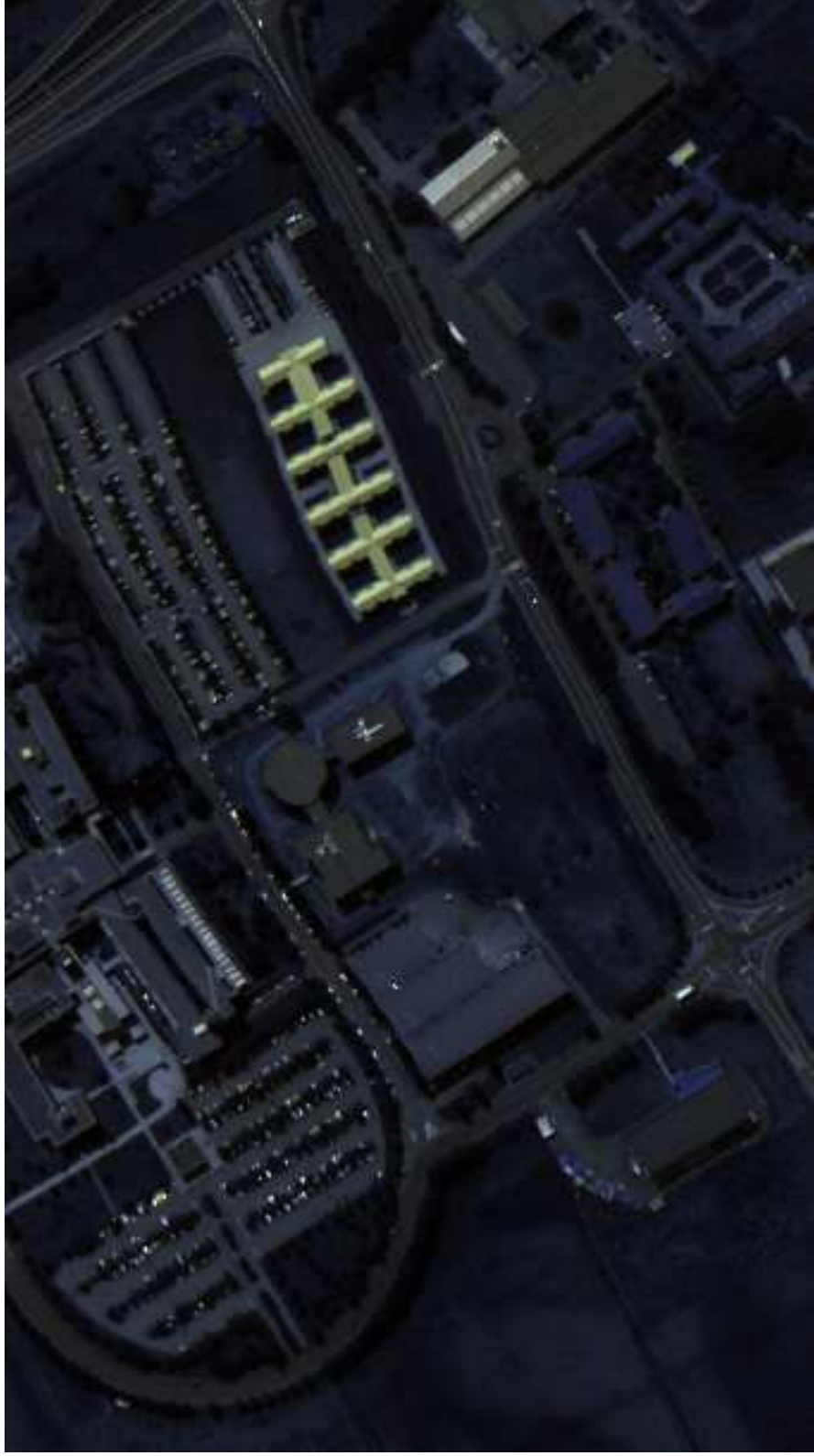}}}\hspace{6pt}
		\subfigure[]{%
			\resizebox*{3cm}{!}{\includegraphics{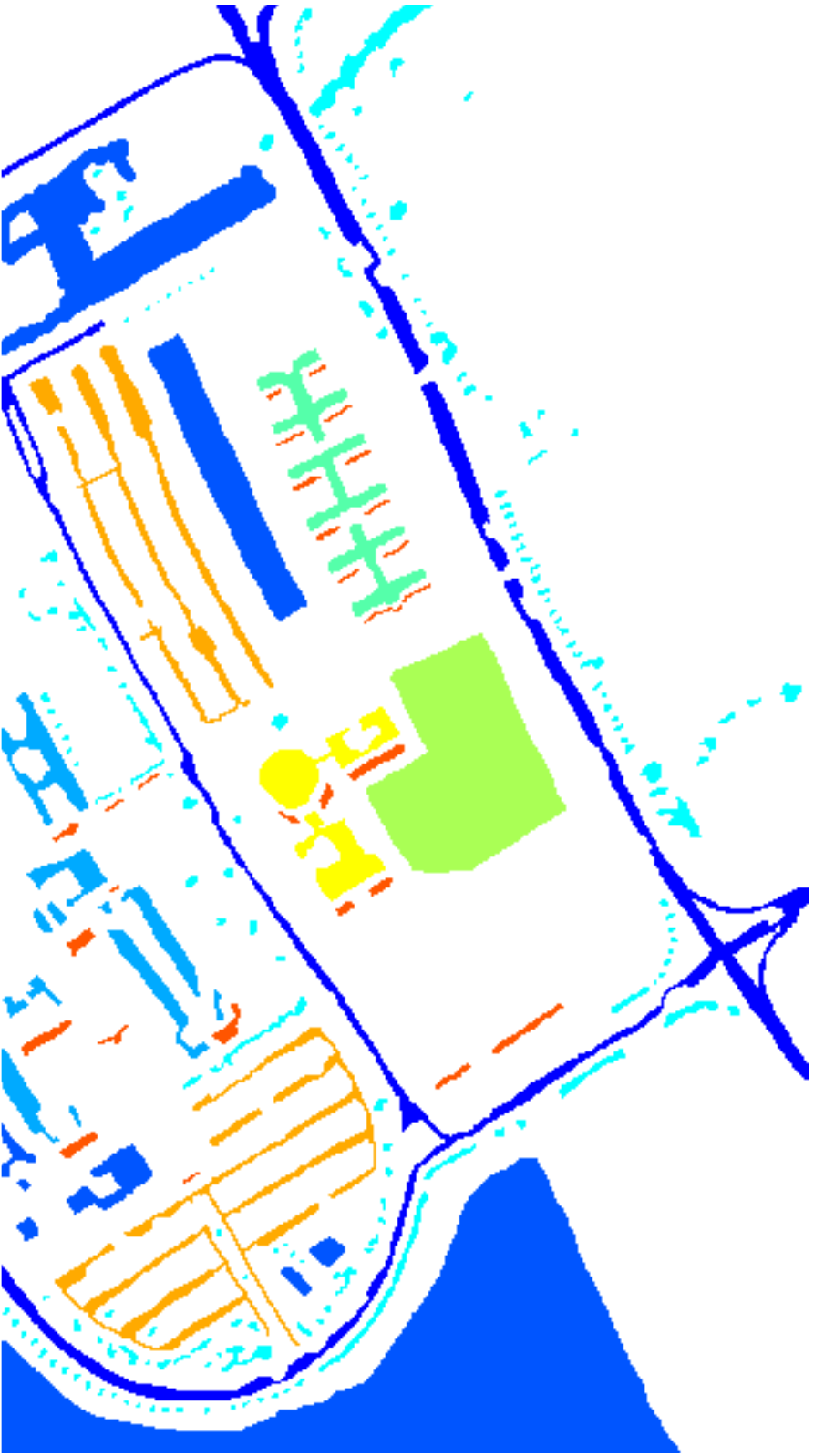}}}\hspace{6pt}
		
		\subfigure {%
			\resizebox*{!}{0.2cm}{\includegraphics{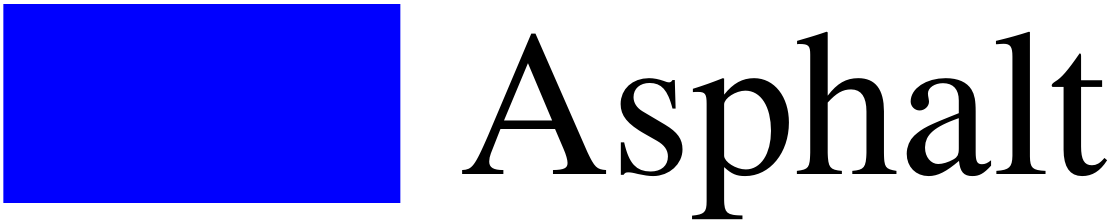}}}\hspace{1pt}
		\subfigure {%
			\resizebox*{!}{0.2cm}{\includegraphics{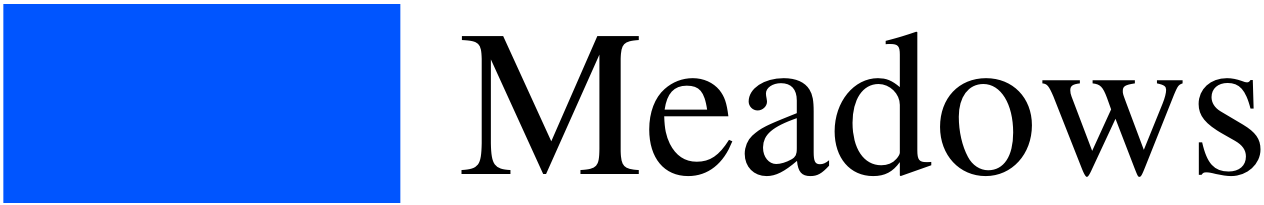}}}\hspace{1pt}
		\subfigure {%
			\resizebox*{!}{0.2cm}{\includegraphics{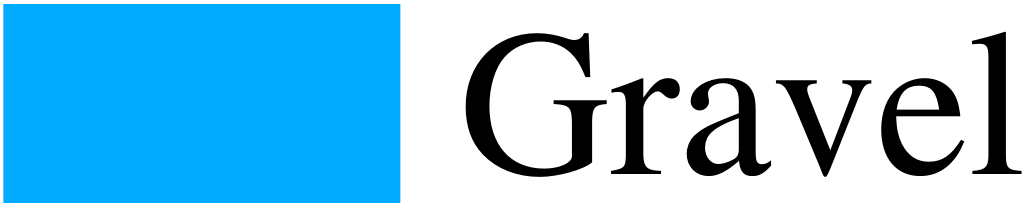}}}\hspace{1pt}
		\subfigure {%
			\resizebox*{!}{0.2cm}{\includegraphics{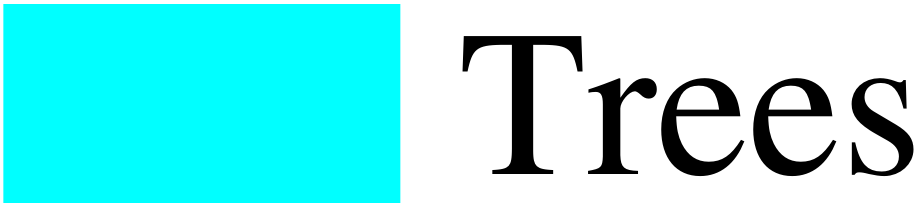}}}\hspace{1pt}
		\subfigure {%
			\resizebox*{!}{0.2cm}{\includegraphics{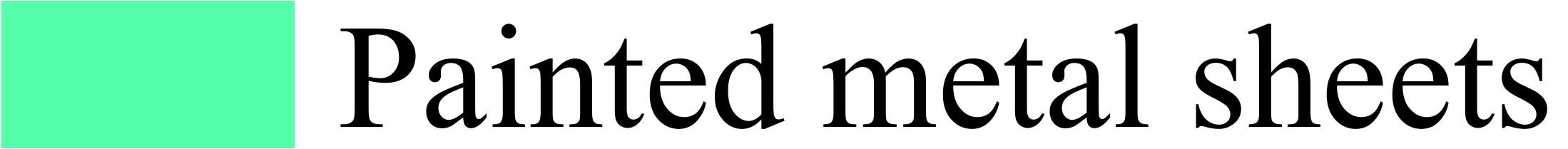}}}\hspace{0pt}
		\subfigure {%
			\resizebox*{!}{0.2cm}{\includegraphics{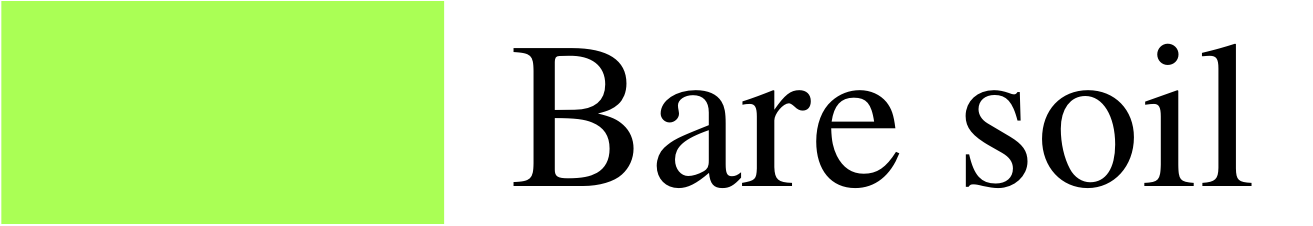}}}\hspace{1pt}	
		\subfigure {%
			\resizebox*{!}{0.2cm}{\includegraphics{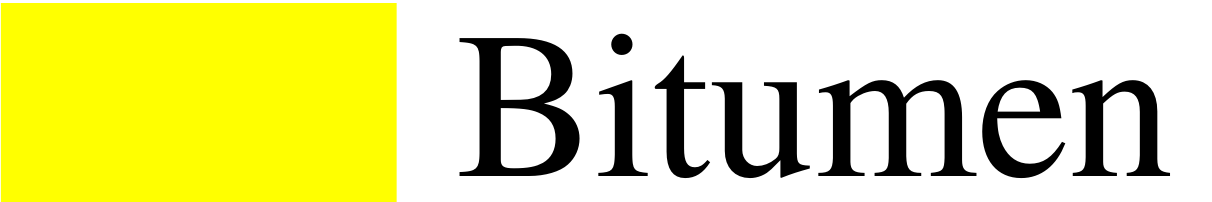}}}\hspace{1pt}
		\subfigure {%
			\resizebox*{!}{0.2cm}{\includegraphics{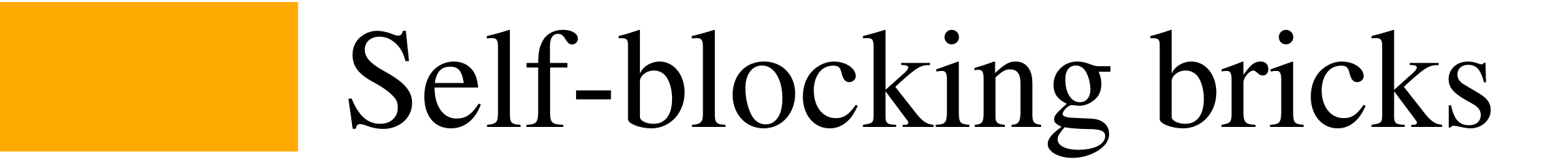}}}\hspace{0pt}
		\subfigure {%
			\resizebox*{!}{0.2cm}{\includegraphics{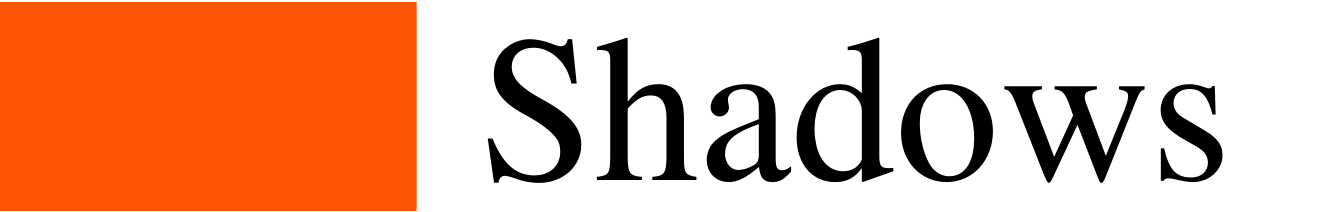}}}\hspace{0pt}	
		\caption{University of Pavia. (a) False color image. (b) Ground truth map.} 
		\label{PUSgtfc}
	\end{figure}

	The University of Pavia dataset captures the Pavia University of Italy with the ROSIS sensor. This dataset consists of $610\times340$ pixels with a spatial resolution of 1.3 m $\times$ 1.3 m and has 103 spectral channels in the wavelength ranging from 0.43 $\mu$m to 0.86 $\mu$m after removing noisy bands. The University of Pavia dataset includes 9 land-cover classes, such as `Asphalt', `Meadows', `Gravel', etc., which are displayed in Fig.~\ref{PUSgtfc}. Table~\ref{PUSnum} shows the amounts of labeled and unlabeled pixels of each class.
	
	\begin{table}[!t]
		\centering
		\caption{Numbers of Labeled and Unlabeled Pixels of All Classes in University of Pavia Dataset}
		\begin{tabular}{cccc}
			\toprule
			ID   & \multicolumn{1}{c}{Class} & \#Labeled & \#Unlabeled \\
			\midrule
			1     & \multicolumn{1}{c}{Asphalt} & 30    & 6601 \\
			2     & Meadows & 30    & 18619 \\
			3     & Gravel & 30    & 2069 \\
			4     & Trees & 30    & 3034 \\
			5     & Painted metal sheets & 30    & 1315 \\
			6     & Bare soil & 30    & 4999 \\
			7     & Bitumen & 30    & 1300 \\
			8     & Self-blocking bricks & 30    & 3652 \\
			9     & Shadows & 30    & 917 \\
			\bottomrule
		\end{tabular}%
		\label{PUSnum}%
	\end{table}%
	
	\subsubsection{Salinas}
	
\begin{figure}[!t]
		\centering
		\subfigure[]{%
			\resizebox*{2.6cm}{!}{\includegraphics{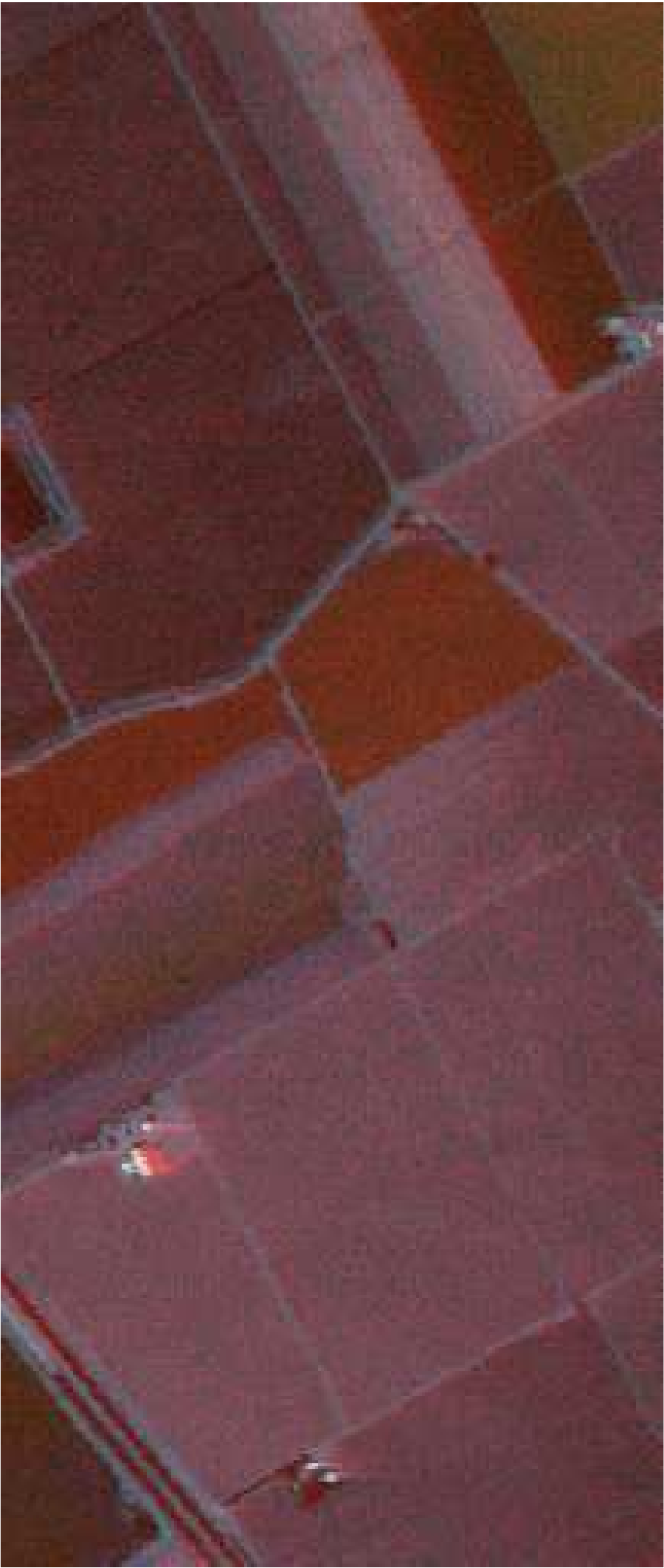}}}\hspace{6pt}
		\subfigure[]{%
			\resizebox*{2.6cm}{!}{\includegraphics{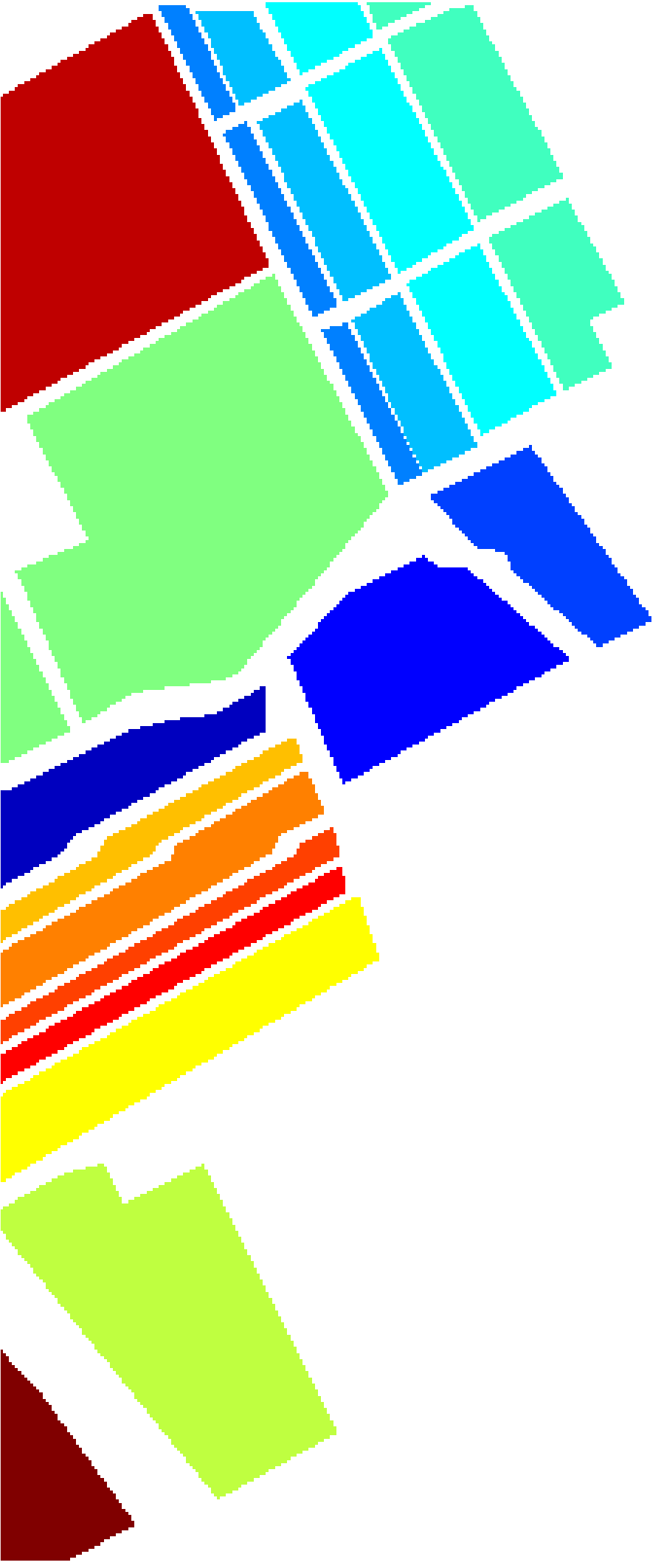}}}\hspace{0pt}
		
		\subfigure {%
			\resizebox*{!}{0.2cm}{\includegraphics{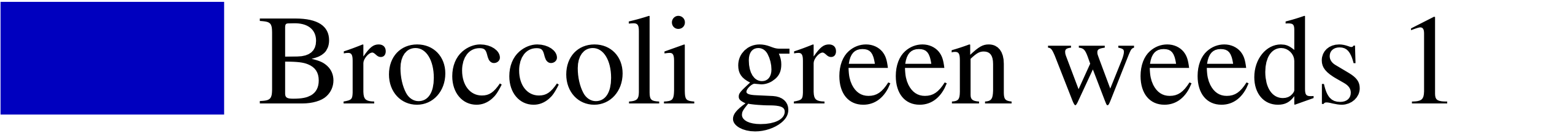}}}\hspace{1pt}
		\subfigure {%
			\resizebox*{!}{0.2cm}{\includegraphics{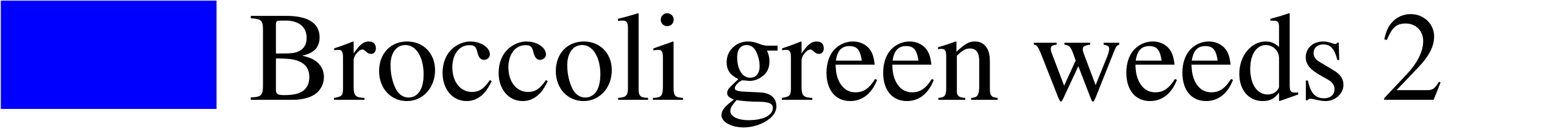}}}\hspace{1pt}
		\subfigure {%
			\resizebox*{!}{0.18cm}{\includegraphics{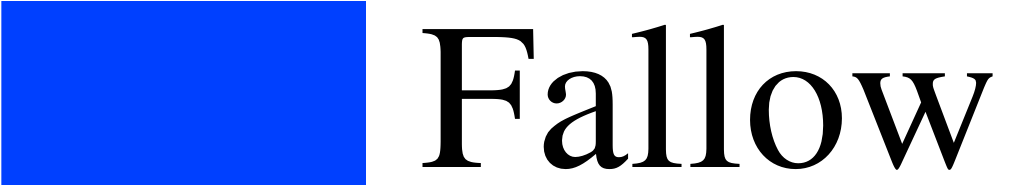}}}\hspace{1pt}
		\subfigure {%
			\resizebox*{!}{0.2cm}{\includegraphics{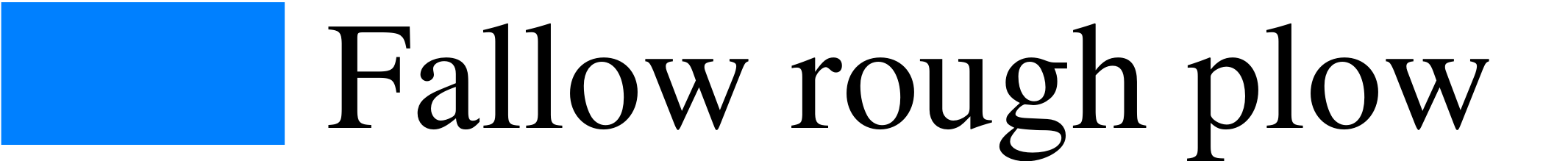}}}\hspace{1pt}
		\subfigure {%
			\resizebox*{!}{0.2cm}{\includegraphics{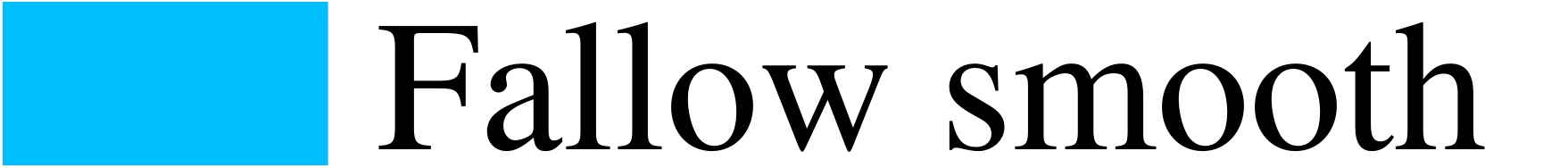}}}\hspace{0pt}
		\subfigure {%
			\resizebox*{!}{0.2cm}{\includegraphics{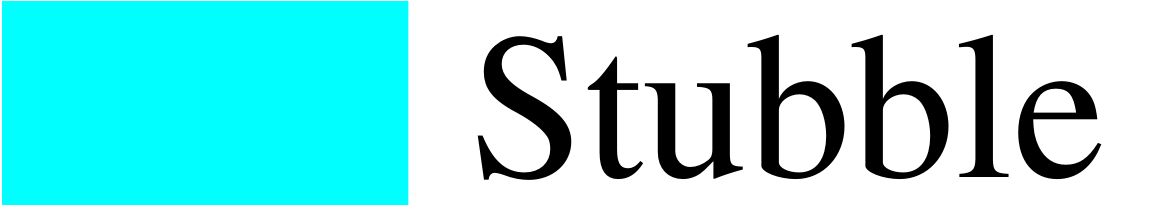}}}\hspace{1pt}	
		\subfigure {%
			\resizebox*{!}{0.205cm}{\includegraphics{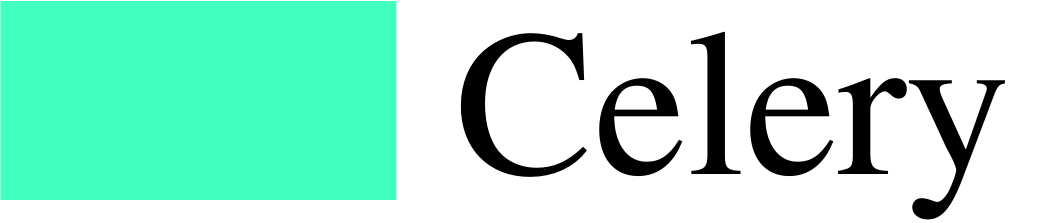}}}\hspace{1pt}
		\subfigure {%
			\resizebox*{!}{0.2cm}{\includegraphics{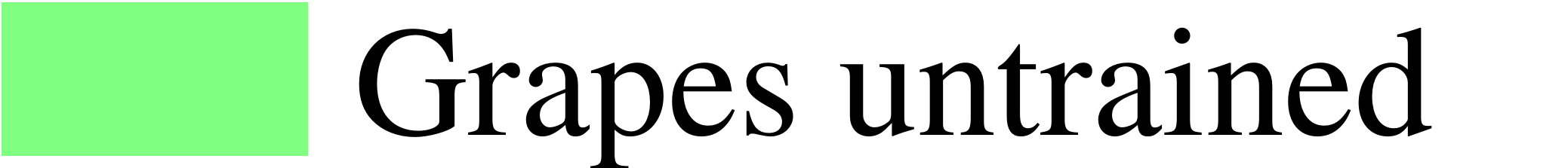}}}\hspace{0pt}
		\subfigure {%
			\resizebox*{!}{0.2cm}{\includegraphics{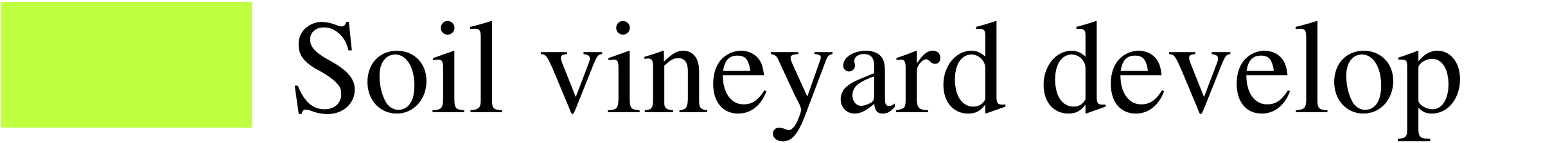}}}\hspace{0pt}	
		\subfigure {%
			\resizebox*{!}{0.2cm}{\includegraphics{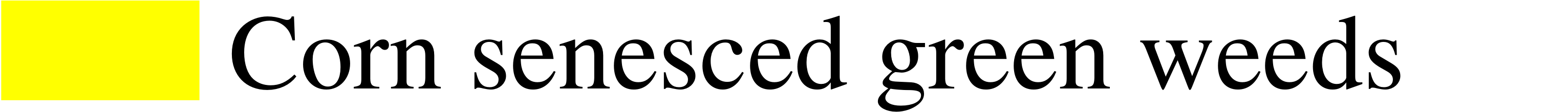}}}\hspace{0pt}
		\subfigure {%
			\resizebox*{!}{0.2cm}{\includegraphics{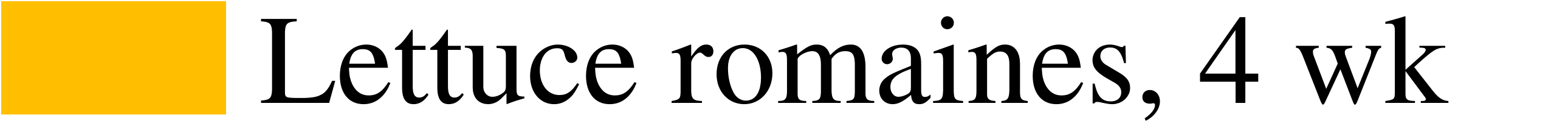}}}\hspace{0pt}
		\subfigure {%
			\resizebox*{!}{0.2cm}{\includegraphics{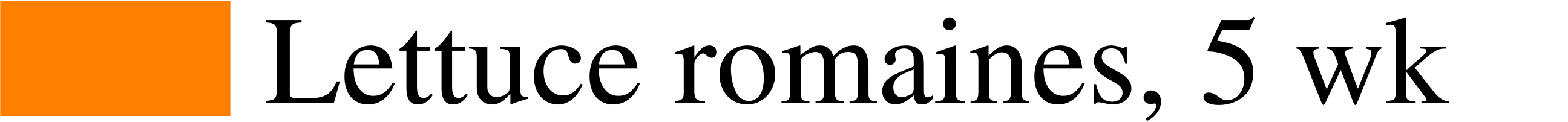}}}\hspace{0pt}		
		\subfigure {%
			\resizebox*{!}{0.2cm}{\includegraphics{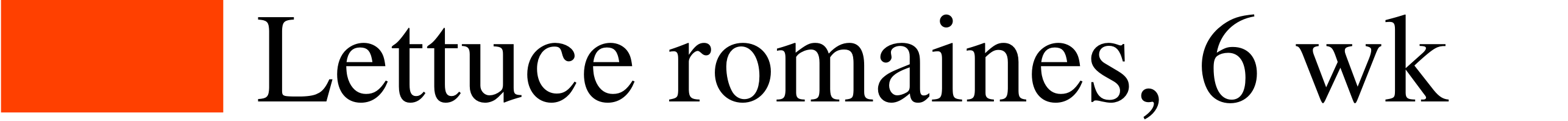}}}\hspace{0pt}	
		\subfigure {%
			\resizebox*{!}{0.2cm}{\includegraphics{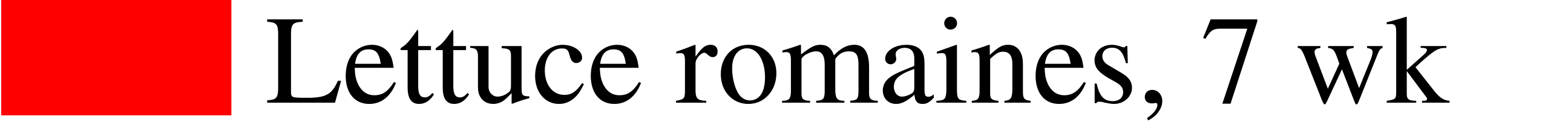}}}\hspace{0pt}
		\subfigure {%
			\resizebox*{!}{0.2cm}{\includegraphics{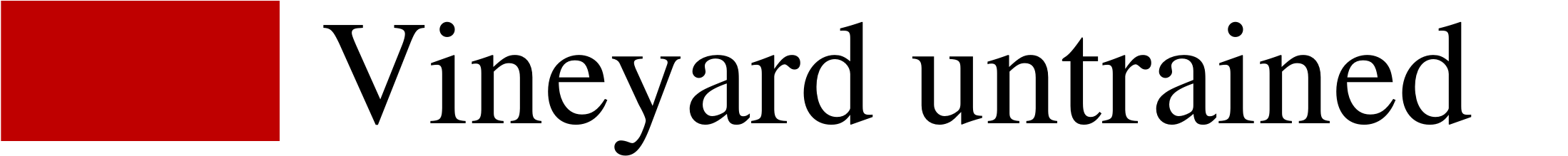}}}\hspace{0pt}		
		\subfigure {%
			\resizebox*{!}{0.2cm}{\includegraphics{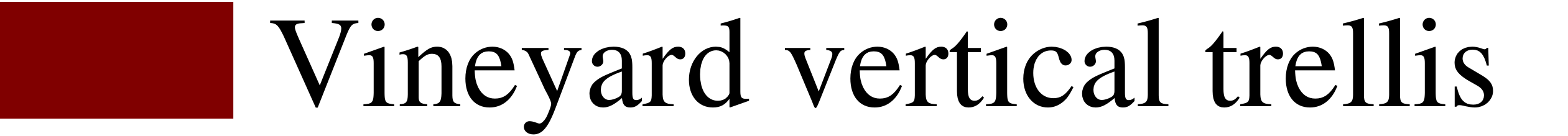}}}\hspace{0pt}

		\caption{Salinas. (a) False color image. (b) Ground truth map.} 
		\label{SAgtfc}
	\end{figure}

	The Salinas dataset is another classic HSI which is collected by the AVIRIS sensor over Salinas Valley, California. This dataset comprises 204 spectral bands (20 water absorption bands are removed) and $512\times 217$ pixels with a spatial resolution of 3.7 m. The Salinas dataset contains 16 land-cover classes, such as `Fallow', `Stubble', `Celery', and so on. Fig.~\ref{SAgtfc} exhibits the false color image and ground truth map of the Salinas dataset. The numbers of labeled and unlabeled pixels of different classes are listed in Table~\ref{SAnum}.

	\begin{table}[!t]
		\centering
		\caption{Numbers of Labeled and Unlabeled Pixels of All Classes in Salinas Dataset}
		\begin{tabular}{cccc}
			\toprule
			ID   & Class & \#Labeled  & \#Unlabeled \\
			\midrule
			1     & Broccoli green weeds 1  & 30    & 1979 \\
			2     & Broccoli green weeds 2 & 30    & 3696 \\
			3     & Fallow & 30    & 1946 \\
			4     & Fallow rough plow & 30    & 1364 \\
			5     & Fallow smooth & 30    & 2648 \\
			6     & Stubble & 30    & 3929 \\
			7     & Celery & 30    & 3549 \\
			8     & Grapes untrained & 30    & 11241 \\
			9     & Soil vineyard develop & 30    & 6173 \\
			10    & Corn senesced green weeds & 30    & 3248 \\
			11    & Lettuce romaines, 4 wk & 30    & 1038 \\
			12    & Lettuce romaines, 5 wk & 30    & 1897 \\
			13    & Lettuce romaines, 6 wk & 30    & 886 \\
			14    & Lettuce romaines, 7 wk & 30    & 1040 \\
			15    & Vineyard untrained & 30    & 7238 \\
			16    & Vineyard vertical trellis & 30    & 1777 \\

			\bottomrule
		\end{tabular}%
		\label{SAnum}%
	\end{table}%
	
	\begin{table*}[!t]
		\centering
		\caption{Per-Class Accuracy, OA, AA (\%), and Kappa Coefficient Achieved by Different Methods on Indian Pines Dataset}
		\begin{tabular}{cccccccc}
			\toprule
			ID    & GCN \cite{Kipf2016Semi} & S$^{2}$GCN \cite{8474300} & R-2D-CNN \cite{Yang2018Hyperspectral} & CNN-PPF \cite{Li2016Hyperspectral} & MFL \cite{6882821} & JSDF \cite{7360896} & CAD-GCN \\
			\midrule
			1     & 95.00$\pm$2.80 & \textbf{100.00$\pm$0.00} & \textbf{100.00$\pm$0.00} & 95.00$\pm$2.64 & 97.64$\pm$0.88 & \textbf{100.00$\pm$0.00} & \textbf{100.00$\pm$0.00} \\
			2     & 56.71$\pm$4.42 & 84.43$\pm$2.50 & 54.94$\pm$2.23 & 73.53$\pm$5.61 & 67.93$\pm$0.42 & \textbf{90.75$\pm$3.19} & 89.24$\pm$4.11 \\
			3     & 51.50$\pm$2.56 & 82.87$\pm$5.53 & 73.31$\pm$4.33 & 81.34$\pm$3.76 & 71.03$\pm$0.63 & 77.84$\pm$3.81 & \textbf{94.20$\pm$3.26} \\
			4     & 84.64$\pm$3.16 & 93.08$\pm$1.95 & 84.06$\pm$12.98 & 91.84$\pm$3.53 & 85.84$\pm$0.70 & \textbf{99.86$\pm$0.33} & 98.52$\pm$1.15 \\
			5     & 83.71$\pm$3.20 & \textbf{97.13$\pm$1.34} & 87.64$\pm$0.31 & 93.69$\pm$0.84 & 89.36$\pm$0.48 & 87.20$\pm$2.73 & 95.12$\pm$1.89 \\
			6     & 94.03$\pm$2.11 & 97.29$\pm$1.27 & 91.21$\pm$4.34 & 97.46$\pm$1.01 & 97.66$\pm$0.27 & \textbf{98.54$\pm$0.28} & 95.42$\pm$5.02 \\
			7     & 92.31$\pm$0.00 & 92.31$\pm$0.00 & \textbf{100.00$\pm$0.00} & 75.38$\pm$8.73 & 95.06$\pm$0.79 & \textbf{100.00$\pm$0.00} & 98.65$\pm$2.68 \\
			8     & 96.61$\pm$1.86 & 99.03$\pm$0.93 & 99.11$\pm$0.95 & 98.01$\pm$0.69 & 99.62$\pm$0.05 & \textbf{99.80$\pm$0.31} & 99.68$\pm$0.54 \\
			9     & \textbf{100.00$\pm$0.00} & \textbf{100.00$\pm$0.00} & \textbf{100.00$\pm$0.00} & \textbf{100.00$\pm$0.00} & 98.00$\pm$0.94 & \textbf{100.00$\pm$0.00} & \textbf{100.00$\pm$0.00} \\
			10    & 77.47$\pm$1.24 & \textbf{93.77$\pm$3.72} & 70.81$\pm$5.11 & 82.30$\pm$1.55 & 76.41$\pm$0.64 & 89.99$\pm$4.24 & 89.09$\pm$4.19 \\
			11    & 56.56$\pm$1.53 & 84.98$\pm$2.82 & 56.35$\pm$1.08 & 62.64$\pm$3.32 & 73.78$\pm$0.59 & 76.75$\pm$5.12 & \textbf{92.18$\pm$2.00} \\
			12    & 58.29$\pm$6.58 & 80.05$\pm$5.17 & 63.06$\pm$12.81 & 88.92$\pm$2.50 & 70.92$\pm$0.80 & 87.10$\pm$2.82 & \textbf{96.03$\pm$2.65} \\
			13    & \textbf{100.00$\pm$0.00} & 99.43$\pm$0.00 & 98.86$\pm$1.62 & 98.80$\pm$0.57 & 98.80$\pm$0.08 & 99.89$\pm$0.36 & 99.75$\pm$0.29 \\
			14    & 80.03$\pm$3.93 & 96.73$\pm$0.92 & 88.74$\pm$2.58 & 86.49$\pm$2.23 & 90.12$\pm$0.53 & 97.21$\pm$2.78 & \textbf{99.36$\pm$0.69} \\
			15    & 69.55$\pm$6.66 & 86.80$\pm$3.42 & 87.08$\pm$2.78 & 86.71$\pm$4.36 & 96.05$\pm$0.35 & \textbf{99.58$\pm$0.68} & 99.04$\pm$1.12 \\
			16    & 98.41$\pm$0.00 & \textbf{100.00$\pm$0.00} & 97.62$\pm$1.12 & 92.70$\pm$3.45 & 97.54$\pm$0.23 & \textbf{100.00$\pm$0.00} & 98.18$\pm$3.89 \\
			\midrule
			OA    & 69.24$\pm$1.56 & 89.49$\pm$1.08 & 72.11$\pm$1.28 & 80.09$\pm$1.56 & 80.22$\pm$0.20 & 88.34$\pm$1.39 & \textbf{94.06$\pm$0.85} \\
			AA    & 80.93$\pm$1.71 & 92.99$\pm$1.04 & 84.55$\pm$1.79 & 87.80$\pm$1.53 & 87.85$\pm$0.19 & 94.03$\pm$0.55 & \textbf{96.53$\pm$0.55} \\
			Kappa & 65.27$\pm$1.80 & 88.00$\pm$1.23 & 68.66$\pm$1.46 & 77.52$\pm$1.74 & 77.59$\pm$0.22 & 86.80$\pm$1.55 & \textbf{93.22$\pm$0.97} \\
			\bottomrule
		\end{tabular}%
		\label{IPClassificationResults}%
	\end{table*}%
	
	\subsection{Experimental Settings}
	
	In our experiments, the proposed CAD-GCN algorithm is implemented by TensorFlow with Adam optimizer. For all the adopted three datasets mentioned in Section \ref{dataset}, usually 30 labeled pixels (i.e., examples) per class are randomly chosen for training, and 15 labeled pixels are chosen if the corresponding class contains less than 30 pixels. During training, 90\% of the labeled examples are used to learn the network parameters and 10\% are used as validation set to tune the hyperparameters. Meanwhile, all the unlabeled examples are used as the test set to evaluate the classification performance.

	To evaluate the classification ability of our proposed CAD-GCN, other recent state-of-the-art HSI classification methods are also utilized for comparison. Specifically, we employ two GCN-based methods, i.e., Graph Convolutional Network (GCN) \cite{Kipf2016Semi} and Spectral-Spatial Graph Convolutional Network (S$^2$GCN) \cite{8474300}, together with two CNN-based methods, i.e., Recurrent 2D-CNN (R-2D-CNN) \cite{Yang2018Hyperspectral} and CNN-Pixel-Pair Features (CNN-PPF) \cite{Li2016Hyperspectral}. Meanwhile, we also compare the proposed CAD-GCN with two traditional HSI classification methods, namely Joint collaborative representation and SVM with Decision Fusion (JSDF) \cite{7360896} and Multiple Feature Learning (MFL) \cite{6882821}, respectively. All these methods are implemented ten times with different labeled pixels on each hyperspectral dataset, and the mean accuracies together with the standard deviations over these ten independent implementations are reported.

	\begin{figure*}[!t]
		\centering
		\subfigure[]{%
			\label{IPClassificationMaps_gt}
			\resizebox*{3.15cm}{!}{\includegraphics{IPgt.pdf}}}\hspace{18pt}
		\subfigure[]{%
			\label{IP_GCN}
			\resizebox*{3.2cm}{!}{\includegraphics{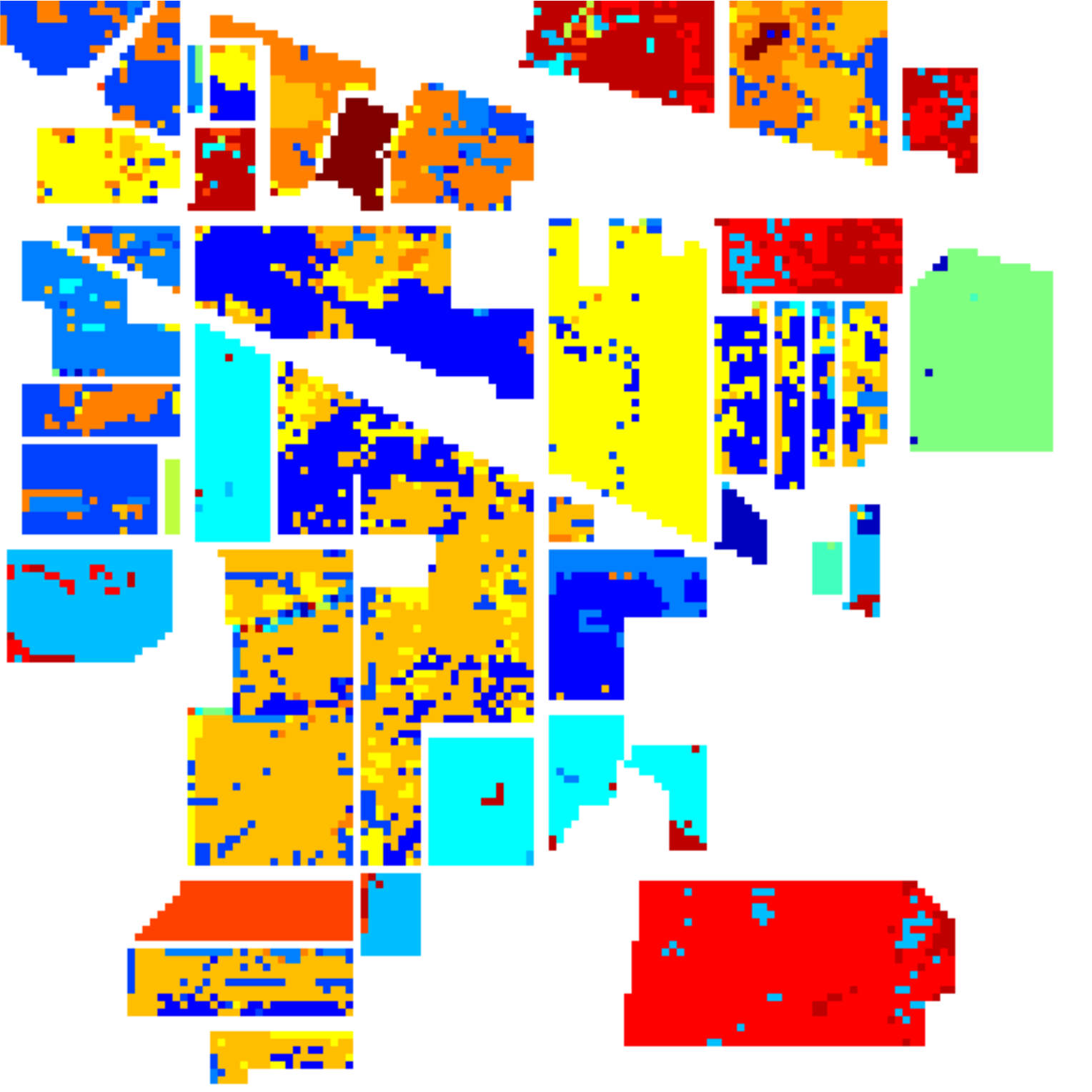}}}\hspace{15pt}	
		\subfigure[]{%
			\label{IP_S2GCN}
			\resizebox*{3.2cm}{!}{\includegraphics{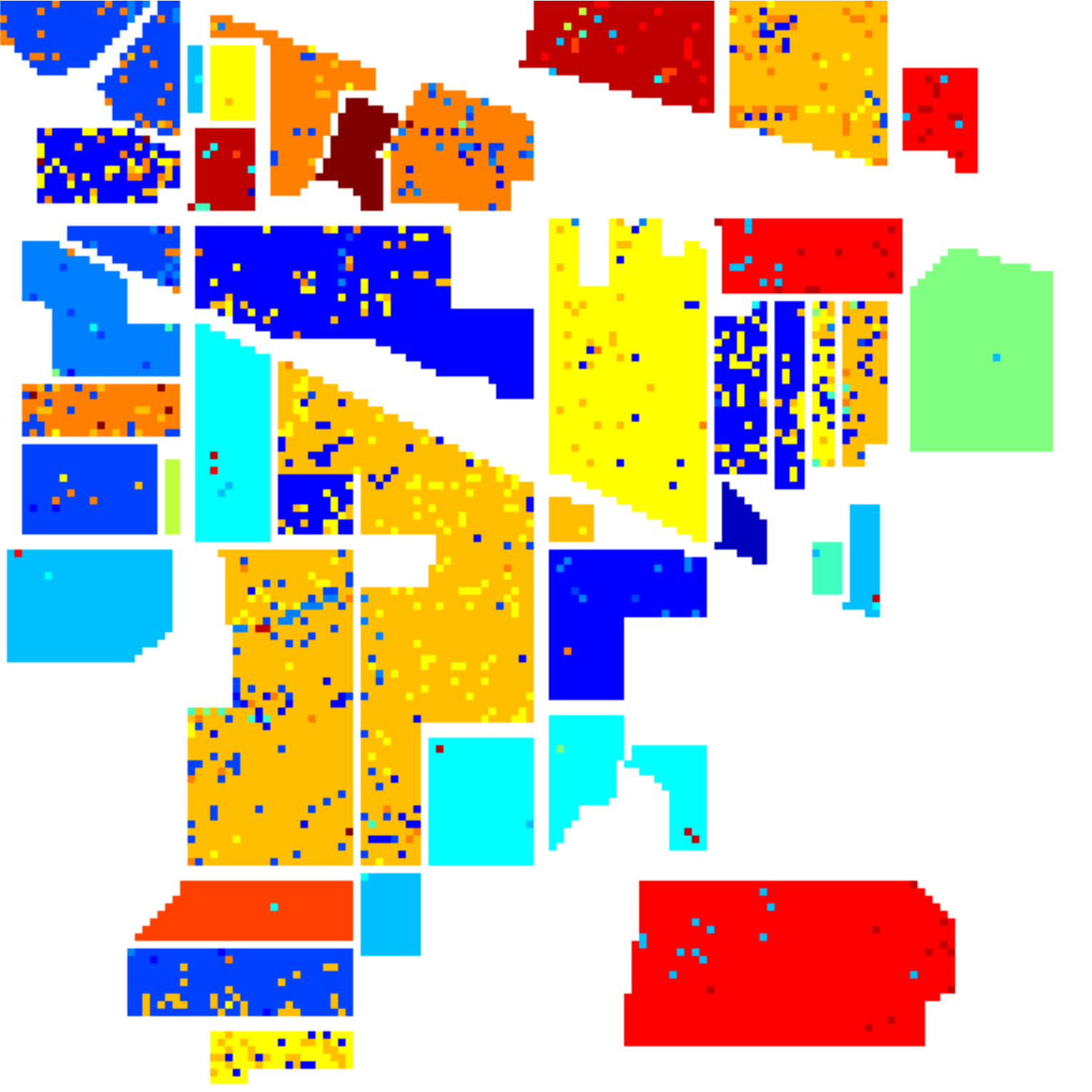}}}\hspace{15pt}	
		\subfigure[]{%
			\label{IP_R2DCNN}
			\resizebox*{3.2cm}{!}{\includegraphics{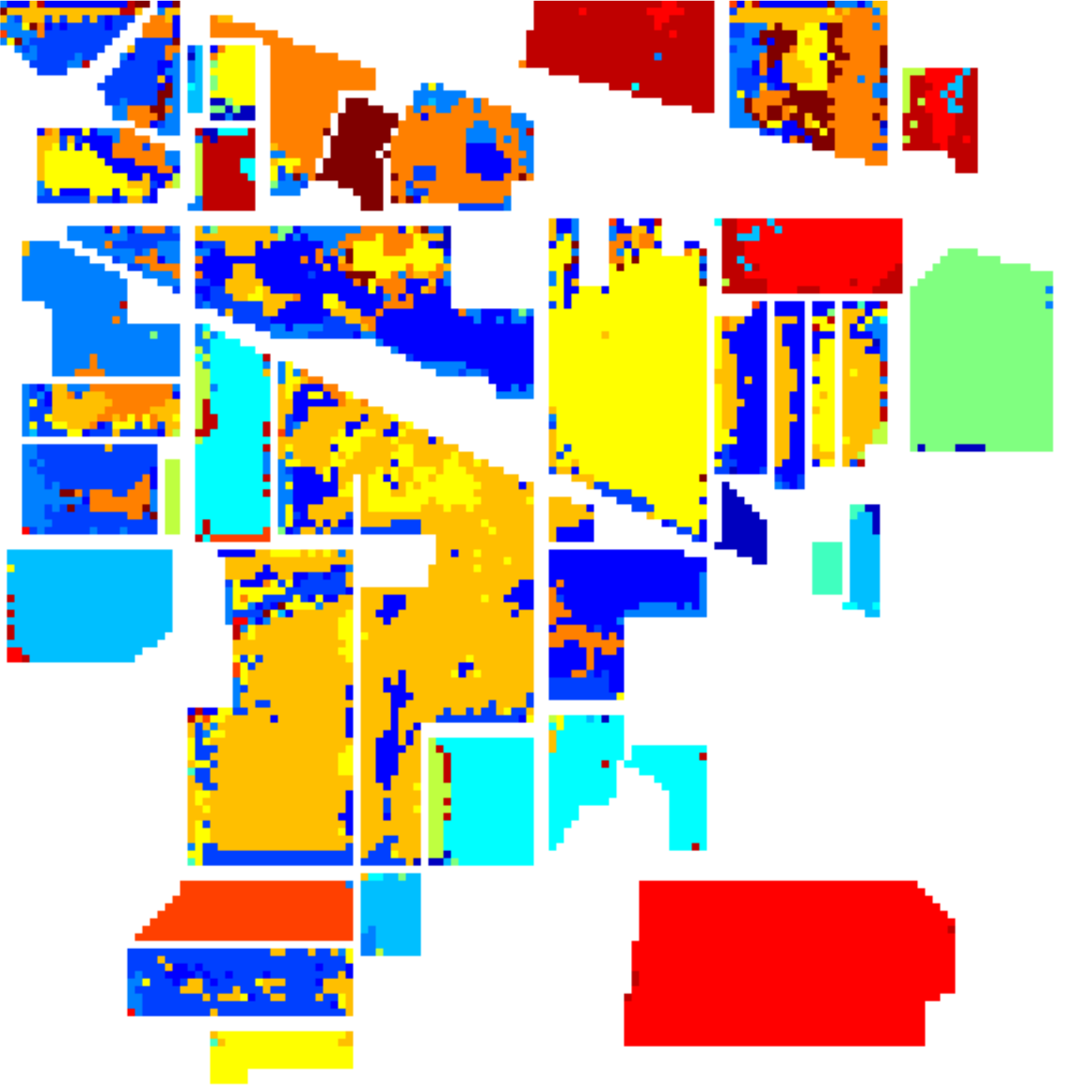}}}\hspace{28pt}	
		\subfigure[]{%
			\label{IP_CNN_PPF}
			\resizebox*{3.2cm}{!}{\includegraphics{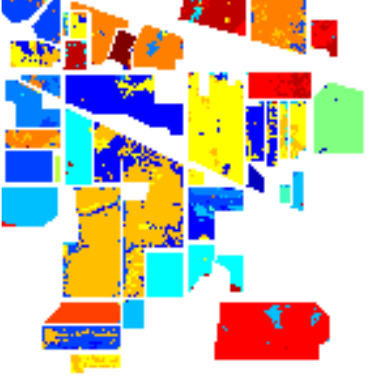}}}\hspace{15pt}	
		\subfigure[]{%
			\label{IP_NMFL}
			\resizebox*{3.2cm}{!}{\includegraphics{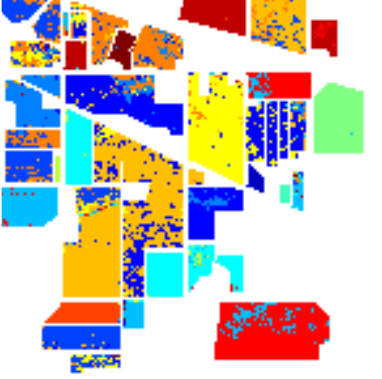}}}\hspace{15pt}		
		\subfigure[]{%
			\label{IP_JSDF}
			\resizebox*{3.2cm}{!}{\includegraphics{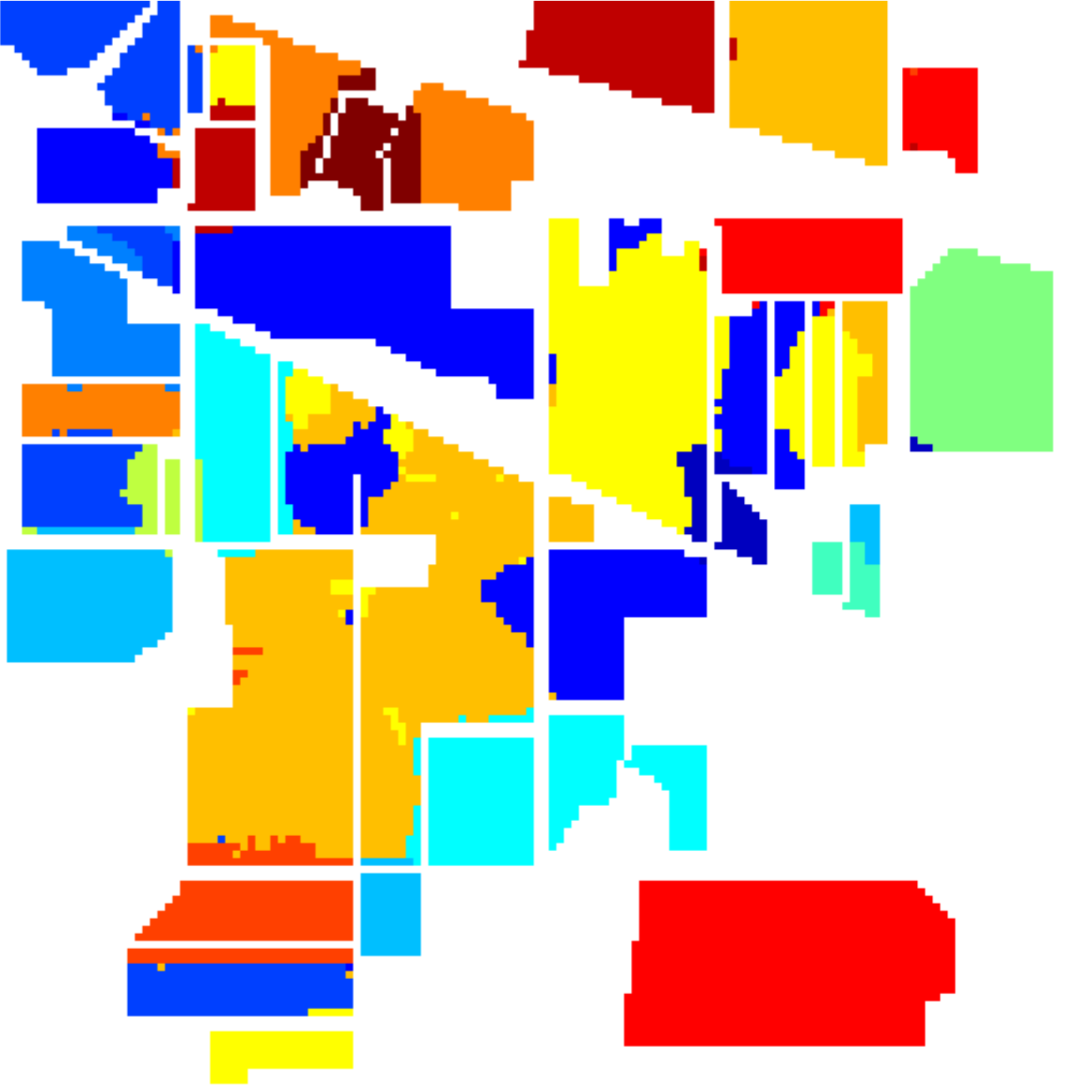}}}\hspace{15pt}	
		\subfigure[]{%
			\label{IP_CAD_GCN}
			\resizebox*{3.2cm}{!}{\includegraphics{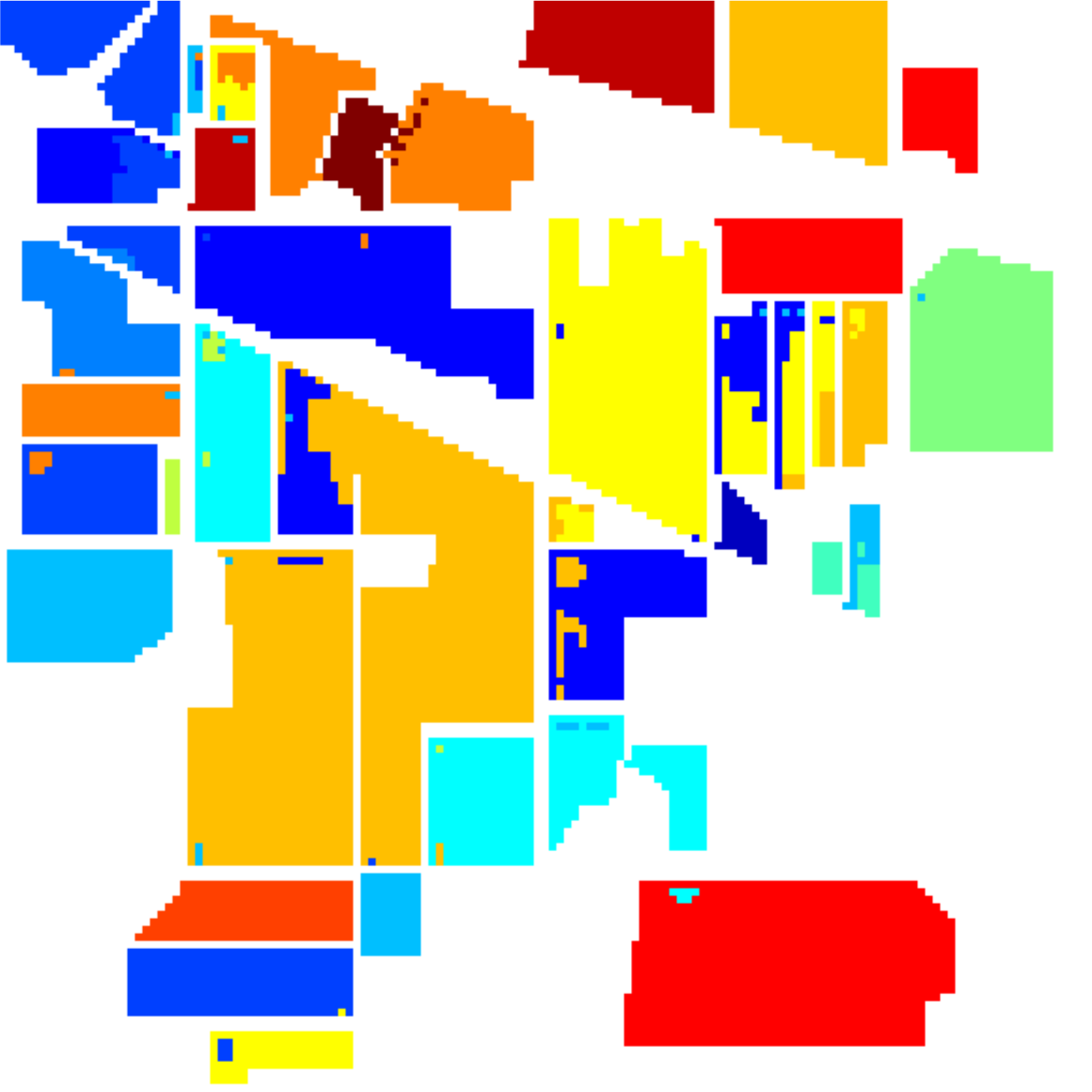}}}\hspace{0pt}
		
		\subfigure {%
			\resizebox*{!}{0.2cm}{\includegraphics{IPclass1.pdf}}}\hspace{9pt}
		\subfigure {%
			\resizebox*{!}{0.2cm}{\includegraphics{IPclass2.pdf}}}\hspace{9pt}
		\subfigure {%
			\resizebox*{!}{0.2cm}{\includegraphics{IPclass3.pdf}}}\hspace{9pt}
		\subfigure {%
			\resizebox*{!}{0.2cm}{\includegraphics{IPclass4.pdf}}}\hspace{9pt}
		\subfigure {%
			\resizebox*{!}{0.2cm}{\includegraphics{IPclass5.pdf}}}\hspace{9pt}
		\subfigure {%
			\resizebox*{!}{0.2cm}{\includegraphics{IPclass6.pdf}}}\hspace{9pt}
		\subfigure {%
			\resizebox*{!}{0.2cm}{\includegraphics{IPclass7.pdf}}}\hspace{9pt}
		\subfigure {%
			\resizebox*{!}{0.2cm}{\includegraphics{IPclass8.pdf}}}\hspace{9pt}
		
		\subfigure {%
			\resizebox*{!}{0.2cm}{\includegraphics{IPclass9.pdf}}}\hspace{9pt}
		\subfigure {%
			\resizebox*{!}{0.2cm}{\includegraphics{IPclass10.pdf}}}\hspace{9pt}
		\subfigure {%
			\resizebox*{!}{0.2cm}{\includegraphics{IPclass11.pdf}}}\hspace{9pt}
		\subfigure {%
			\resizebox*{!}{0.2cm}{\includegraphics{IPclass12.pdf}}}\hspace{9pt}
		\subfigure {%
			\resizebox*{!}{0.19cm}{\includegraphics{IPclass13.pdf}}}\hspace{9pt}
		\subfigure {%
			\resizebox*{!}{0.19cm}{\includegraphics{IPclass14.pdf}}}\hspace{9pt}
		\subfigure {%
			\resizebox*{!}{0.23cm}{\includegraphics{IPclass15.pdf}}}\hspace{4pt}
		\subfigure {%
			\resizebox*{!}{0.23cm}{\includegraphics{IPclass16.pdf}}}\hspace{0pt}
		\caption{Classification maps obtained by different methods on Indian Pines dataset. (a) Ground truth map; (b) GCN; (c) S$^{2}$GCN; (d) R-2D-CNN; (e) CNN-PPF; (f) MFL; (g) JSDF; (h) CAD-GCN.} 
		\label{IPClassificationMaps}
	\end{figure*}
	
	\subsection{Classification Results}
	\label{ClassificationResults}
	
	To show the effectiveness of our proposed CAD-GCN, here we quantitatively and qualitatively evaluate the classification performance by comparing CAD-GCN with the aforementioned baseline methods.
	
	\subsubsection{Results on the Indian Pines Dataset}

	The quantitative results acquired by different methods on the Indian Pines dataset are presented in Table~\ref{IPClassificationResults}, and the highest record regarding each class (i.e., each row) has been highlighted in bold. As shown by Table~\ref{IPClassificationResults}, our proposed CAD-GCN achieves the top level performance among all the methods in terms of OA, AA, and Kappa coefficient, and the standard deviations are very small as well. Meanwhile, the proposed CAD-GCN acquires stable and very high classification accuracies on most of the land-cover classes. All these statistics demonstrate the effectiveness of our CAD-GCN in HSI classification.
	
	The classification maps generated by different methods on the Indian Pines dataset are exhibited in Fig.~\ref{IPClassificationMaps}. To facilitate the comparison among the investigated methods, the ground truth map is also provided in Fig.~\ref{IPClassificationMaps_gt} . A visual inspection reveals that the proposed CAD-GCN method produces much more compact classification map and shows fewer misclassifications than other methods. More concretely, in the classification maps of GCN, S$^{2}$GCN, R-2D-CNN, CNN-PPF, and MFL, the errors are almost uniformly distributed (the salt-and-pepper effect in the homogeneous regions), while in the classification maps of JSDF and our proposed CAD-GCN, the errors only appear in some highly heterogeneous areas, where the spatial separability between classes is quite low. For instance, in Figs.~\ref{IP_GCN}-\ref{IP_NMFL}, the middle and the bottom left parts of the classification maps which correspond to `Soybean-mintill' are highly confusing. Moreover, by comparing CAD-GCN with JSDF, we can also find that JSDF produces more errors around class boundaries than our CAD-GCN method, which reveals the good discriminability of the proposed CAD-GCN in boundary regions.

	\subsubsection{Results on the University of Pavia Dataset}

	\begin{table*}[!t]
		\centering
		\caption{Per-Class Accuracy, OA, AA (\%), and Kappa Coefficient Achieved by Different Methods on University of Pavia Dataset}
		\begin{tabular}{cccccccc}
			\toprule
			ID    & GCN \cite{Kipf2016Semi} & S$^{2}$GCN \cite{8474300} & R-2D-CNN \cite{Yang2018Hyperspectral} & CNN-PPF \cite{Li2016Hyperspectral} & MFL \cite{6882821} & JSDF \cite{7360896} & CAD-GCN \\
			\midrule
			1     & 69.78$\pm$4.71 & 92.87$\pm$3.79 & 84.96$\pm$0.56 & \textbf{95.73$\pm$0.80} & 92.46$\pm$0.27 & 82.40$\pm$4.07 & 83.17$\pm$3.61 \\
			2     & 54.10$\pm$10.54 & 87.06$\pm$4.47 & 79.99$\pm$2.29 & 84.01$\pm$1.99 & 87.22$\pm$0.56 & 90.76$\pm$3.74 & \textbf{95.49$\pm$2.13} \\
			3     & 69.69$\pm$4.48 & 87.97$\pm$4.77 & 89.49$\pm$0.17 & 86.45$\pm$1.94 & 81.59$\pm$0.74 & 86.71$\pm$4.14 & \textbf{97.04$\pm$1.73} \\
			4     & 91.23$\pm$7.02 & 90.85$\pm$0.94 & \textbf{98.12$\pm$0.65} & 91.70$\pm$2.06 & 93.69$\pm$0.42 & 92.88$\pm$2.16 & 78.16$\pm$5.19 \\
			5     & 98.74$\pm$0.11 & \textbf{100.00$\pm$0.00} & 99.85$\pm$0.11 & 99.93$\pm$0.04 & 99.26$\pm$0.05 & \textbf{100.00$\pm$0.00} & 98.01$\pm$1.34 \\
			6     & 65.34$\pm$10.53 & 88.69$\pm$2.64 & 76.79$\pm$7.40 & 93.57$\pm$1.28 & 90.38$\pm$0.58 & 94.30$\pm$4.55 & \textbf{96.70$\pm$1.34} \\
			7     & 86.64$\pm$4.68 & 98.88$\pm$1.08 & 88.69$\pm$4.57 & 93.53$\pm$0.72 & \textbf{99.40$\pm$0.05} & 96.62$\pm$1.37 & 99.05$\pm$1.25 \\
			8     & 72.26$\pm$2.63 & 89.97$\pm$3.28 & 67.54$\pm$5.67 & 83.83$\pm$1.60 & 81.74$\pm$0.55 & \textbf{94.69$\pm$3.74} & 93.83$\pm$3.23 \\
			9     & \textbf{99.93$\pm$0.06} & 98.89$\pm$0.53 & 99.84$\pm$0.08 & 99.47$\pm$0.34 & 99.80$\pm$0.03 & 99.56$\pm$0.36 & 82.66$\pm$4.54 \\
			\midrule
			OA    & 66.19$\pm$3.43 & 89.74$\pm$1.70 & 82.38$\pm$0.88 & 88.72$\pm$0.95 & 89.14$\pm$0.26 & 90.82$\pm$1.30 & \textbf{92.32$\pm$0.98} \\
			AA    & 78.63$\pm$1.23 & 92.80$\pm$0.47 & 87.25$\pm$0.68 & 92.02$\pm$0.37 & 91.73$\pm$0.14 & \textbf{93.10$\pm$0.65} & 91.57$\pm$0.71 \\
			Kappa & 58.39$\pm$3.28 & 86.65$\pm$2.06 & 77.31$\pm$0.97 & 85.43$\pm$1.18 & 85.91$\pm$0.32 & 88.02$\pm$1.62 & \textbf{89.91$\pm$1.25} \\
			\bottomrule
		\end{tabular}%
		\label{PUSClassificationResults}%
	\end{table*}%
	
	\begin{figure*}[!t]
		\centering
		\subfigure[]{%
			\label{PUSclassgt}
			\resizebox*{3cm}{!}{\includegraphics{PUS_gt-eps-converted-to.pdf}}}\hspace{25pt}	
		\subfigure[]{%
			\resizebox*{3cm}{!}{\includegraphics{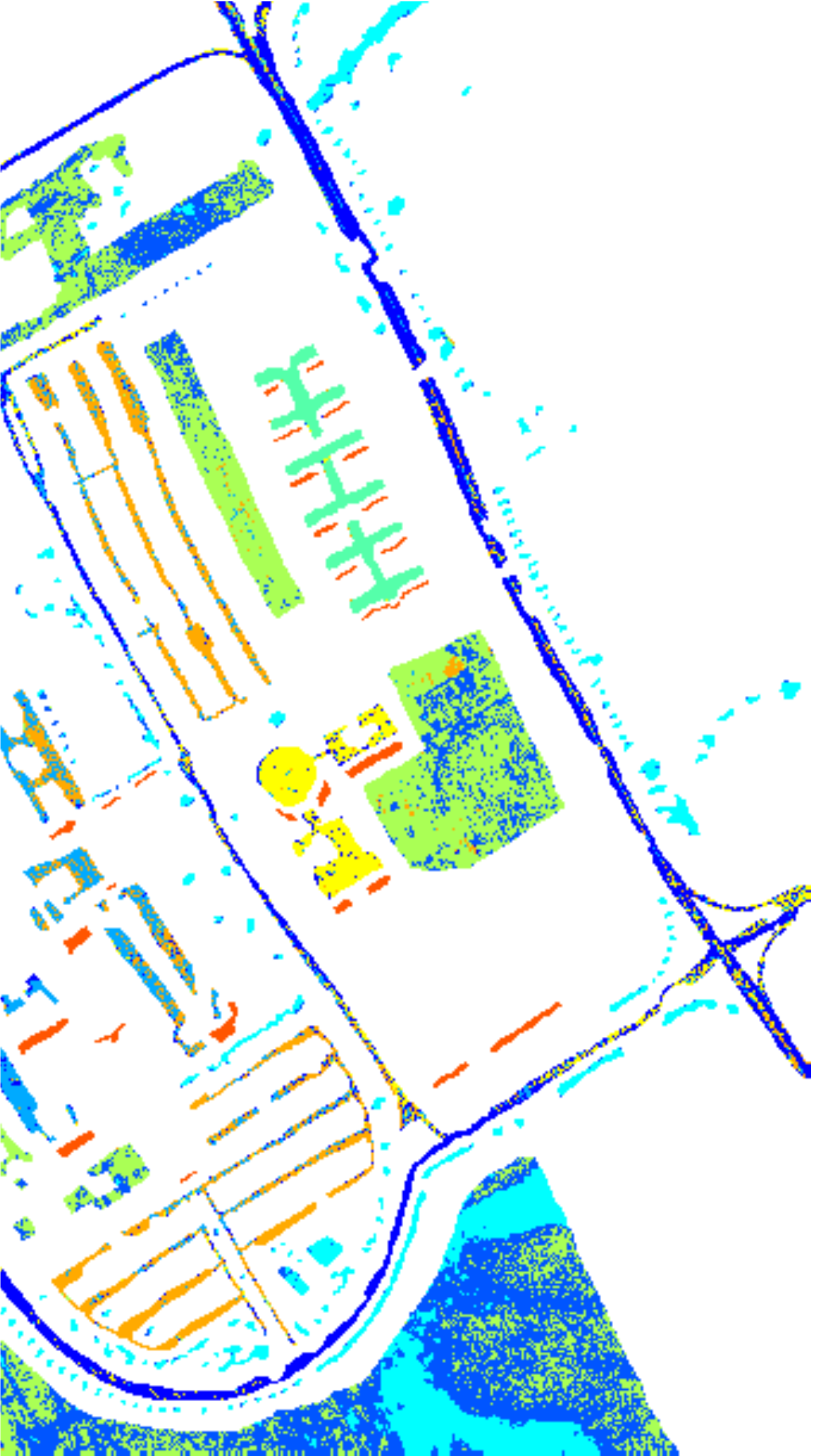}}}\hspace{25pt}
		\subfigure[]{%
			\resizebox*{3cm}{!}{\includegraphics{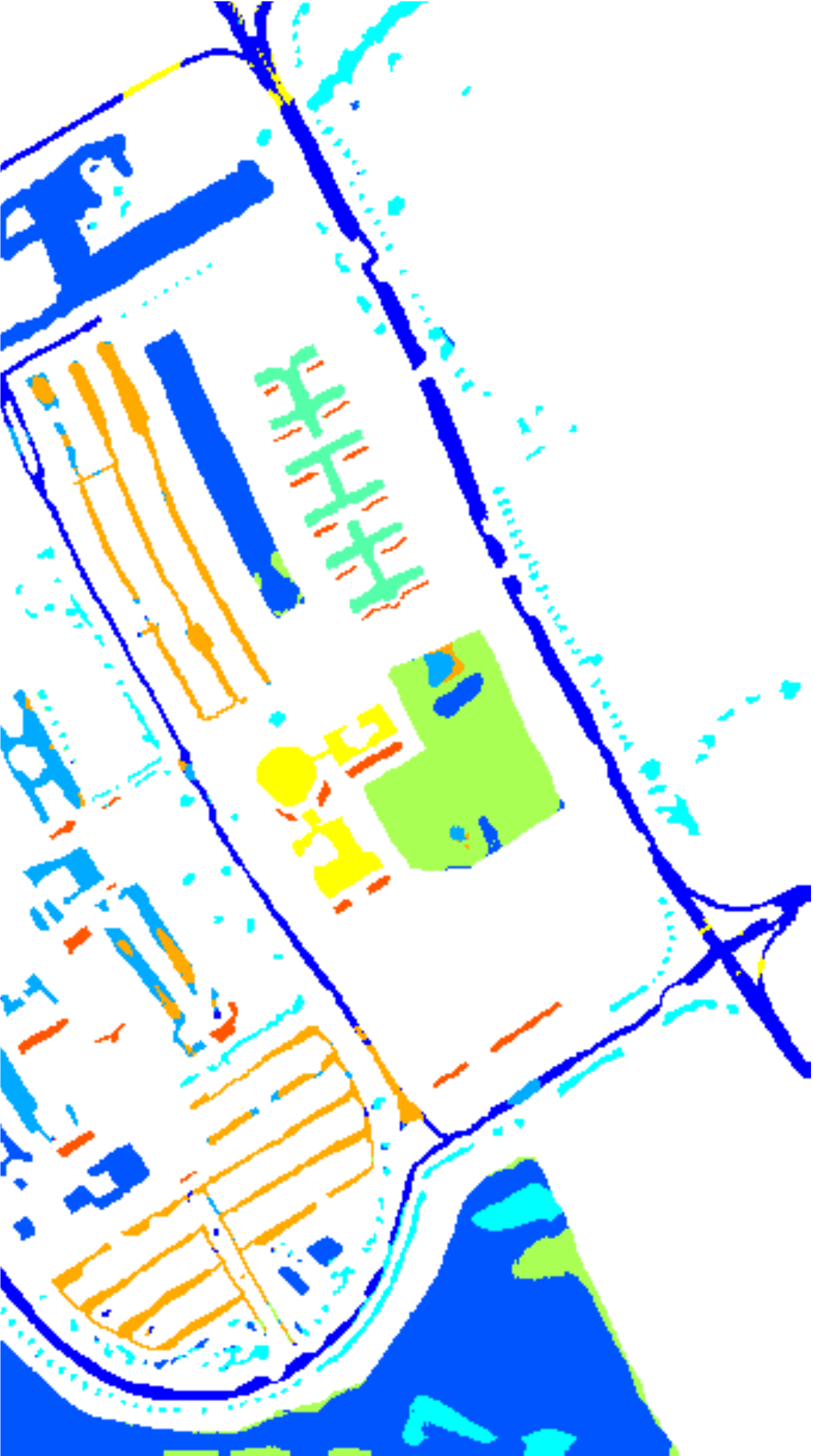}}}\hspace{25pt}
		\subfigure[]{%
			\resizebox*{3cm}{!}{\includegraphics{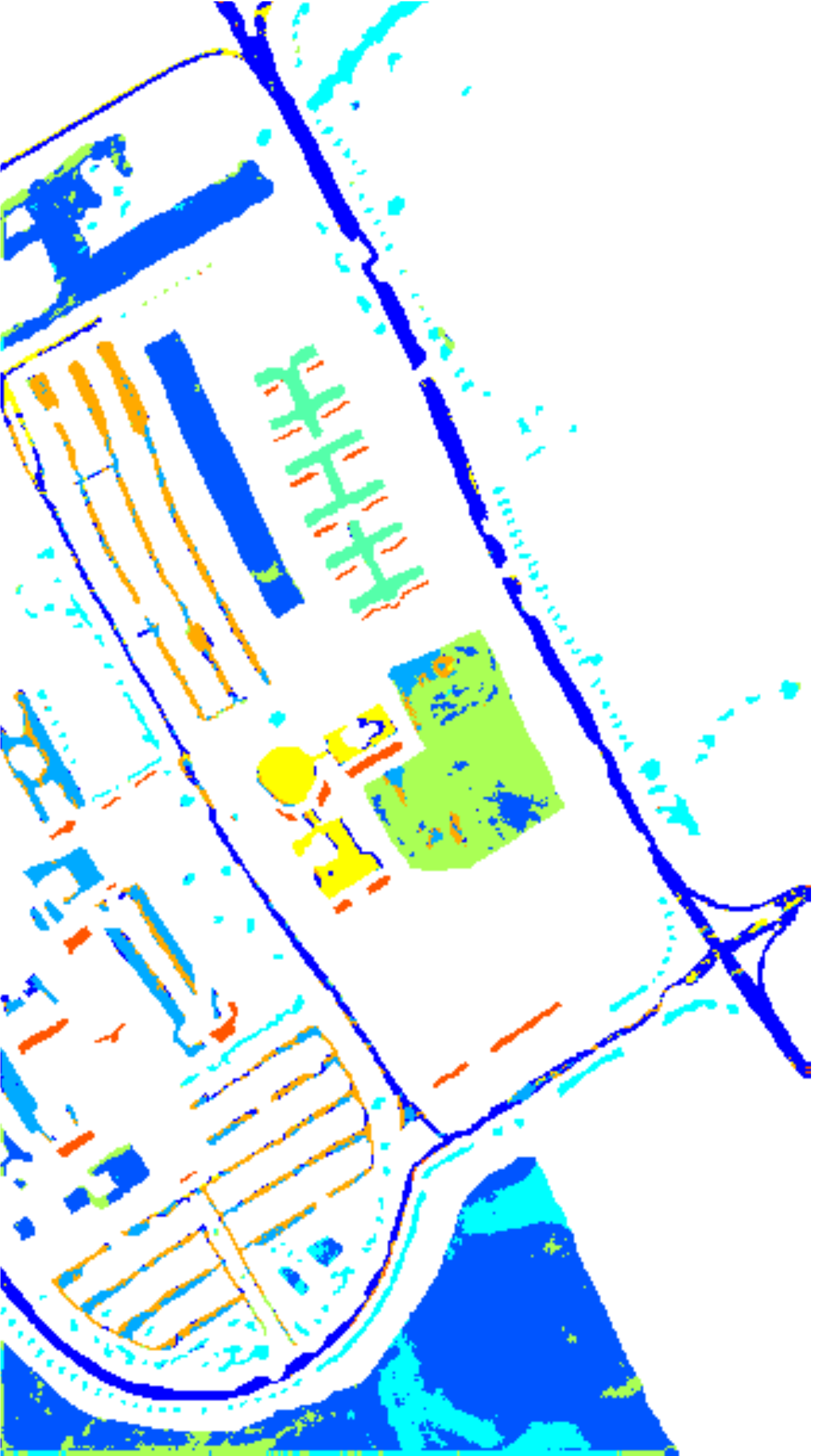}}}\hspace{25pt}	
		\subfigure[]{%
			\resizebox*{3cm}{!}{\includegraphics{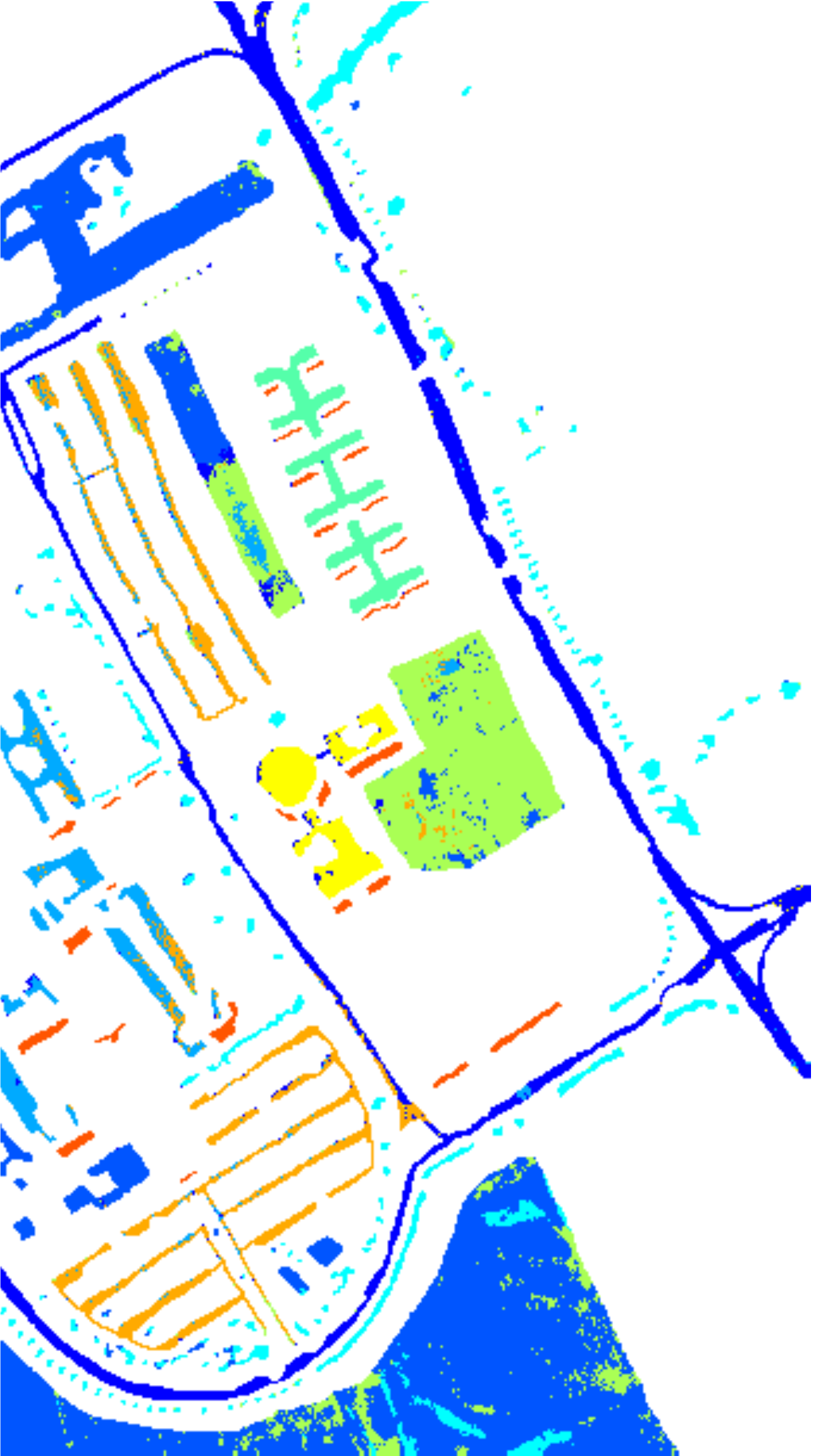}}}\hspace{25pt}
		\subfigure[]{%
			\resizebox*{3cm}{!}{\includegraphics{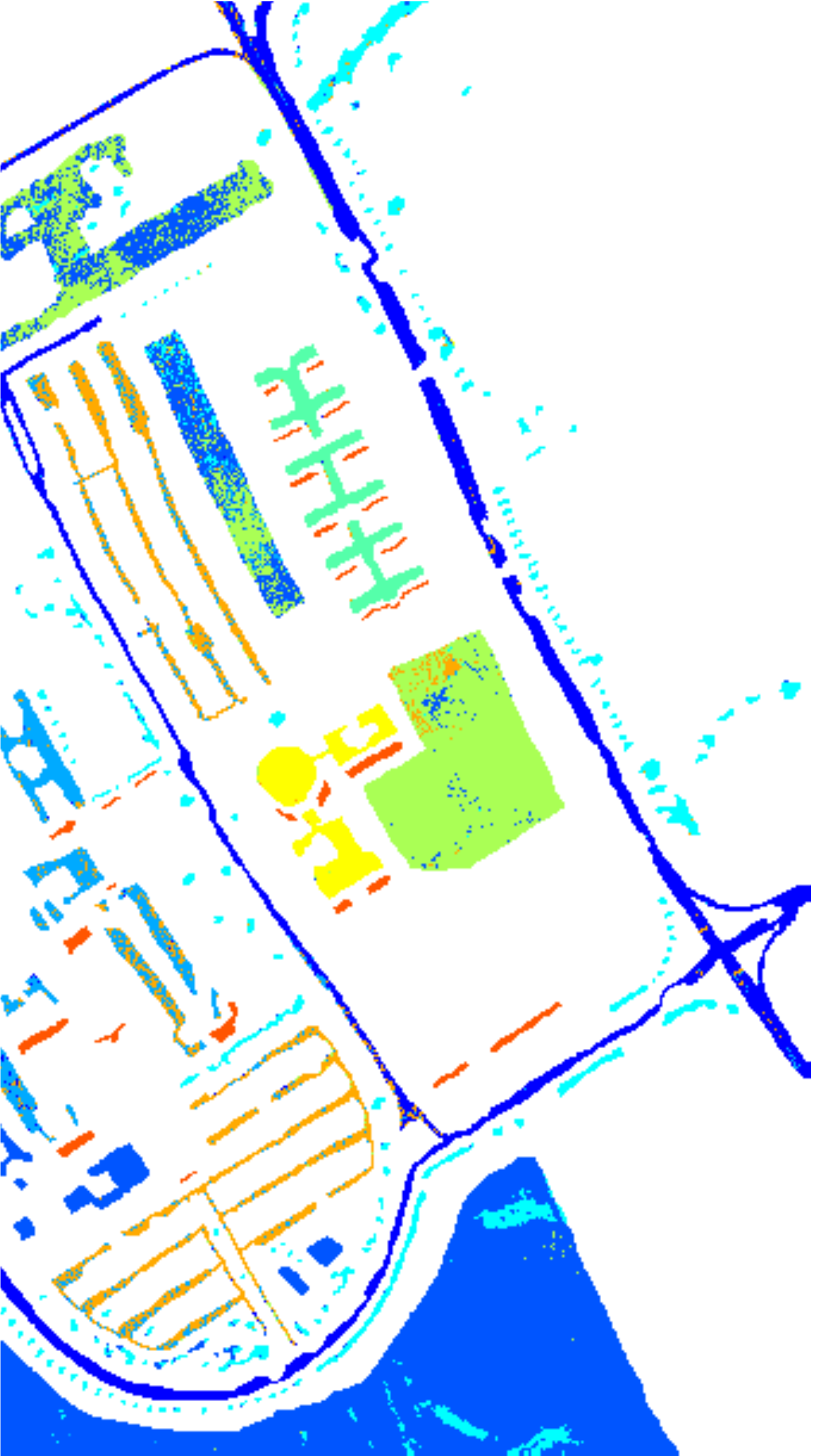}}}\hspace{25pt}		
		\subfigure[]{%
			\resizebox*{3cm}{!}{\includegraphics{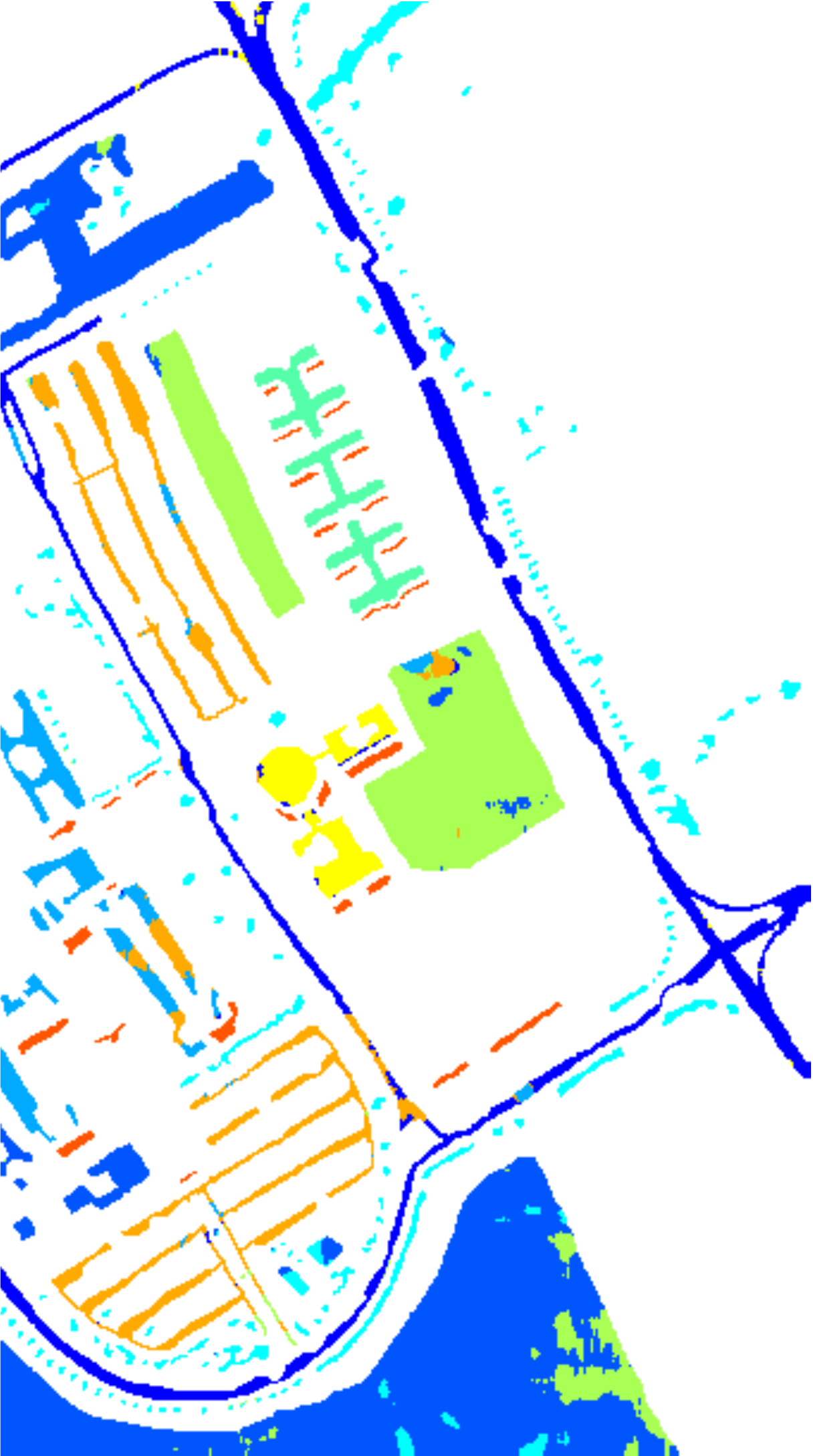}}}\hspace{25pt}	
		\subfigure[]{%
			\label{PUSclassCADGCN}
			\resizebox*{3cm}{!}{\includegraphics{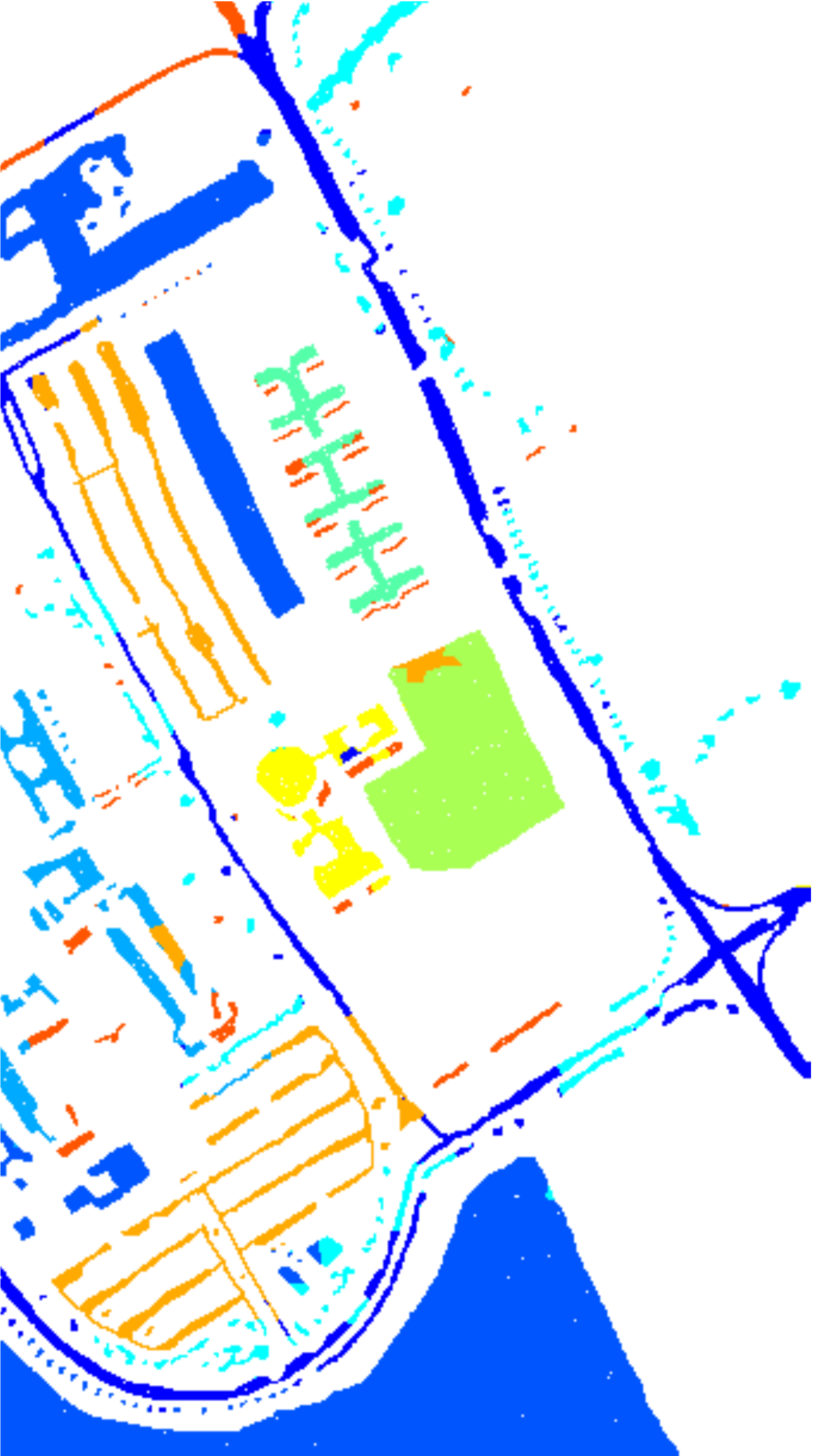}}}\hspace{0pt}	
		
		\subfigure {%
			\resizebox*{!}{0.2cm}{\includegraphics{PUSclass1.pdf}}}\hspace{10pt}
		\subfigure {%
			\resizebox*{!}{0.2cm}{\includegraphics{PUSclass2.pdf}}}\hspace{10pt}
		\subfigure {%
			\resizebox*{!}{0.2cm}{\includegraphics{PUSclass3.pdf}}}\hspace{10pt}
		\subfigure {%
			\resizebox*{!}{0.2cm}{\includegraphics{PUSclass4.pdf}}}\hspace{10pt}
		\subfigure {%
			\resizebox*{!}{0.2cm}{\includegraphics{PUSclass5.pdf}}}\hspace{10pt}
		\subfigure {%
			\resizebox*{!}{0.2cm}{\includegraphics{PUSclass6.pdf}}}\hspace{10pt}	
		\subfigure {%
			\resizebox*{!}{0.2cm}{\includegraphics{PUSclass7.pdf}}}\hspace{10pt}
		\subfigure {%
			\resizebox*{!}{0.2cm}{\includegraphics{PUSclass8.pdf}}}\hspace{10pt}
		\subfigure {%
			\resizebox*{!}{0.2cm}{\includegraphics{PUSclass9.pdf}}}\hspace{0pt}
		
		\caption{Classification maps obtained by different methods on University of Pavia dataset. (a) Ground truth map; (b) GCN; (c) S$^{2}$GCN; (d) R-2D-CNN; (e) CNN-PPF; (f) MFL; (g) JSDF; (h) CAD-GCN.} 
		\label{PUSClassificationMaps}
	\end{figure*}
	
	In Table~\ref{PUSClassificationResults}, different methods are compared on the aforementioned three datasets, where per-class accuracy, OA, AA, and Kappa coefficient are reported, and the highest value in each row is highlighted in bold. From Table~\ref{PUSClassificationResults}, we can conclude that our proposed CAD-GCN method achieves the best result in terms of OA and Kappa coefficient among all the competitors. Compared with the two CNN-based methods (i.e., R-2D-CNN and CNN-PPF), the proposed CAD-GCN increases the OA by 9.94\% and 3.60\%, respectively, which suggests that the refined contextual relations captured by our CAD-GCN is superior to the spatial context characterized by the fixed convolutional kernels of CNN.
	
	Fig.~\ref{PUSClassificationMaps} visualizes the classification results generated by the seven different methods on the University of Pavia dataset. As depicted in Fig.~\ref{PUSclassCADGCN}, the classification map of our proposed CAD-GCN are noticeably closer to the ground truth map (see Fig.~\ref{PUSclassgt}) than other methods, which is consistent with previous results in Table~\ref{PUSClassificationResults}. Although GCN and S$^{2}$GCN are able to capture the relations among graph nodes, they are not originally designed for accurately encoding the contextual relations of HSI. Different from GCN and S$^{2}$GCN, our CAD-GCN employs graph projection and dynamic graph refinement operations to effectively exploit the improved contextual relations of HSI. As a result, GCN and S$^{2}$GCN which use the fixed coarse graph convolution produce more errors than our proposed CAD-GCN.
	
	\subsubsection{Results on the Salinas Dataset}
	
	\begin{table*}[!t]
		\centering
		\caption{Per-Class Accuracy, OA, AA (\%), and Kappa Coefficient Achieved by Different Methods on Salinas Dataset}
		\begin{tabular}{cccccccc}
			\toprule
			ID    & GCN \cite{Kipf2016Semi} & S$^{2}$GCN \cite{8474300} & R-2D-CNN \cite{Yang2018Hyperspectral} & CNN-PPF \cite{Li2016Hyperspectral} & MFL \cite{6882821} & JSDF \cite{7360896} & CAD-GCN \\
			\midrule
			1     & 98.62$\pm$0.86 & 99.01$\pm$0.44 & 98.17$\pm$0.34 & 99.77$\pm$0.21 & 98.41$\pm$0.09 & \textbf{100.00$\pm$0.00} & \textbf{100.00$\pm$0.00} \\
			2     & 99.07$\pm$1.21 & 99.18$\pm$0.59 & 97.79$\pm$1.10 & 98.69$\pm$0.89 & 99.04$\pm$0.06 & \textbf{100.00$\pm$0.00} & \textbf{100.00$\pm$0.00} \\
			3     & 97.03$\pm$1.10 & 97.15$\pm$2.76 & 94.56$\pm$3.84 & 99.50$\pm$0.49 & 99.74$\pm$0.04 & \textbf{100.00$\pm$0.00} & 99.96$\pm$0.09 \\
			4     & 99.28$\pm$0.49 & 99.11$\pm$0.55 & 96.86$\pm$1.03 & 99.81$\pm$0.04 & 98.43$\pm$0.14 & \textbf{99.93$\pm$0.09} & 98.12$\pm$1.62 \\
			5     & 98.58$\pm$0.79 & 97.55$\pm$2.35 & 97.54$\pm$1.08 & 96.64$\pm$1.26 & 98.53$\pm$0.02 & \textbf{99.77$\pm$0.31} & 98.05$\pm$0.93 \\
			6     & 99.58$\pm$0.30 & 99.32$\pm$0.35 & 98.88$\pm$0.44 & 99.32$\pm$0.86 & 98.97$\pm$0.11 & \textbf{100.00$\pm$0.00} & 99.66$\pm$0.31 \\
			7     & 99.13$\pm$0.25 & 99.06$\pm$0.27 & 97.98$\pm$1.12 & 99.59$\pm$0.13 & 99.14$\pm$0.03 & \textbf{99.99$\pm$0.01} & 99.15$\pm$1.73 \\
			8     & 67.94$\pm$8.33 & 70.68$\pm$5.20 & 69.76$\pm$9.72 & 74.77$\pm$4.01 & 69.74$\pm$0.86 & 87.79$\pm$4.89 & \textbf{95.40$\pm$3.15} \\
			9     & 98.50$\pm$0.85 & 98.32$\pm$1.79 & 97.34$\pm$0.57 & 98.99$\pm$0.18 & 98.95$\pm$0.04 & 99.67$\pm$0.33 & \textbf{100.00$\pm$0.00} \\
			10    & 89.64$\pm$1.57 & 90.97$\pm$2.59 & 88.04$\pm$4.54 & 89.32$\pm$3.04 & 90.66$\pm$0.29 & 96.53$\pm$2.55 & \textbf{98.10$\pm$1.17} \\
			11    & 94.80$\pm$2.98 & 98.00$\pm$1.65 & 93.43$\pm$1.56 & 97.65$\pm$1.49 & 93.85$\pm$0.28 & 99.76$\pm$0.21 & \textbf{99.86$\pm$0.26} \\
			12    & 99.71$\pm$0.08 & 99.56$\pm$0.59 & 96.79$\pm$0.99 & 99.82$\pm$0.30 & 97.85$\pm$0.31 & \textbf{100.00$\pm$0.00} & 97.94$\pm$0.84 \\
			13    & 97.99$\pm$0.61 & 97.83$\pm$0.72 & 95.54$\pm$0.81 & 97.70$\pm$0.50 & 99.12$\pm$0.10 & \textbf{100.00$\pm$0.00} & 97.96$\pm$1.01 \\
			14    & 93.58$\pm$2.60 & 95.75$\pm$1.65 & 92.98$\pm$1.85 & 94.14$\pm$1.22 & 94.52$\pm$0.32 & 98.71$\pm$0.72 & \textbf{99.16$\pm$0.71} \\
			15    & 66.18$\pm$9.08 & 70.36$\pm$3.62 & 74.40$\pm$7.97 & 79.12$\pm$1.99 & 71.09$\pm$0.83 & 81.86$\pm$5.26 & \textbf{97.52$\pm$1.92} \\
			16    & 97.24$\pm$1.21 & 96.90$\pm$1.97 & 91.87$\pm$1.87 & 98.65$\pm$0.31 & 99.37$\pm$0.05 & 98.99$\pm$0.63 & \textbf{99.76$\pm$0.67} \\
			\midrule
			OA    & 87.16$\pm$0.85 & 88.39$\pm$1.01 & 87.63$\pm$1.26 & 90.52$\pm$0.77 & 88.36$\pm$0.22 & 94.67$\pm$0.77 & \textbf{98.23$\pm$0.54} \\
			AA    & 93.55$\pm$0.39 & 94.30$\pm$0.47 & 92.62$\pm$0.49 & 95.22$\pm$0.34 & 94.21$\pm$0.08 & 97.69$\pm$0.34 & \textbf{98.79$\pm$0.22} \\
			Kappa & 85.74$\pm$0.92 & 87.10$\pm$1.12 & 86.28$\pm$1.37 & 89.46$\pm$0.85 & 87.06$\pm$0.24 & 94.06$\pm$0.85 & \textbf{98.03$\pm$0.60} \\
			\bottomrule
		\end{tabular}%
		\label{SAClassificationResults}%
	\end{table*}%
	
	\begin{figure*}[!t]
		\centering
		\subfigure[]{%
			\label{SAclassgt}
			\resizebox*{2.3cm}{!}{\includegraphics{SA_gt.pdf}}}\hspace{30pt}	
		\subfigure[]{%
			\resizebox*{2.3cm}{!}{\includegraphics{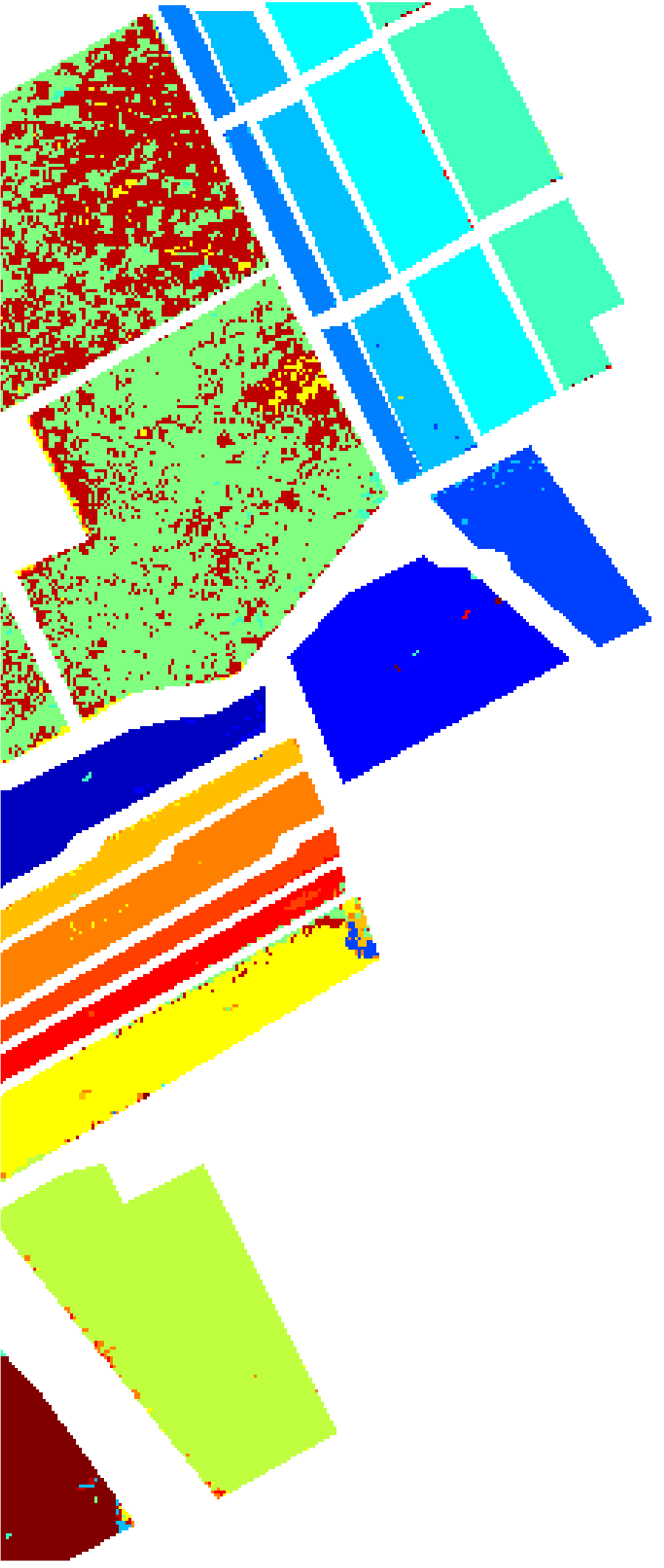}}}\hspace{30pt}
		\subfigure[]{%
			\resizebox*{2.3cm}{!}{\includegraphics{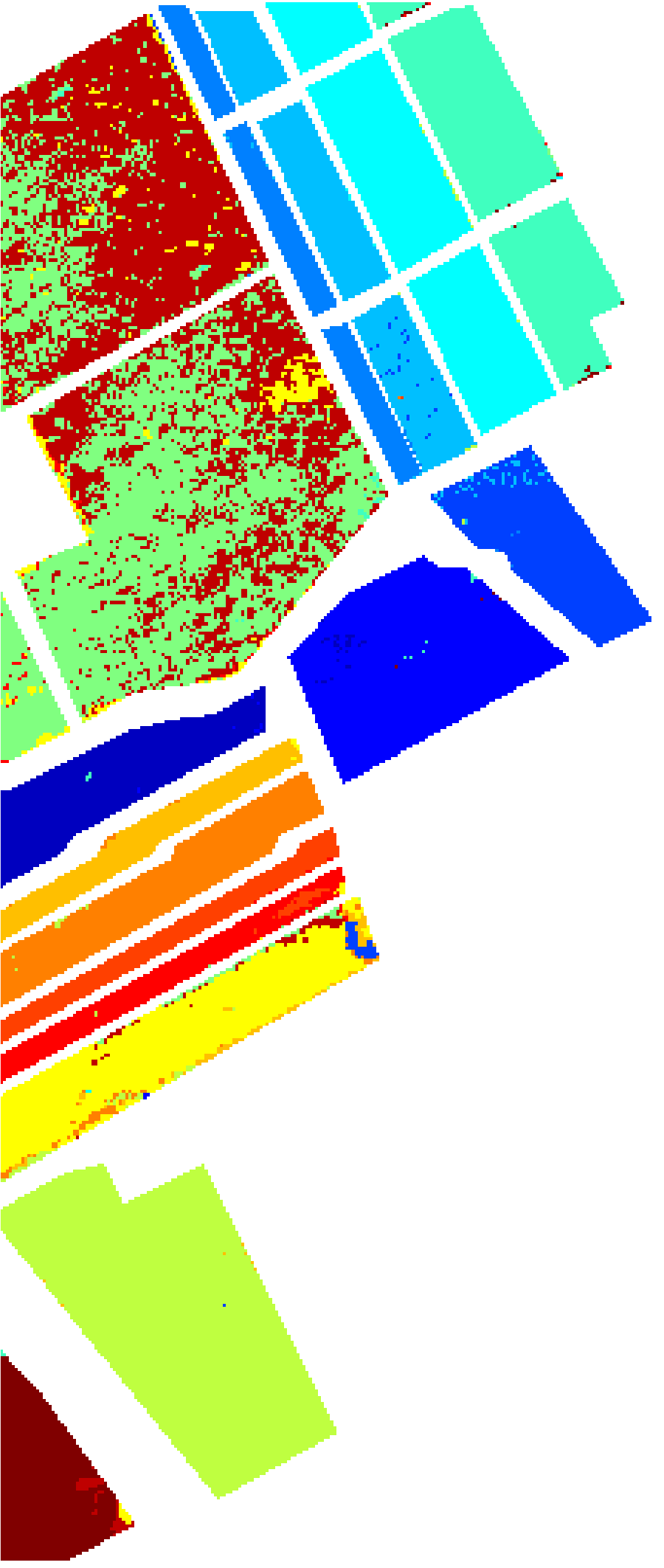}}}\hspace{35pt}
		\subfigure[]{%
			\resizebox*{2.3cm}{!}{\includegraphics{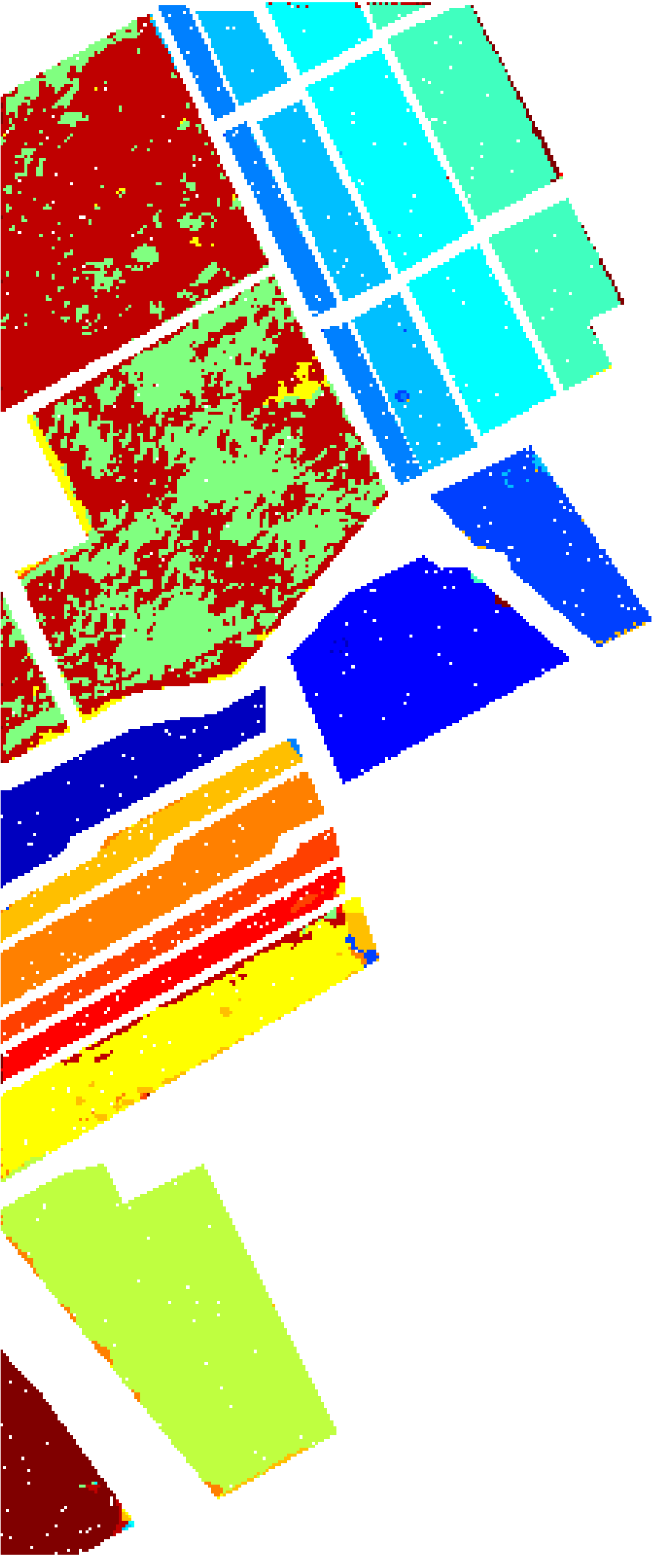}}}\hspace{0pt}
		
		\subfigure[]{%
			\resizebox*{2.3cm}{!}{\includegraphics{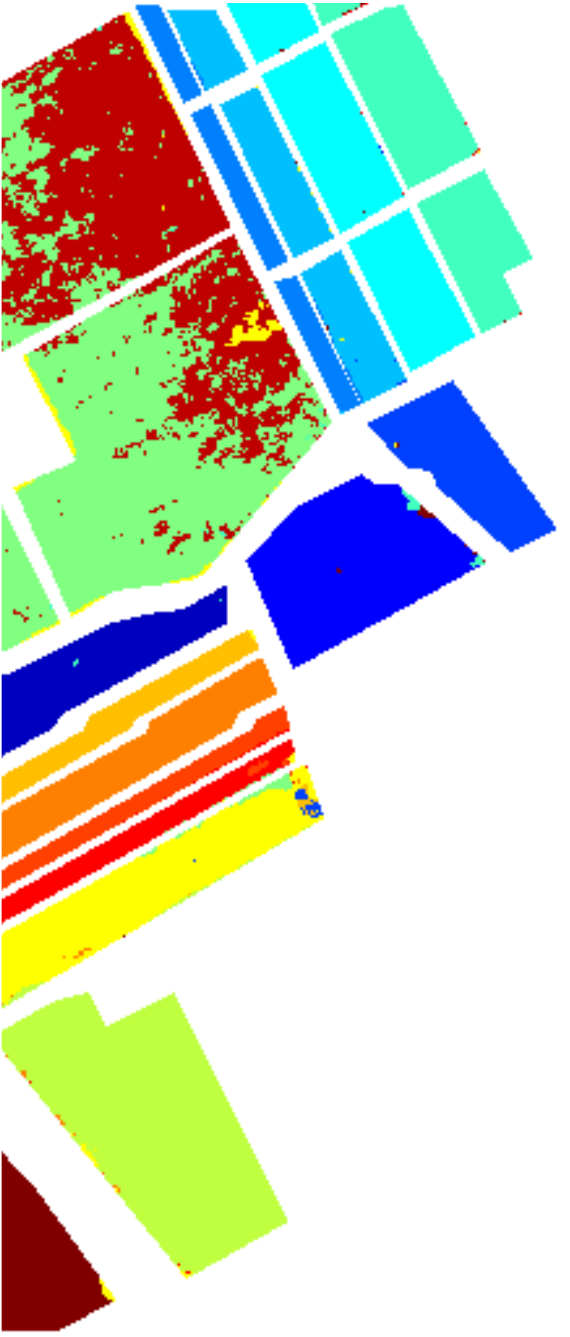}}}\hspace{30pt}
		\subfigure[]{%
			\resizebox*{2.3cm}{!}{\includegraphics{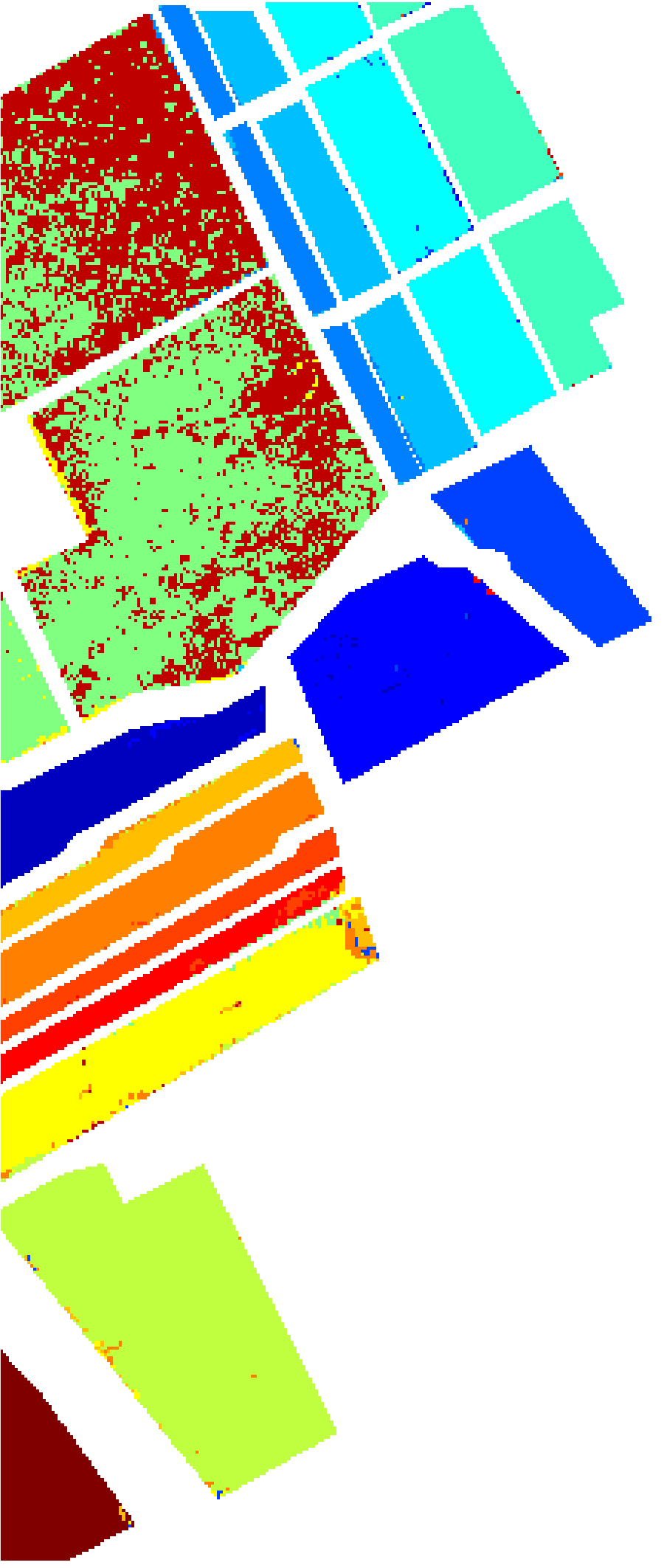}}}\hspace{30pt}		
		\subfigure[]{%
			\resizebox*{2.3cm}{!}{\includegraphics{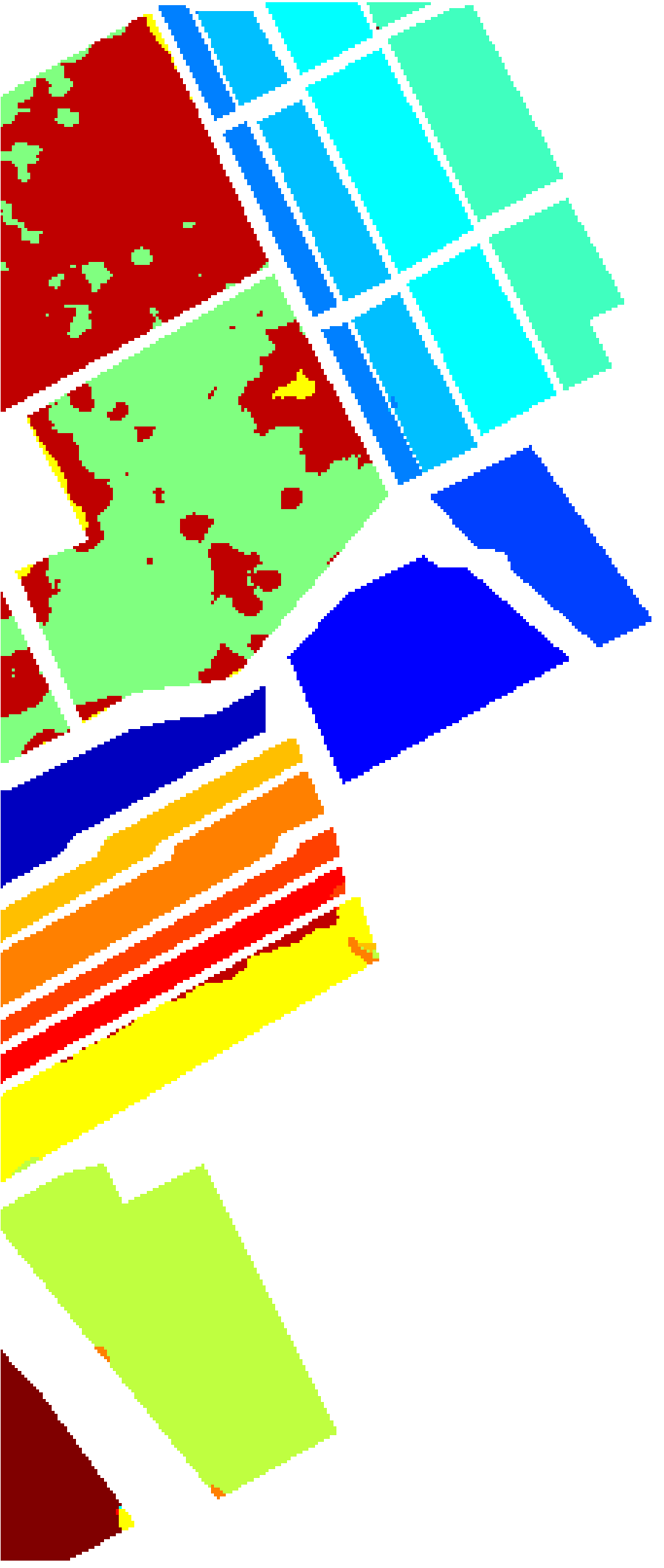}}}\hspace{30pt}	
		\subfigure[]{%
			\label{SAclassCADGCN}
			\resizebox*{2.3cm}{!}{\includegraphics{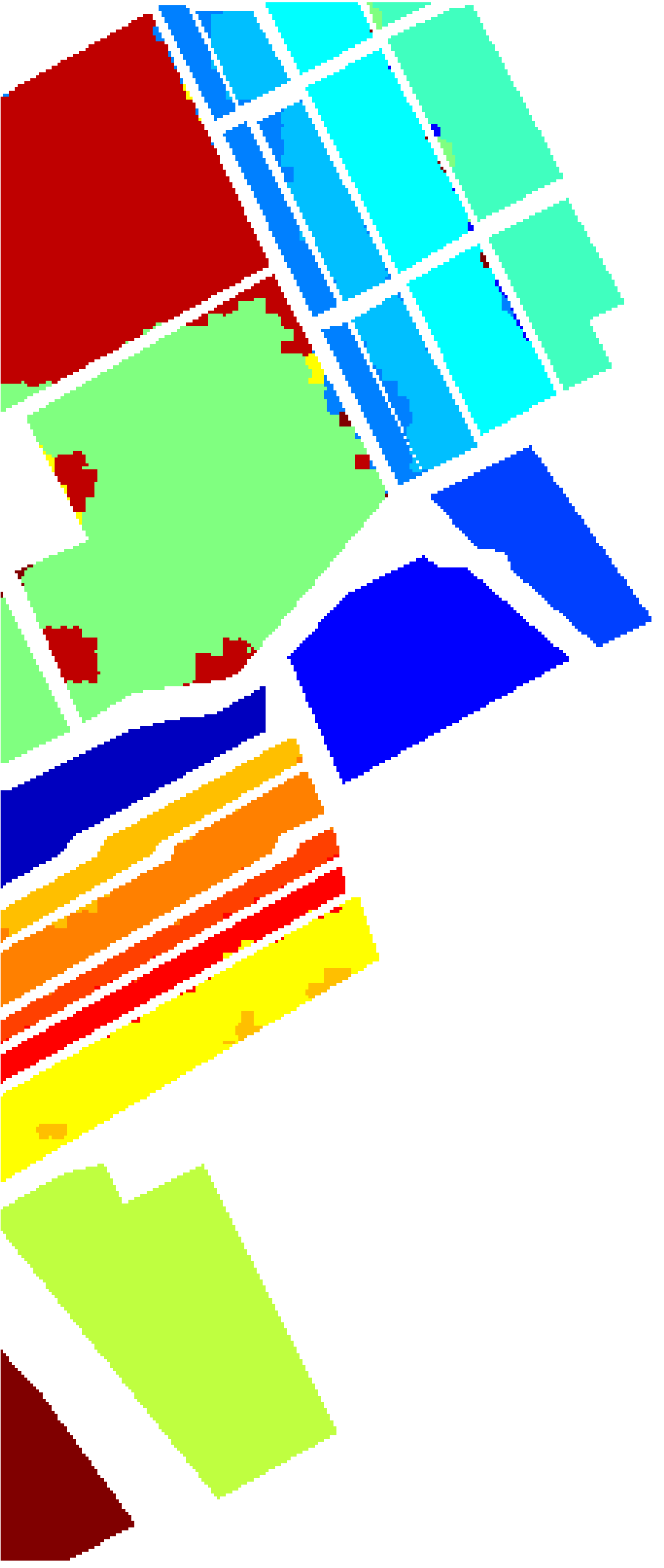}}}\hspace{0pt}
		
		\subfigure {%
			\resizebox*{!}{0.2cm}{\includegraphics{SAclass1.pdf}}}\hspace{1pt}
		\subfigure {%
			\resizebox*{!}{0.2cm}{\includegraphics{SAclass2.pdf}}}\hspace{1pt}
		\subfigure {%
			\resizebox*{!}{0.18cm}{\includegraphics{SAclass3.pdf}}}\hspace{1pt}
		\subfigure {%
			\resizebox*{!}{0.2cm}{\includegraphics{SAclass4.pdf}}}\hspace{1pt}
		\subfigure {%
			\resizebox*{!}{0.2cm}{\includegraphics{SAclass5.pdf}}}\hspace{0pt}
		\subfigure {%
			\resizebox*{!}{0.2cm}{\includegraphics{SAclass6.pdf}}}\hspace{1pt}	
		\subfigure {%
			\resizebox*{!}{0.205cm}{\includegraphics{SAclass7.pdf}}}\hspace{1pt}
		\subfigure {%
			\resizebox*{!}{0.2cm}{\includegraphics{SAclass8.pdf}}}\hspace{0pt}
		\subfigure {%
			\resizebox*{!}{0.2cm}{\includegraphics{SAclass9.pdf}}}\hspace{0pt}	
		\subfigure {%
			\resizebox*{!}{0.2cm}{\includegraphics{SAclass10.pdf}}}\hspace{0pt}
		\subfigure {%
			\resizebox*{!}{0.2cm}{\includegraphics{SAclass11.pdf}}}\hspace{0pt}
		\subfigure {%
			\resizebox*{!}{0.2cm}{\includegraphics{SAclass12.pdf}}}\hspace{0pt}		
		\subfigure {%
			\resizebox*{!}{0.2cm}{\includegraphics{SAclass13.pdf}}}\hspace{0pt}	
		\subfigure {%
			\resizebox*{!}{0.2cm}{\includegraphics{SAclass14.pdf}}}\hspace{0pt}
		\subfigure {%
			\resizebox*{!}{0.2cm}{\includegraphics{SAclass15.pdf}}}\hspace{0pt}		
		\subfigure {%
			\resizebox*{!}{0.2cm}{\includegraphics{SAclass16.pdf}}}\hspace{0pt}		
		
		\caption{Classification maps obtained by different methods on Salinas dataset. (a) Ground truth map; (b) GCN; (c) S$^{2}$GCN; (d) R-2D-CNN; (e) CNN-PPF; (f) MFL; (g) JSDF; (h) CAD-GCN.} 
		\label{SAClassificationMaps}
	\end{figure*}
	
	Table~\ref{SAClassificationResults} presents the experimental results of different methods on the Salinas dataset. The proposed CAD-GCN is obviously superior to the CNN-based methods (i.e., R-2D-CNN and CNN-PPF) and all the other competitors. For instance, in Table~\ref{SAClassificationResults}, CAD-GCN yields over 10\% higher OA than R-2D-CNN and approximately 8\% higher OA than CNN-PPF. Especially in some classes such as `Grapes untrained' ($\rm{ID}=8$) and `Vineyard untrained' ($\rm{ID}=15$), the class-specific accuracies of our proposed CAD-GCN are even approximately 20\% higher than those of the CNN-based methods.
	
	Fig.~\ref{SAClassificationMaps} provides a visual comparison of the classification results obtained by different methods. It is observable that some areas in the classification map of our proposed CAD-GCN are less noisy than those of other methods, e.g., the regions of `Grapes untrained' and `Vineyard untrained', which is in consistence with the results listed in Table~\ref{SAClassificationResults}.

	\subsection{Impact of the Number of Labeled Examples}
	
	\begin{figure*}[!t]
		\centering
		\subfigure[]{
			\label{IPclassnum}
			\resizebox*{5.3cm}{!}{\includegraphics{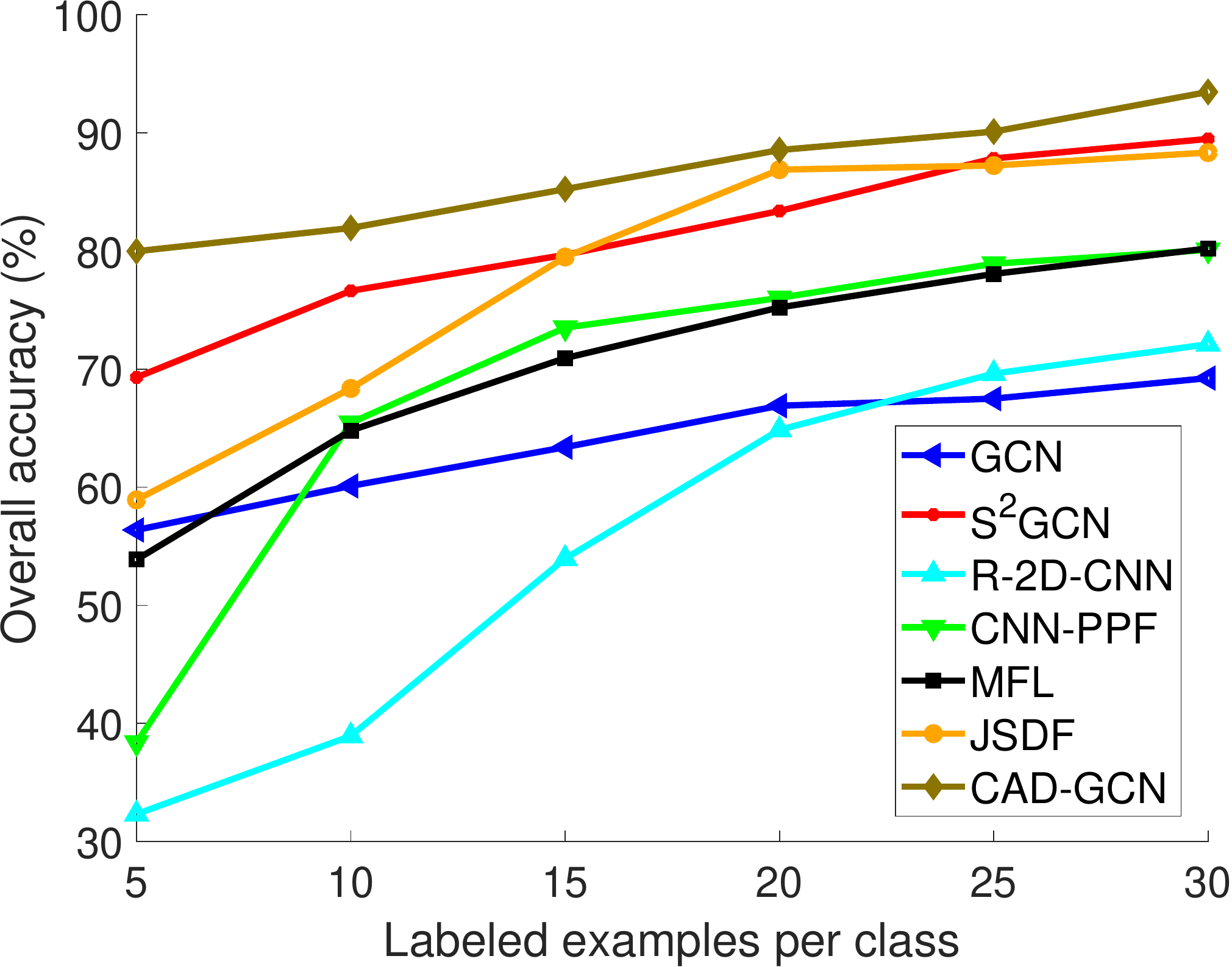}}}
		\subfigure[]{
			\label{paviaUclassnum}
			\resizebox*{5.3 cm}{!}{\includegraphics{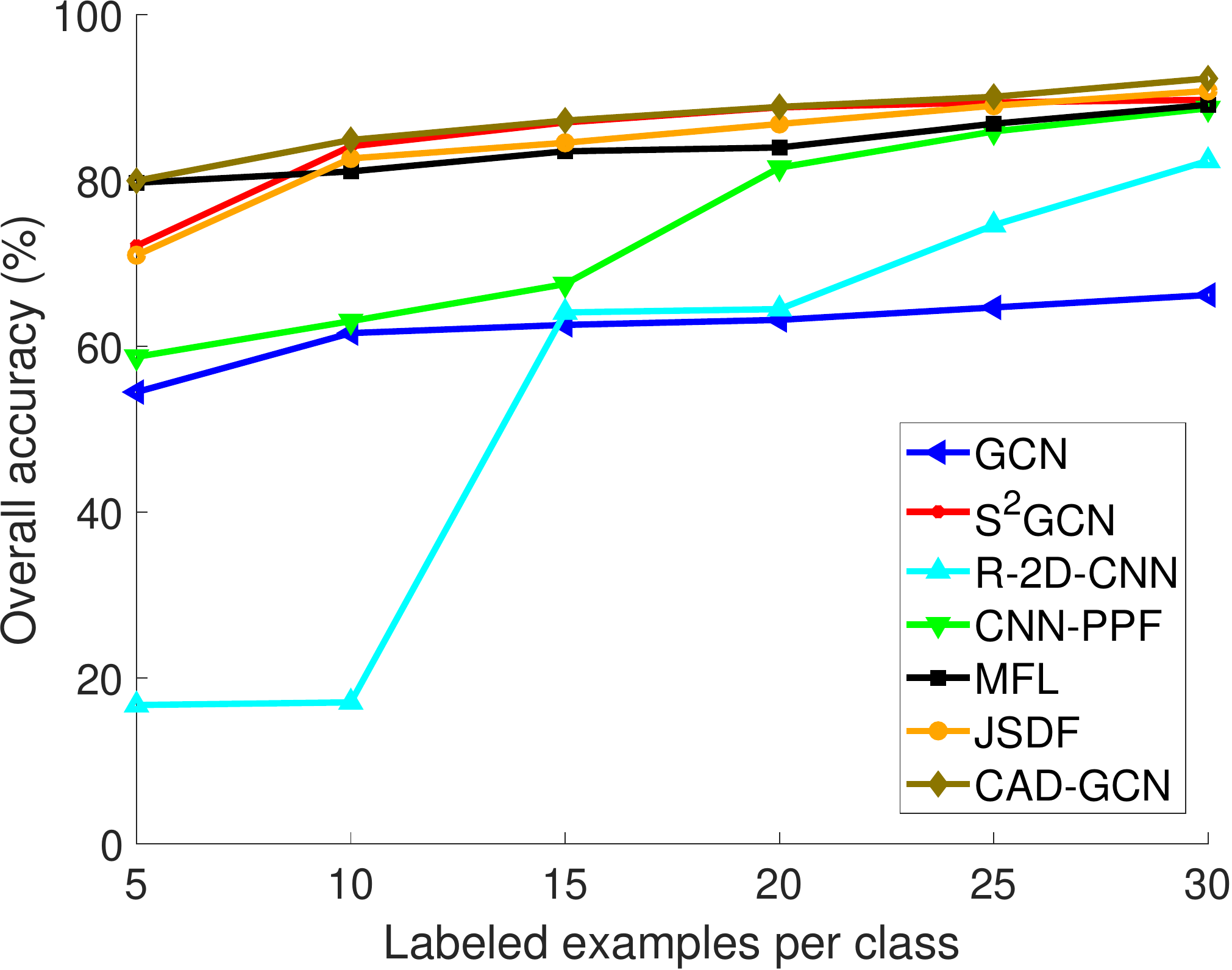}}}
		\subfigure[]{
			\label{SAclassnum}
			\resizebox*{5.3cm}{!}{\includegraphics{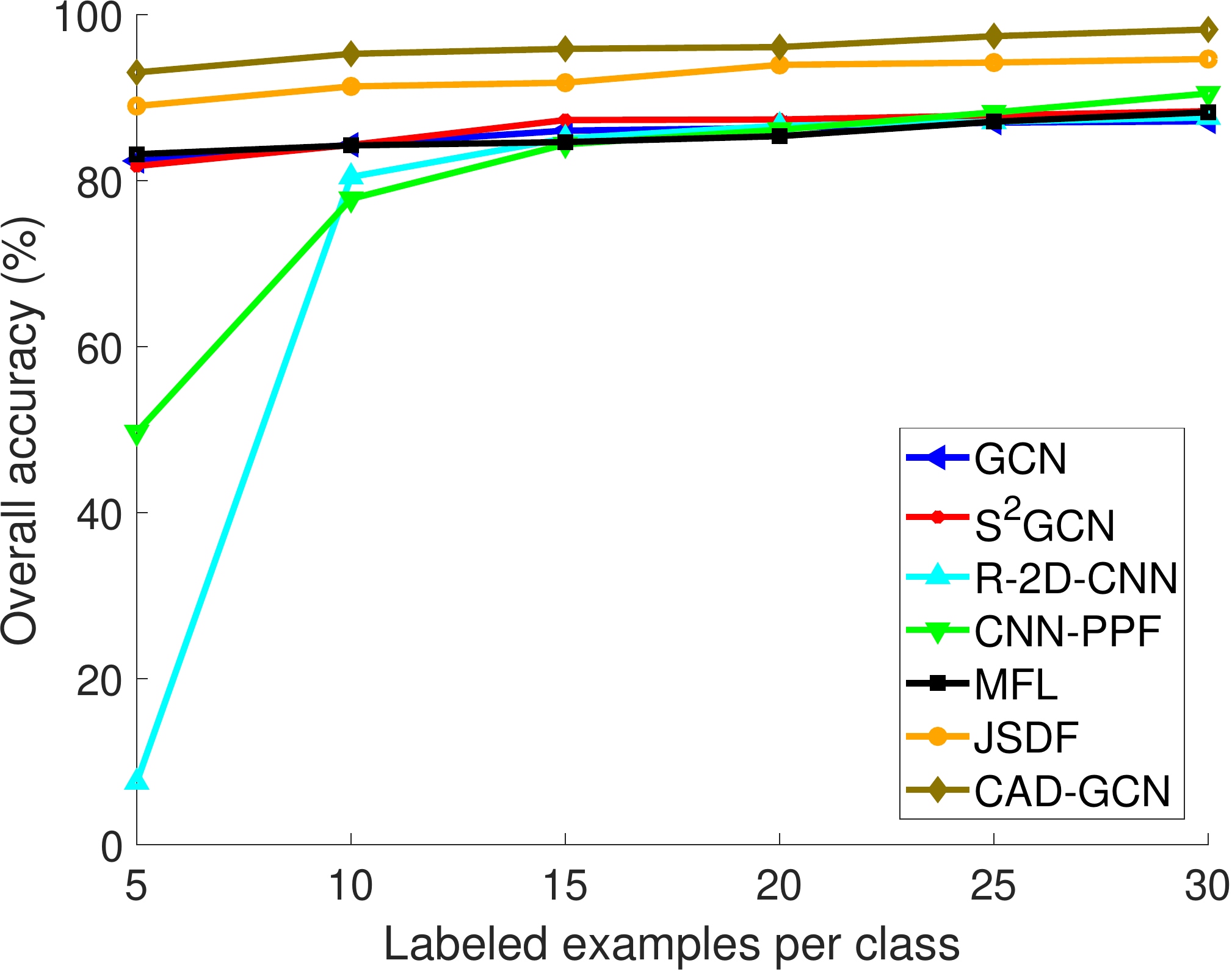}}}
		\caption{Overall accuracies of various methods under different numbers of labeled examples per class. (a) Indian Pines dataset; (b) University of Pavia dataset; (c) Salinas dataset.} 
		\label{classnum_acc}
	\end{figure*}	
	
	In this experiment, the classification performances of the aforementioned seven methods with different numbers of labeled examples (i.e., pixels) for training are investigated. To be specific, we vary the number of labeled examples per class form 5 to 30 with an interval of 5, and report the OA acquired by all the methods on the Indian Pines, the University of Pavia, and the Salinas datasets (see Fig.~\ref{classnum_acc}). From the results, we can find that the proposed CAD-GCN consistently outperforms the GCN, S$^{2}$GCN and all the other competitors on the three datasets, which verifies the effectiveness of contextual relations captured by CAD-GCN. Another interesting observation is that even if the labeled examples are quite limited (i.e., 5 or 10 labeled examples per class), our CAD-GCN still achieves relatively high OA, which suggests good stability of CAD-GCN in HSI classification tasks.
	
	\subsection{Impact of Hyperparameters}
	
	\begin{figure}[!]
		\centering
		\subfigure[]{%
			\resizebox*{3.8cm}{!}{\includegraphics{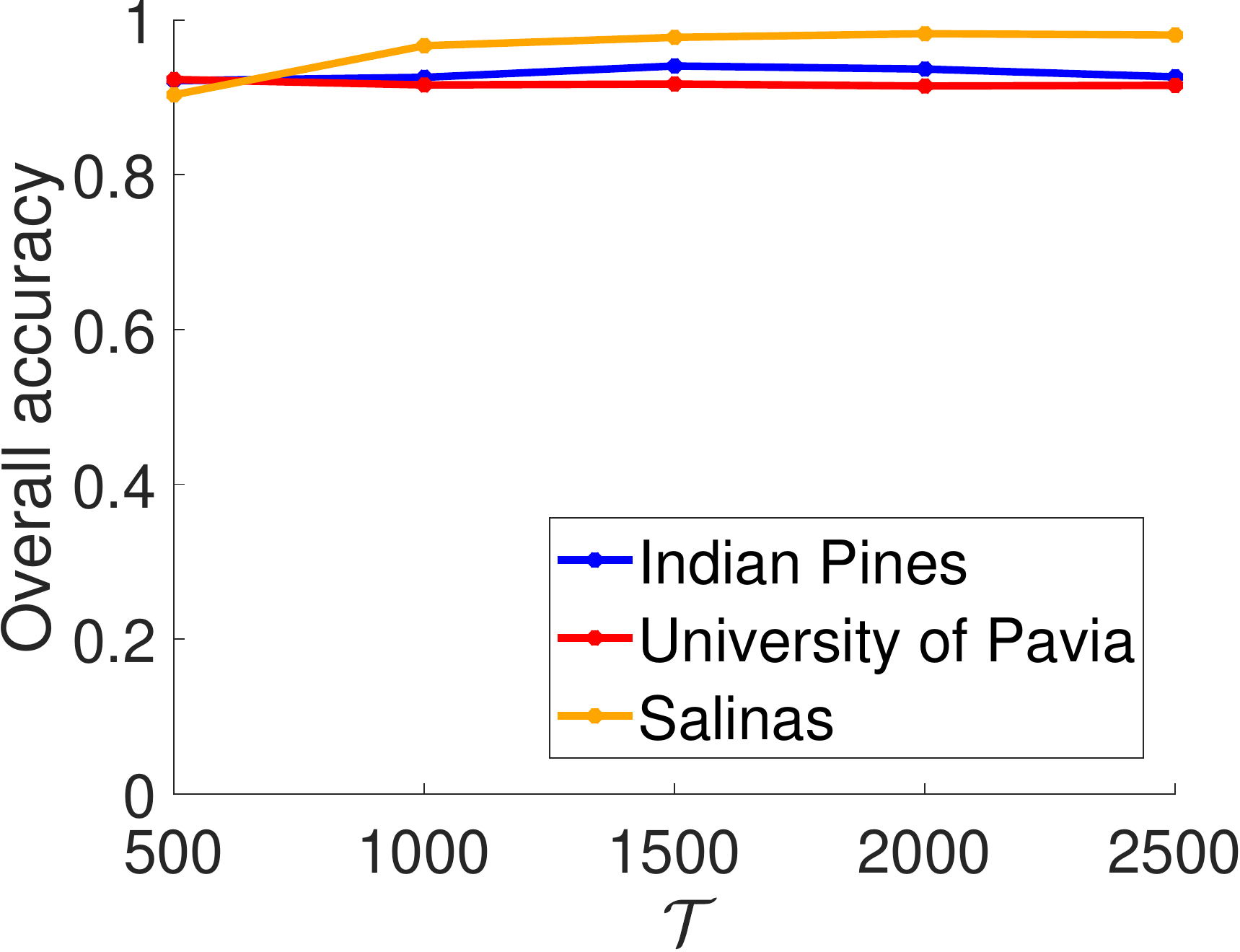}}}\hspace{10pt}
		\subfigure[]{%
			\resizebox*{3.8cm}{2.9cm}{\includegraphics{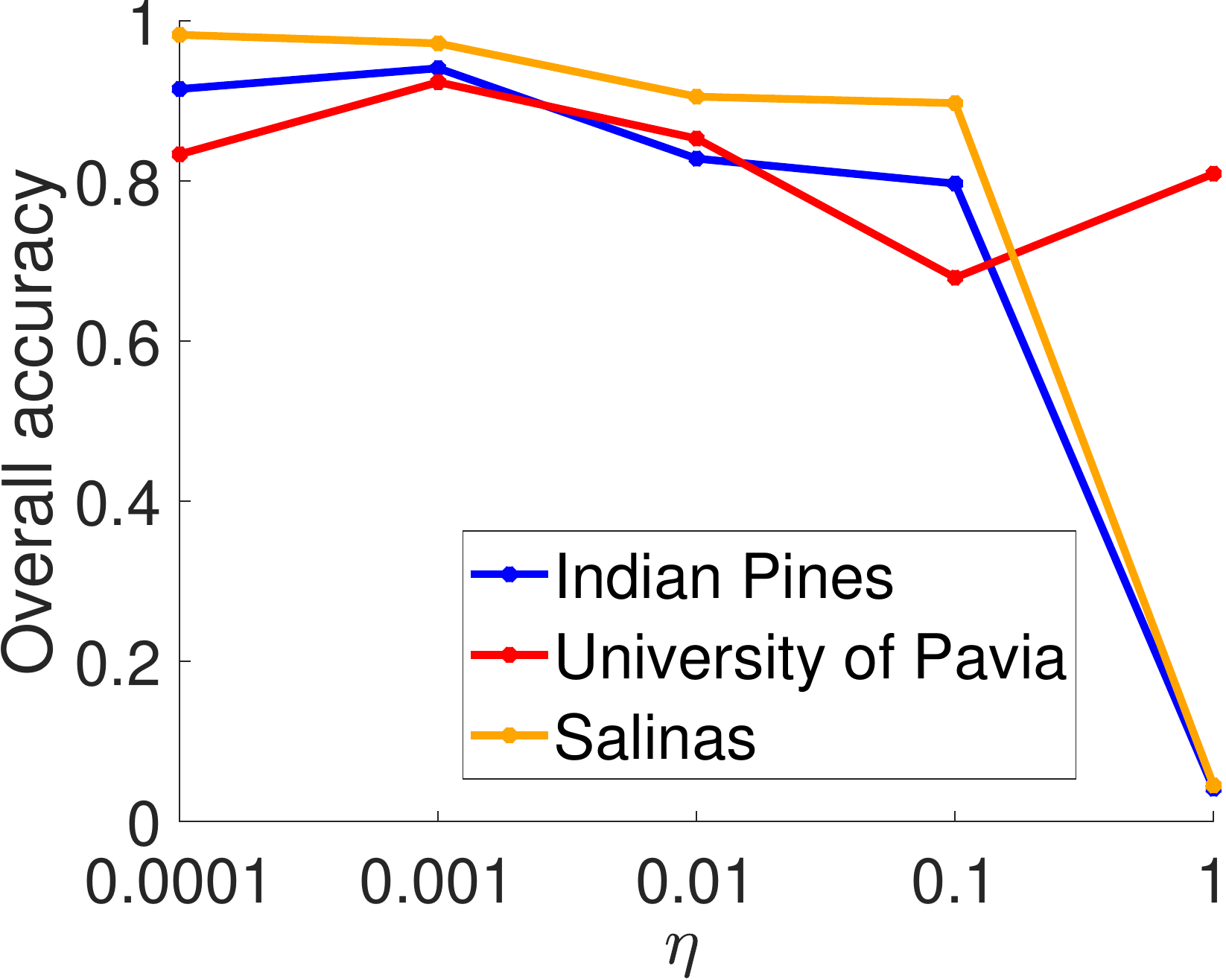}}}\hspace{0pt}
		
		\subfigure[] {%
			\resizebox*{3.8cm}{!}{\includegraphics{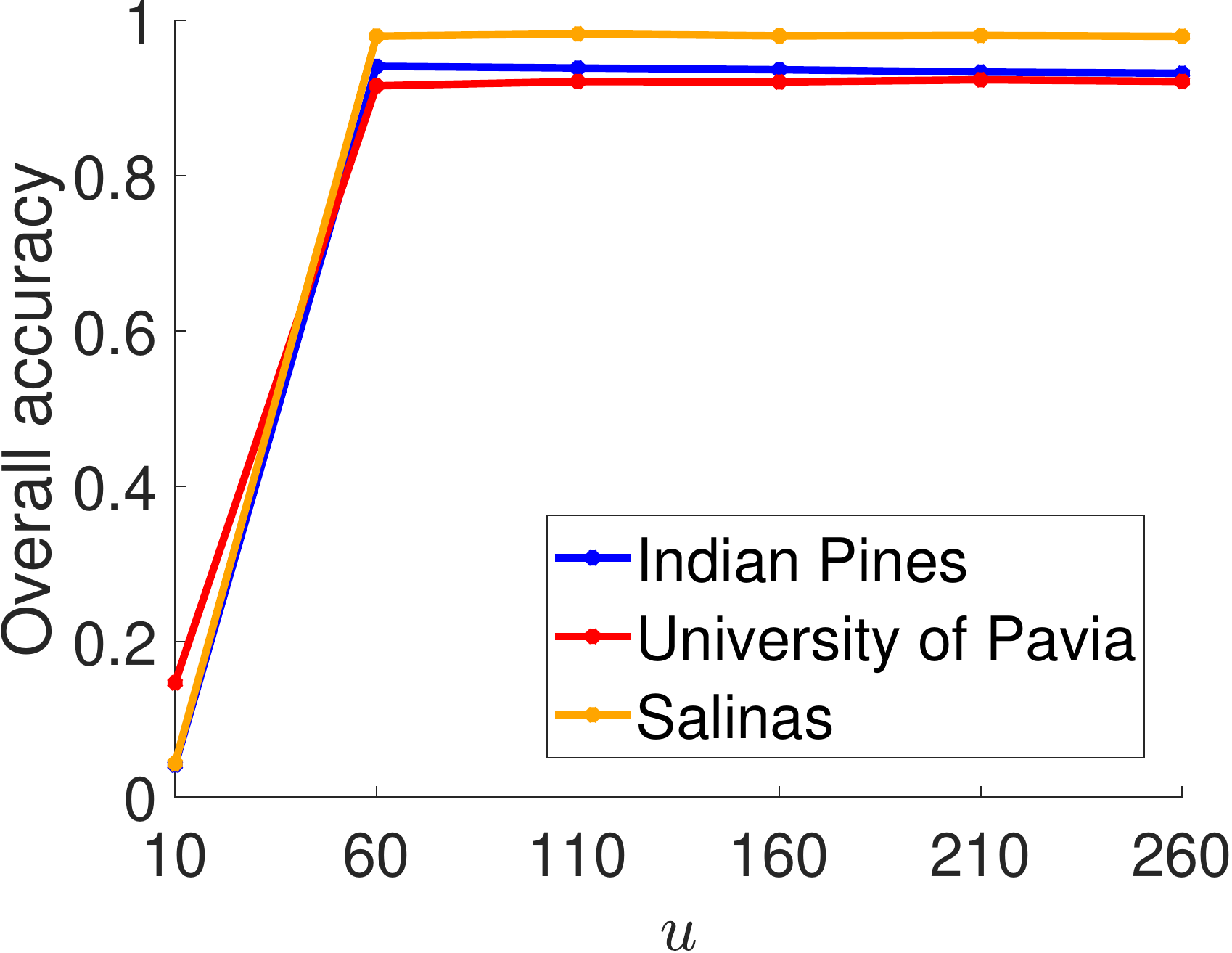}}}\hspace{9pt}
		\subfigure[] {%
			\label{para_beta}
			\resizebox*{3.8cm}{!}{\includegraphics{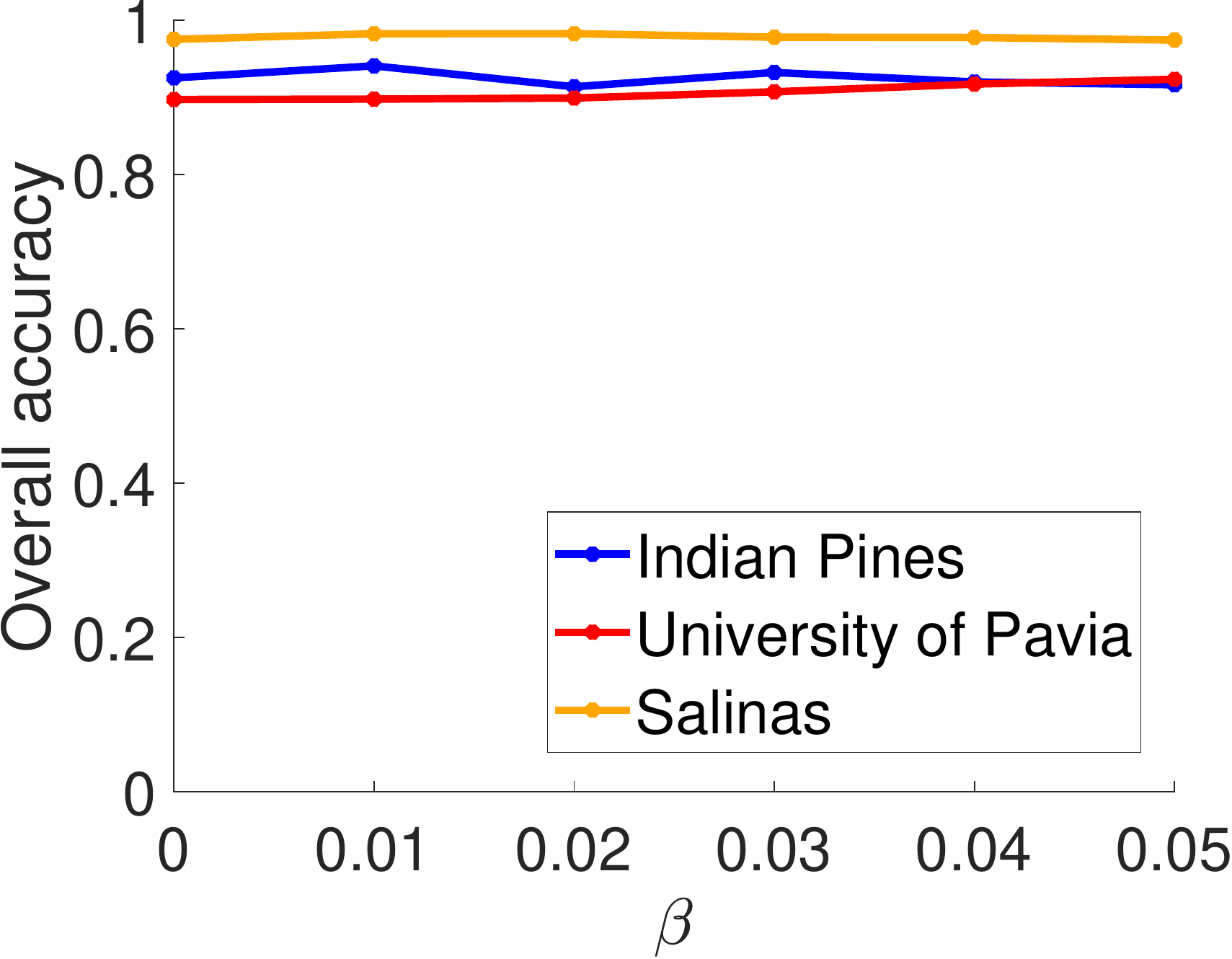}}}\hspace{0pt}
		

		\caption{Parametric sensitivity of (a) $\mathcal{T}$, (b) $\eta$, (c) $u$, and (d) $\beta$ of our proposed CAD-GCN.} 
		\label{ParametricSensitivity}
	\end{figure}
	
	There are several important hyperparameters that should be manually tuned in the designed CAD-GCN architecture. Here, we will evaluate in detail the sensitivity of the classification performance to different hyperparameter settings of the proposed CAD-GCN. Since GCN-based methods usually do not require deep structure to achieve excellent performance \cite{8474300, Gao:2018:LLG:3219819.3219947}, we empirically employ two convolutional layers for all the three datasets. The hyperparameters to be pre-tuned manually mainly include the number of iterations $\mathcal{T}$, the learning rate $\eta$, the number of hidden units $u$, and the threshold $\beta$ used in dynamic graph refinement. Therefore, we examine the test accuracy of CAD-GCN by varying one of $\mathcal{T}$, $\eta$, $u$, and $\beta$, and meanwhile fixing the remaining hyperparameters to a constant value \cite{7010929}. The results on Indian Pines, University of Pavia, and Salinas datasets are shown in Fig.~\ref{ParametricSensitivity}.
	
	From the curves presented in Fig.~\ref{ParametricSensitivity}, it can be observed that the aforementioned four hyperparameters are critical for the proposed CAD-GCN to achieve satisfactory performance. In order to obtain promising classification performance, we set the hyperparameters to $\mathcal{T} = 1500$, $\eta = 0.001$, $u = 60$, $\beta = 0.01$ for the Indian Pines dataset, $\mathcal{T} = 500$, $\eta = 0.001$, $u = 210$, $\beta = 0.05$ for the University of Pavia dataset, and $\mathcal{T} = 2000$, $\eta = 0.0001$, $u = 110$, $\beta = 0.02$ for the Salinas dataset, respectively. Note that when the threshold $\beta$ used in the edge filter (see Eq.~\eqref{EdgeFilter}) becomes too large, the useful information contained in the graph may be filtered out. Hence, we search for the optimal $\beta$ between zero and a relatively small value 0.05.

	\subsection{Ablation Study}
	
	\begin{table}[!]
		\scriptsize
		\centering
		\caption{Per-Class Accuracy, OA, AA (\%), and Kappa Coefficient Achieved by Different Model Settings on Indian Pines Dataset}
		\begin{tabular}{ccccc}
			\toprule
			ID    & CAD-GCN-v1 & CAD-GCN-v2 & CAD-GCN-v3 & CAD-GCN \\
			\midrule
			1     & 98.85$\pm$2.05 & 99.03$\pm$1.98 & \textbf{100.00$\pm$0.00} & \textbf{100.00$\pm$0.00} \\
			2     & 86.90$\pm$4.83 & 86.52$\pm$4.78 & 77.42$\pm$1.88 & \textbf{89.24$\pm$4.11} \\
			3     & 92.85$\pm$4.98 & 92.94$\pm$3.75 & 92.28$\pm$0.60 & \textbf{94.20$\pm$3.26} \\
			4     & 98.04$\pm$1.96 & 97.92$\pm$1.95 & \textbf{99.69$\pm$1.40} & 98.52$\pm$1.15 \\
			5     & 94.63$\pm$3.22 & 93.75$\pm$3.10 & 93.59$\pm$1.42 & \textbf{95.12$\pm$1.89} \\
			6     & 97.50$\pm$2.01 & \textbf{97.84$\pm$1.99} & 96.11$\pm$0.65 & 95.42$\pm$5.02 \\
			7     & 96.14$\pm$5.42 & 96.41$\pm$5.18 & \textbf{100.00$\pm$0.00} & 98.65$\pm$2.68 \\
			8     & 99.43$\pm$1.31 & 98.93$\pm$1.64 & 96.08$\pm$0.98 & \textbf{99.68$\pm$0.54} \\
			9     & \textbf{100.00$\pm$0.00} & \textbf{100.00$\pm$0.00} & \textbf{100.00$\pm$0.00} & \textbf{100.00$\pm$0.00} \\
			10    & 90.31$\pm$1.89 & 89.14$\pm$3.77 & \textbf{92.21$\pm$0.80} & 89.09$\pm$4.19 \\
			11    & 89.43$\pm$3.88 & 89.01$\pm$3.69 & 88.14$\pm$2.13 & \textbf{92.18$\pm$2.00} \\
			12    & 91.61$\pm$3.91 & 93.59$\pm$3.27 & 87.90$\pm$1.13 & \textbf{96.03$\pm$2.65} \\
			13    & 99.83$\pm$0.32 & 99.81$\pm$0.32 & \textbf{99.86$\pm$0.64} & 99.75$\pm$0.29 \\
			14    & 97.69$\pm$2.21 & 98.88$\pm$2.41 & 96.08$\pm$0.26 & \textbf{99.36$\pm$0.69} \\
			15    & 98.34$\pm$2.56 & 99.00$\pm$1.00 & 95.32$\pm$0.33 & \textbf{99.04$\pm$1.12} \\
			16    & \textbf{99.05$\pm$1.18} & 98.20$\pm$2.71 & 94.13$\pm$1.55 & 98.18$\pm$3.89 \\
			\midrule
			OA    & 92.68$\pm$1.00 & 92.66$\pm$1.14 & 90.30$\pm$0.45 & \textbf{94.06$\pm$0.85} \\
			AA    & 95.66$\pm$0.60 & 95.69$\pm$0.64 & 94.30$\pm$0.22 & \textbf{96.53$\pm$0.55} \\
			Kappa & 91.67$\pm$1.13 & 91.64$\pm$1.27 & 88.94$\pm$0.50 & \textbf{93.22$\pm$0.97} \\
			\bottomrule
		\end{tabular}%
		\label{TestCADGCN_IP}%
	\end{table}%
	
	\begin{table}[!t]
		\scriptsize
		\centering
		\caption{Per-Class Accuracy, OA, AA (\%), and Kappa Coefficient Achieved by Different Model Settings on University of Pavia Dataset}
		\begin{tabular}{ccccc}
			\toprule
			ID    & CAD-GCN-v1 & CAD-GCN-v2 & CAD-GCN-v3 & CAD-GCN \\
			\midrule
			1     & 79.31$\pm$8.17 & 81.64$\pm$7.14 & 83.15$\pm$3.37 & \textbf{83.17$\pm$3.61} \\
			2     & 92.09$\pm$3.89 & 90.07$\pm$3.75 & 93.88$\pm$3.80 & \textbf{95.49$\pm$2.13} \\
			3     & 95.40$\pm$1.98 & 95.69$\pm$3.53 & 88.12$\pm$6.87 & \textbf{97.04$\pm$1.73} \\
			4     & 76.85$\pm$5.91 & 74.97$\pm$5.70 & \textbf{89.14$\pm$7.11} & 78.16$\pm$5.19 \\
			5     & \textbf{98.75$\pm$0.72} & 98.62$\pm$1.32 & 98.26$\pm$1.22 & 98.01$\pm$1.34 \\
			6     & 97.10$\pm$3.96 & \textbf{98.40$\pm$1.18} & 97.52$\pm$2.10 & 96.70$\pm$1.34 \\
			7     & 97.06$\pm$3.38 & 97.71$\pm$3.67 & 95.79$\pm$2.89 & \textbf{99.05$\pm$1.25} \\
			8     & 92.95$\pm$3.22 & \textbf{94.98$\pm$3.08} & 85.40$\pm$4.80 & 93.83$\pm$3.23 \\
			9     & 82.69$\pm$4.70 & 82.96$\pm$3.90 & \textbf{94.23$\pm$3.90} & 82.66$\pm$4.54 \\
			\midrule
			OA    & 89.99$\pm$1.76 & 89.70$\pm$1.35 & 91.50$\pm$1.67 & \textbf{92.32$\pm$0.98} \\
			AA    & 90.24$\pm$1.68 & 90.56$\pm$0.96 & \textbf{91.72$\pm$1.21} & 91.57$\pm$0.71 \\
			Kappa & 86.94$\pm$2.26 & 86.61$\pm$1.68 & 88.86$\pm$2.08 & \textbf{89.91$\pm$1.25} \\
			\bottomrule
		\end{tabular}%
		\label{TestCADGCN_paviaU}%
	\end{table}%

	\begin{table}[!t]
		\scriptsize
		\centering
		\caption{Per-Class Accuracy, OA, AA (\%), and Kappa Coefficient Achieved by Different Model Settings on Salinas Dataset}
		\begin{tabular}{ccccc}
			\toprule
			ID    & CAD-GCN-v1 & CAD-GCN-v2 & CAD-GCN-v3 & CAD-GCN \\
			\midrule
			1     & \textbf{100.00$\pm$0.00} & \textbf{100.00$\pm$0.00} & 99.58$\pm$0.75 & \textbf{100.00$\pm$0.00} \\
			2     & \textbf{100.00$\pm$0.00} & \textbf{100.00$\pm$0.00} & 99.98$\pm$0.03 & \textbf{100.00$\pm$0.00} \\
			3     & 99.98$\pm$0.06 & \textbf{100.00$\pm$0.00} & 99.78$\pm$0.20 & 99.96$\pm$0.09 \\
			4     & 98.34$\pm$1.40 & \textbf{99.33$\pm$0.63} & 97.01$\pm$2.49 & 98.12$\pm$1.62 \\
			5     & 97.03$\pm$1.64 & 96.21$\pm$1.94 & 96.73$\pm$1.78 & \textbf{98.05$\pm$0.93} \\
			6     & 99.59$\pm$0.35 & 99.52$\pm$0.37 & 98.99$\pm$0.59 & \textbf{99.66$\pm$0.31} \\
			7     & 99.37$\pm$1.49 & 98.82$\pm$1.83 & \textbf{99.54$\pm$0.40} & 99.15$\pm$1.73 \\
			8     & 94.39$\pm$2.61 & 95.17$\pm$3.67 & 88.17$\pm$2.24 & \textbf{95.40$\pm$3.15} \\
			9     & \textbf{100.00$\pm$0.00} & \textbf{100.00$\pm$0.00} & 99.57$\pm$0.82 & \textbf{100.00$\pm$0.00} \\
			10    & 97.01$\pm$1.97 & 96.96$\pm$2.58 & 94.61$\pm$3.01 & \textbf{98.10$\pm$1.17} \\
			11    & 99.77$\pm$0.51 & 99.16$\pm$0.90 & 96.74$\pm$2.55 & \textbf{99.86$\pm$0.26} \\
			12    & 97.52$\pm$0.72 & 97.68$\pm$1.24 & \textbf{99.16$\pm$0.79} & 97.94$\pm$0.84 \\
			13    & 97.37$\pm$1.32 & \textbf{98.01$\pm$0.45} & 97.90$\pm$1.11 & 97.96$\pm$1.01 \\
			14    & 98.89$\pm$2.38 & \textbf{99.34$\pm$0.27} & 96.99$\pm$2.15 & 99.16$\pm$0.71 \\
			15    & \textbf{97.67$\pm$1.27} & 97.16$\pm$1.74 & 91.14$\pm$2.23 & 97.52$\pm$1.92 \\
			16    & 99.66$\pm$1.01 & 99.71$\pm$0.89 & 98.95$\pm$1.33 & \textbf{99.76$\pm$0.67} \\
			\midrule
			OA    & 97.90$\pm$0.53 & 97.95$\pm$0.71 & 95.36$\pm$0.55 & \textbf{98.23$\pm$0.54} \\
			AA    & 98.54$\pm$0.30 & 98.57$\pm$0.32 & 97.18$\pm$0.39 & \textbf{98.79$\pm$0.22} \\
			Kappa & 97.66$\pm$0.59 & 97.72$\pm$0.79 & 94.84$\pm$0.61 & \textbf{98.03$\pm$0.60} \\
			\bottomrule
		\end{tabular}%
		\label{TestCADGCN_SA}%
	\end{table}%
	
	As is mentioned in the introduction, the proposed CAD-GCN contains three critical parts for improving the contextual relations, i.e., the graph projection framework, the dynamic refinement of node similarities, and the edge filter. To shed light on the contributions of these three components, every time we report the classification results of CAD-GCN without one of the three components on the three adopted datasets (namely, the Indian Pines, the University of Pavia, and the Salinas). For simplicity, we adopt `CAD-GCN-v1', `CAD-GCN-v2', and `CAD-GCN-v3' to represent the reduced model by removing dynamic refinement of node similarities, the edge filter, and the graph projection framework, respectively. Table~\ref{TestCADGCN_IP}, Table~\ref{TestCADGCN_paviaU}, and Table~\ref{TestCADGCN_SA} exhibit the comparative results on the aforementioned datasets. It can be obviously observed that lacking any one of the components will inevitably hurt the OA. Therefore, the graph projection framework, the dynamic refinement of node similarities, and the edge filter work collaboratively to render satisfactory classification performance.
	
	\subsection{Running Time}
	
	\begin{table}[!t]
		\centering
		\caption{Running Time Comparison (In Seconds) of Different Methods. `IP' Denotes Indian Pines Dataset and `PaviaU' Denotes University of Pavia Dataset}
		\resizebox{85mm}{7.5mm}{
			\begin{tabular}{cccccc}
				\toprule
				Dataset & GCN \cite{Kipf2016Semi} & S$^{2}$GCN \cite{8474300} & R-2D-CNN \cite{Yang2018Hyperspectral} & CNN-PPF \cite{Li2016Hyperspectral} & \multicolumn{1}{l}{CAD-GCN} \\
				\midrule
				IP    & 58   & 71    & 2156  & 1495  & 44  \\
				paviaU & 1783     & 1803  & 2272  & 1545  & 90  \\
				Salinas & 3497  & 3528  & 2361  & 1769  & 556  \\
				\bottomrule
			\end{tabular}%
		}
		\label{RunningTime}%
	\end{table}%
	
	In order to reveal the advantage of our proposed CAD-GCN to the baselines in terms of efficiency, in Table~\ref{RunningTime}, we report the running time of different deep models, including GCN, S$^{2}$GCN, R-2D-CNN, CNN-PPF, and the proposed CAD-GCN on three datasets (i.e., the Indian Pines, the University of Pavia, and the Salinas), where the number of labeled pixels per class is kept identical to the experiments presented in Section \ref{ClassificationResults}. The codes for all methods are written in Python, and the running time is reported on a server with a 3.60-GHz Intel Xeon CPU with 264 GB of RAM and a Tesla P40 GPU. Compared with the running time of GCN, S$^{2}$GCN, our proposed CAD-GCN shows remarkably higher efficiency in large-scale datasets (i.e., the University of Pavia and the Salinas dataset), which is owing much to the employment of graph projection operation. Since the graph size can be significantly reduced by graph projection, our proposed CAD-GCN exhibits high efficiency on all the three datasets. The comparison results demonstrate that our proposed method is effective and efficient for HSI classification.

	\section{Conclusion}
	\label{Conclusions}
	
	In this paper, we have developed a novel Context-Aware Dynamic Graph Convolutional Network (CAD-GCN) for HSI classification. In order to capture long range contextual relations, we move beyond regular image grids by learning the pixel-to-region assignment, and further encode the contextual relations among regions, so that the regions which are originally far away in the 2D space can be connected by successive graph convolutions. Moreover, we enable the node similarities and connective relationships to be dynamically updated via learning the improved distance metric and the edge filter. Therefore, the contextual relations among pixels can be gradually refined along with graph convolution, which significantly improves the performance of CAD-GCN on representation and classification of HSI. The experimental results on three real-world HSI datasets indicate that the proposed CAD-GCN is able to yield better performance when compared with the state-of-the-art HSI classification methods.
	
	\ifCLASSOPTIONcaptionsoff
	\newpage
	\fi

	
	
	%

	\bibliographystyle{IEEEtran}
	\bibliography{IEEEexample}

\begin{thebibliography}{10}
\providecommand{\url}[1]{#1}
\csname url@samestyle\endcsname
\providecommand{\newblock}{\relax}
\providecommand{\bibinfo}[2]{#2}
\providecommand{\BIBentrySTDinterwordspacing}{\spaceskip=0pt\relax}
\providecommand{\BIBentryALTinterwordstretchfactor}{4}
\providecommand{\BIBentryALTinterwordspacing}{\spaceskip=\fontdimen2\font plus
\BIBentryALTinterwordstretchfactor\fontdimen3\font minus
  \fontdimen4\font\relax}
\providecommand{\BIBforeignlanguage}[2]{{%
\expandafter\ifx\csname l@#1\endcsname\relax
\typeout{** WARNING: IEEEtran.bst: No hyphenation pattern has been}%
\typeout{** loaded for the language `#1'. Using the pattern for}%
\typeout{** the default language instead.}%
\else
\language=\csname l@#1\endcsname
\fi
#2}}
\providecommand{\BIBdecl}{\relax}
\BIBdecl

\bibitem{298007}
J.~C. {Harsanyi} and C.-I. {Chang}, ``Hyperspectral image classification and
  dimensionality reduction: an orthogonal subspace projection approach,''
  \emph{IEEE Trans. Geosci. Remote Sens.}, vol.~32, no.~4, pp. 779--785, Jul.
  1994.

\bibitem{Zhang2018Diverse}
M.~{Zhang}, W.~{Li}, and Q.~{Du}, ``Diverse region-based {CNN} for
  hyperspectral image classification,'' \emph{IEEE Trans. Image Process.},
  vol.~27, no.~6, pp. 2623--2634, Jun. 2018.

\bibitem{1323134}
F.~{Melgani} and L.~{Bruzzone}, ``Classification of hyperspectral remote
  sensing images with support vector machines,'' \emph{IEEE Trans. Geosci.
  Remote Sens.}, vol.~42, no.~8, pp. 1778--1790, Aug. 2004.

\bibitem{5766028}
Y.~{Chen}, N.~M. {Nasrabadi}, and T.~D. {Tran}, ``Hyperspectral image
  classification using dictionary-based sparse representation,'' \emph{IEEE
  Trans. Geosci. Remote Sens.}, vol.~49, no.~10, pp. 3973--3985, Oct. 2011.

\bibitem{8101519}
L.~{He}, J.~{Li}, C.~{Liu}, and S.~{Li}, ``Recent advances on spectral-spatial
  hyperspectral image classification: An overview and new guidelines,''
  \emph{IEEE Trans. Geosci. Remote Sens.}, vol.~56, no.~3, pp. 1579--1597, Mar.
  2018.

\bibitem{6297992}
M.~{Fauvel}, Y.~{Tarabalka}, J.~A. {Benediktsson}, J.~{Chanussot}, and J.~C.
  {Tilton}, ``Advances in spectral-spatial classification of hyperspectral
  images,'' \emph{Proc. IEEE}, vol. 101, no.~3, pp. 652--675, Mar. 2013.

\bibitem{4071761}
R.~L. {Kettig} and D.~A. {Landgrebe}, ``Classification of multispectral image
  data by extraction and classification of homogeneous objects,'' \emph{IEEE
  Trans. Geosci. Electron.}, vol.~14, no.~1, pp. 19--26, Jan. 1976.

\bibitem{Wang2005A}
Y.~{Wang}, K.~{Loe}, T.~{Tan}, and J.~{Wu}, ``A dynamic hidden markov random
  field model for foreground and shadow segmentation,'' in \emph{Proc. WACV},
  vol.~1, Jan. 2005, pp. 474--480.

\bibitem{8271995}
X.~{Cao}, F.~{Zhou}, L.~{Xu}, D.~{Meng}, Z.~{Xu}, and J.~{Paisley},
  ``Hyperspectral image classification with markov random fields and a
  convolutional neural network,'' \emph{IEEE Trans. Image Process.}, vol.~27,
  no.~5, pp. 2354--2367, May. 2018.

\bibitem{5779697}
B.~{Zhang}, S.~{Li}, X.~{Jia}, L.~{Gao}, and M.~{Peng}, ``Adaptive markov
  random field approach for classification of hyperspectral imagery,''
  \emph{IEEE Geosci. Remote Sens. Lett.}, vol.~8, no.~5, pp. 973--977, Sep.
  2011.

\bibitem{6305471}
G.~{Moser} and S.~B. {Serpico}, ``Combining support vector machines and markov
  random fields in an integrated framework for contextual image
  classification,'' \emph{IEEE Trans. Geosci. Remote Sens.}, vol.~51, no.~5,
  pp. 2734--2752, May. 2013.

\bibitem{1396321}
J.~A. {Benediktsson}, J.~A. {Palmason}, and J.~R. {Sveinsson}, ``Classification
  of hyperspectral data from urban areas based on extended morphological
  profiles,'' \emph{IEEE Trans. Geosci. Remote Sens.}, vol.~43, no.~3, pp.
  480--491, Mar. 2005.

\bibitem{5887411}
L.~{Shen} and S.~{Jia}, ``Three-dimensional gabor wavelets for pixel-based
  hyperspectral imagery classification,'' \emph{IEEE Trans. Geosci. Remote
  Sens.}, vol.~49, no.~12, pp. 5039--5046, Dec. 2011.

\bibitem{1576697}
G.~{Camps-Valls}, L.~{Gomez-Chova}, J.~{Munoz-Mari}, J.~{Vila-Frances}, and
  J.~{Calpe-Maravilla}, ``Composite kernels for hyperspectral image
  classification,'' \emph{IEEE Geosci. Remote Sens. Lett.}, vol.~3, no.~1, pp.
  93--97, Jan. 2006.

\bibitem{7147814}
L.~{Fang}, S.~{Li}, W.~{Duan}, J.~{Ren}, and J.~A. {Benediktsson},
  ``Classification of hyperspectral images by exploiting spectral-spatial
  information of superpixel via multiple kernels,'' \emph{IEEE Trans. Geosci.
  Remote Sens.}, vol.~53, no.~12, pp. 6663--6674, Dec. 2015.

\bibitem{7117347}
Y.~{Xu}, Z.~{Wu}, and Z.~{Wei}, ``Spectral-spatial classification of
  hyperspectral image based on low-rank decomposition,'' \emph{IEEE J. Sel.
  Topics Appl. Earth Observ. Remote Sens.}, vol.~8, no.~6, pp. 2370--2380, Jun.
  2015.

\bibitem{Kipf2016Semi}
T.~N. Kipf and M.~Welling, ``Semi-supervised classification with graph
  convolutional networks,'' in \emph{Proc. ICLR}, 2017.

\bibitem{Wang2017Non}
X.~Wang, R.~Girshick, A.~Gupta, and K.~He, ``Non-local neural networks,'' pp.
  7794--7803, Jun. 2017.

\bibitem{Yin2018Beyond}
Y.~Li and A.~Gupta, ``Beyond grids: Learning graph representations for visual
  recognition,'' in \emph{Proc. Adv. Neural Inf. Process. Syst.}, 2018, pp.
  9225--9235.

\bibitem{Luo2016Understanding}
W.~Luo, Y.~Li, R.~Urtasun, and R.~Zemel, ``Understanding the effective
  receptive field in deep convolutional neural networks,'' pp. 4898--4906,
  2016.

\bibitem{Li2018Adaptive}
R.~Li, S.~Wang, F.~Zhu, and J.~Huang, ``Adaptive graph convolutional neural
  networks,'' in \emph{Proc. AAAI}, 2018.

\bibitem{5942156}
A.~{Villa}, J.~A. {Benediktsson}, J.~{Chanussot}, and C.~{Jutten},
  ``Hyperspectral image classification with independent component discriminant
  analysis,'' \emph{IEEE Trans. Geosci. Remote Sens.}, vol.~49, no.~12, pp.
  4865--4876, Dec. 2011.

\bibitem{Zhang2016Simultaneous}
L.~{Zhang}, Q.~{Zhang}, B.~{Du}, X.~{Huang}, Y.~Y. {Tang}, and D.~{Tao},
  ``Simultaneous spectral-spatial feature selection and extraction for
  hyperspectral images,'' \emph{IEEE Trans. Cybern.}, vol.~48, no.~1, pp.
  16--28, Jan. 2018.

\bibitem{6912942}
K.~{Liu}, Y.~{Lin}, and C.~{Chen}, ``Linear spectral mixture analysis via
  multiple-kernel learning for hyperspectral image classification,'' \emph{IEEE
  Trans. Geosci. Remote Sens.}, vol.~53, no.~4, pp. 2254--2269, Apr. 2015.

\bibitem{6926746}
Y.~{Zhou}, J.~{Peng}, and C.~L.~P. {Chen}, ``Extreme learning machine with
  composite kernels for hyperspectral image classification,'' \emph{IEEE J.
  Sel. Topics Appl. Earth Observ. Remote Sens.}, vol.~8, no.~6, pp. 2351--2360,
  Jun. 2015.

\bibitem{4683346}
X.~{Huang} and L.~{Zhang}, ``An adaptive mean-shift analysis approach for
  object extraction and classification from urban hyperspectral imagery,''
  \emph{IEEE Trans. Geosci. Remote Sens.}, vol.~46, no.~12, pp. 4173--4185,
  Dec. 2008.

\bibitem{6923420}
Y.~Y. {Tang}, Y.~{Lu}, and H.~{Yuan}, ``Hyperspectral image classification
  based on three-dimensional scattering wavelet transform,'' \emph{IEEE Trans.
  Geosci. Remote Sens.}, vol.~53, no.~5, pp. 2467--2480, May. 2015.

\bibitem{7046411}
J.~{Zabalza}, J.~{Ren}, J.~{Zheng}, J.~{Han}, H.~{Zhao}, S.~{Li}, and
  S.~{Marshall}, ``Novel two-dimensional singular spectrum analysis for
  effective feature extraction and data classification in hyperspectral
  imaging,'' \emph{IEEE Trans. Geosci. Remote Sens.}, vol.~53, no.~8, pp.
  4418--4433, Aug. 2015.

\bibitem{7080913}
J.~{Peng}, Y.~{Zhou}, and C.~L.~P. {Chen}, ``Region-kernel-based support vector
  machines for hyperspectral image classification,'' \emph{IEEE Trans. Geosci.
  Remote Sens.}, vol.~53, no.~9, pp. 4810--4824, Sep. 2015.

\bibitem{7064745}
J.~{Xia}, M.~{Dalla Mura}, J.~{Chanussot}, P.~{Du}, and X.~{He}, ``Random
  subspace ensembles for hyperspectral image classification with extended
  morphological attribute profiles,'' \emph{IEEE Trans. Geosci. Remote Sens.},
  vol.~53, no.~9, pp. 4768--4786, Sep. 2015.

\bibitem{905239}
M.~{Pesaresi} and J.~A. {Benediktsson}, ``A new approach for the morphological
  segmentation of high-resolution satellite imagery,'' \emph{IEEE Trans.
  Geosci. Remote Sens.}, vol.~39, no.~2, pp. 309--320, Feb. 2001.

\bibitem{4686022}
M.~{Fauvel}, J.~A. {Benediktsson}, J.~{Chanussot}, and J.~R. {Sveinsson},
  ``Spectral and spatial classification of hyperspectral data using {SVM}s and
  morphological profiles,'' \emph{IEEE Trans. Geosci. Remote Sens.}, vol.~46,
  no.~11, pp. 3804--3814, Nov. 2008.

\bibitem{5482208}
M.~{Dalla Mura}, J.~A. {Benediktsson}, B.~{Waske}, and L.~{Bruzzone},
  ``Morphological attribute profiles for the analysis of very high resolution
  images,'' \emph{IEEE Trans. Geosci. Remote Sens.}, vol.~48, no.~10, pp.
  3747--3762, Oct. 2010.

\bibitem{5664759}
M.~{Dalla Mura}, A.~{Villa}, J.~A. {Benediktsson}, J.~{Chanussot}, and
  L.~{Bruzzone}, ``Classification of hyperspectral images by using extended
  morphological attribute profiles and independent component analysis,''
  \emph{IEEE Trans. Geosci. Remote Sens.}, vol.~8, no.~3, pp. 542--546, May.
  2011.

\bibitem{6410414}
M.~{Pedergnana}, P.~R. {Marpu}, M.~D. {Mura}, J.~A. {Benediktsson}, and
  L.~{Bruzzone}, ``A novel technique for optimal feature selection in attribute
  profiles based on genetic algorithms,'' \emph{IEEE Trans. Geosci. Remote
  Sens.}, vol.~51, no.~6, pp. 3514--3528, Jun. 2013.

\bibitem{7097693}
J.~{Li}, H.~{Zhang}, and L.~{Zhang}, ``Efficient superpixel-level multitask
  joint sparse representation for hyperspectral image classification,''
  \emph{IEEE Trans. Geosci. Remote Sens.}, vol.~53, no.~10, pp. 5338--5351,
  Oct. 2015.

\bibitem{7119598}
G.~{Zhang}, X.~{Jia}, and J.~{Hu}, ``Superpixel-based graphical model for
  remote sensing image mapping,'' \emph{IEEE Trans. Geosci. Remote Sens.},
  vol.~53, no.~11, pp. 5861--5871, Nov. 2015.

\bibitem{8291607}
B.~{Cui}, X.~{Xie}, X.~{Ma}, G.~{Ren}, and Y.~{Ma}, ``Superpixel-based extended
  random walker for hyperspectral image classification,'' \emph{IEEE Trans.
  Geosci. Remote Sens.}, vol.~56, no.~6, pp. 3233--3243, Jun. 2018.

\bibitem{Hamilton2017Inductive}
W.~Hamilton, Z.~Ying, and J.~Leskovec, ``Inductive representation learning on
  large graphs,'' in \emph{Proc. Adv. Neural Inf. Process. Syst.}, 2017, pp.
  1024--1034.

\bibitem{velivckovic2017graph}
P.~Veli{\v{c}}kovi{\'c}, G.~Cucurull, A.~Casanova, A.~Romero, P.~Lio, and
  Y.~Bengio, ``Graph attention networks,'' \emph{arXiv preprint
  arXiv:1710.10903}, 2017.

\bibitem{Bruna2014Spectral}
J.~Bruna, W.~Zaremba, A.~Szlam, and Y.~LeCun, ``Spectral networks and locally
  connected networks on graphs,'' \emph{arXiv preprint arXiv:1312.6203}, 2013.

\bibitem{Defferrard2016Convolutional}
M.~Defferrard, X.~Bresson, and P.~Vandergheynst, ``Convolutional neural
  networks on graphs with fast localized spectral filtering,'' in \emph{Proc.
  Adv. Neural Inf. Process. Syst.}, 2016, pp. 3844--3852.

\bibitem{Lei2009Relational}
T.~Lei and H.~Liu, ``Relational learning via latent social dimensions,'' in
  \emph{Proc. ACM SIGKDD}.\hskip 1em plus 0.5em minus 0.4em\relax ACM, 2009,
  pp. 817--826.

\bibitem{Ying2018GraphSIGKDD}
K.~C. P. E. W. L.~H. R.~Ying, R.~He and J.~Leskovec, ``Graph convolutional
  neural networks for web-scale recommender systems,'' in \emph{Proc. ACM
  SIGKDD}.\hskip 1em plus 0.5em minus 0.4em\relax ACM, 2018, pp. 974--983.

\bibitem{8029788}
H.~{Zhuang}, C.~{Wang}, C.~{Li}, Q.~{Wang}, and X.~{Zhou}, ``Natural language
  processing service based on stroke-level convolutional networks for chinese
  text classification,'' in \emph{Proc. IEEE ICWS}, Jun. 2017, pp. 404--411.

\bibitem{8100020}
Y.~{Zhang}, S.~{Song}, E.~{Yumer}, M.~{Savva}, J.~{Lee}, H.~{Jin}, and
  T.~{Funkhouser}, ``Physically-based rendering for indoor scene understanding
  using convolutional neural networks,'' in \emph{Proc. IEEE CVPR}, Jul. 2017,
  pp. 5057--5065.

\bibitem{8474300}
A.~{Qin}, Z.~{Shang}, J.~{Tian}, Y.~{Wang}, T.~{Zhang}, and Y.~Y. {Tang},
  ``Spectral-spatial graph convolutional networks for semisupervised
  hyperspectral image classification,'' \emph{IEEE Geosci. Remote Sens. Lett.},
  vol.~16, no.~2, pp. 241--245, Feb. 2019.

\bibitem{7280459}
Z.~Hao, Z.~Yang, W.~Liu, J.~Liang, and Y.~Li, ``Improving deep neural networks
  using softplus units,'' in \emph{Proc. IJCNN}, Jul. 2015, pp. 1--4.

\bibitem{Hammond2009Wavelets}
D.~K. Hammond, P.~Vandergheynst, and R.~Gribonval, ``Wavelets on graphs via
  spectral graph theory,'' \emph{Appl. Comput. Harmon. Anal.}, vol.~30, no.~2,
  pp. 129--150, Dec. 2011.

\bibitem{Radhakrishna2012SLIC}
R.~{Achanta}, A.~{Shaji}, K.~{Smith}, A.~{Lucchi}, P.~{Fua}, and
  S.~{Süsstrunk}, ``Slic superpixels compared to state-of-the-art superpixel
  methods,'' \emph{IEEE Trans. Pattern Anal. Mach. Intell.}, vol.~34, no.~11,
  pp. 2274--2282, Nov. 2012.

\bibitem{Yang2018Hyperspectral}
X.~{Yang}, Y.~{Ye}, X.~{Li}, R.~Y.~K. {Lau}, X.~{Zhang}, and X.~{Huang},
  ``Hyperspectral image classification with deep learning models,'' \emph{IEEE
  Trans. Geosci. Remote Sens.}, vol.~56, no.~9, pp. 5408--5423, Sep. 2018.

\bibitem{Li2016Hyperspectral}
W.~{Li}, G.~{Wu}, F.~{Zhang}, and Q.~{Du}, ``Hyperspectral image classification
  using deep pixel-pair features,'' \emph{IEEE Trans. Geosci. Remote Sens.},
  vol.~55, no.~2, pp. 844--853, Feb. 2017.

\bibitem{6882821}
J.~{Li}, X.~{Huang}, P.~{Gamba}, J.~M. {Bioucas-Dias}, L.~{Zhang}, J.~A.
  {Benediktsson}, and A.~{Plaza}, ``Multiple feature learning for hyperspectral
  image classification,'' \emph{IEEE Trans. Geosci. Remote Sens.}, vol.~53,
  no.~3, pp. 1592--1606, Mar. 2015.

\bibitem{7360896}
C.~{Bo}, H.~{Lu}, and D.~{Wang}, ``Hyperspectral image classification via jcr
  and svm models with decision fusion,'' \emph{IEEE Geosci. Remote Sens.
  Lett.}, vol.~13, no.~2, pp. 177--181, Feb. 2016.

\bibitem{Gao:2018:LLG:3219819.3219947}
H.~Gao, Z.~Wang, and S.~Ji, ``Large-scale learnable graph convolutional
  networks,'' in \emph{Proc. SIGKDD}.\hskip 1em plus 0.5em minus 0.4em\relax
  ACM, 2018, pp. 1416--1424.

\bibitem{7010929}
C.~{Gong}, T.~{Liu}, D.~{Tao}, K.~{Fu}, E.~{Tu}, and J.~{Yang}, ``Deformed
  graph laplacian for semisupervised learning,'' \emph{IEEE Trans. Neural Netw.
  Learn. Syst.}, vol.~26, no.~10, pp. 2261--2274, Oct. 2015.

\end{thebibliography}

\end{document}